\documentclass{IEEEphot}
\usepackage{cite}
\usepackage{amsmath,amssymb,amsfonts}
\usepackage{setspace}
\usepackage{algorithmic}
\usepackage{mathtools}
\usepackage{verbatim}
\usepackage{graphicx,color}
\usepackage{textcomp}
\usepackage{xcolor}
\usepackage{multicol}
\usepackage{xfrac}
\usepackage[thinc]{esdiff}
\usepackage{placeins}
\usepackage{orcidlink} 
\usepackage{setspace}

\usepackage{etoolbox}

\makeatletter
\patchcmd{\@IEEEtranBSTctlwrite}
  {, \the\@year}
  {~\the\@year}
  {}{}
\makeatother

\usepackage{hyperref} 
\hypersetup{pdfpagemode=UseNone}

\renewcommand{\thesection}{\Roman{section}}

\DeclareCollectionInstance{plainmath}{xfrac}{mathdefault}{math}
{
scale-factor = 4.0,
scaling = true
}
\usepackage{cleveref}
\usepackage[english]{babel}
\usepackage[makeroom]{cancel}
\usepackage[flushleft]{threeparttable}
\usepackage[bottom]{footmisc}
\usepackage{algorithm,algorithmic}
\usepackage{booktabs}
\usepackage{multirow}
\usepackage{makecell}
\usepackage{colortbl}
\usepackage{arydshln}
\usepackage{tikz}
\usepackage{acronym}
\usepackage{float}
\usepackage{siunitx}

\newenvironment{leftitemize}
  {\begin{list}{$\bullet$}{%
      \setlength{\leftmargin}{0pt}%
      \setlength{\itemindent}{5pt}%
      \setlength{\labelwidth}{0pt}%
      \setlength{\labelsep}{0.5em}%
      \setlength{\itemsep}{0.5em}%
      \setlength{\topsep}{0.5em}%
    }}
  {\end{list}}

\hypersetup{
    pdftoolbar=true,        
    pdfmenubar=true,        
    pdffitwindow=false,     
    pdfstartview={FitH},    
    pdftitle={Informed, Constrained, Aligned: A Field Analysis on Degeneracy-aware Point Cloud Registration in the Wild},    
    pdfauthor={Turcan Tuna},     
    pdfsubject={IEEE T-FR Journal},   
    pdfcreator={Turcan Tuna},   
    pdfproducer={Turcan Tuna}, 
    colorlinks=true,
    linkcolor=blue,
    linkbordercolor=white,
    filecolor=magenta,
    citecolor=red,
    urlcolor=cyan
    }
\usetikzlibrary{calc}
\usepackage[export]{adjustbox}

\usepackage{pifont}
\newcommand{\cmark}{{\color{darkgreen}\ding{52}}}%
\newcommand{\xmark}{{\color{red}\ding{56}}}%
\newcommand{\sphere}{{\color{bananamania}\ding{108}}}%

\definecolor{darkgreen}{rgb}{0.0, 0.5, 0.0}
\definecolor{darkbrown}{rgb}{0.8, 0.4, 0.2}
\definecolor{alizarin}{rgb}{0.82, 0.1, 0.26}
\definecolor{bittersweet}{rgb}{0.94, 0.5, 0.5} 
\definecolor{bananamania}{rgb}{0.93, 0.86, 0.51}
\definecolor{applegreen}{rgb}{0.67, 0.88, 0.69}
\definecolor{bluebell}{rgb}{0.64, 0.64, 0.82}
\definecolor{bisque}{rgb}{0.95, 0.9, 0.67}

\acrodef{ICP}{Iterative closest point}
\acrodef{DoF}{Degrees-of-freedom}
\acrodef{LiDAR}{Light detection and ranging}
\acrodef{SVD}{Singular value decomposition}
\acrodef{LU}{Lower-Upper}
\acrodef{LM}{Levenberg Marquardt}
\acrodef{TSVD}{Truncated \ac{SVD}}
\acrodef{IRLS}{Iterative re-weighted least squares}
\acrodef{FGR}{Fast global registration}
\acrodef{MAD}{Median of absolute deviation}
\acrodef{GT}{Ground truth}
\acrodef{L-Reg.}{linear optimization with Tikhonov regularization}
\acrodef{NL-Reg.}{Non-linear optimization with Tikhonov regularization}
\acrodef{P2Plane}{Point-to-plane ICP}
\acrodef{Ineq. Con.}{Inequality constraints}
\acrodef{Eq. Con.}{Equality constraints}
\acrodef{SLAM}{Simultaneous Localization and Mapping}
\acrodef{GNSS}{Global Navigation Satellite Systems}
\acrodef{ATE}{Absolute Translation Error}
\acrodef{RTE}{Relative Translation Error}
\acrodef{NL-Solver}{non-linear solver}
\acrodef{RMS}{Redundancy-minimizing point cloud
sampling}

\def\BibTeX{{\rm B\kern-.05em{\sc i\kern-.025em b}\kern-.08em
    T\kern-.1667em\lower.7ex\hbox{E}\kern-.125emX}}
\AtBeginDocument{\definecolor{tmlcncolor}{cmyk}{0.93,0.59,0.15,0.02}\definecolor{NavyBlue}{RGB}{0,86,125}}

\usepackage{balance}

\def\OJlogo{\vspace{-4pt}$<$Society logo(s) and publication title will appear here.$>$}
\def\seclogo{\vspace{10pt}$<$Society logo(s) and publication title will appear here.$>$}

\def\authorrefmark#1{\ensuremath{^{\textbf{#1}}}}
\begin{document}
\receiveddate{XX Month, XXXX}
\reviseddate{XX Month, XXXX}
\accepteddate{XX Month, XXXX}
\publisheddate{XX Month, XXXX}
\currentdate{XX Month, XXXX}
\doiinfo{XXXX.2022.1234567}

\markboth{}{Tuna {et al.}}


\title{Informed, Constrained, Aligned: A Field Analysis on Degeneracy-aware Point Cloud Registration in the Wild}

\author{Turcan~Tuna$^{\orcidlink{0000-0001-8662-4890}}$\authorrefmark{1} (Student Member, IEEE), Julian~Nubert$^{\orcidlink{0000-0001-8949-6134}}$\authorrefmark{1,2} (Student Member, IEEE), Patrick~Pfreundschuh$^{\orcidlink{0000-0002-4127-8119}}$\authorrefmark{3} (Student Member, IEEE), Cesar~Cadena$^{\orcidlink{0000-0002-2972-6011}}$\authorrefmark{1} (Member, IEEE), \\ Shehryar~Khattak$^{\orcidlink{0000-0002-9304-1455}}$\authorrefmark{4} (Member, IEEE), and Marco~Hutter$^{\orcidlink{0000-0002-4285-4990}}$\authorrefmark{1} (Member, IEEE)}
\affil{Robotic Systems Lab, Swiss Federal Institute of Technology (ETH Zürich), 8092 Zürich, Switzerland}
\affil{Max Planck Institute for Intelligent Systems, Stuttgart, Germany.}
\affil{Autonomous Systems Lab, Swiss Federal Institute of Technology (ETH Zürich), 8092 Zürich, Switzerland}
\affil{Jet Propulsion Laboratory (JPL), California Institute of Technology (Caltech), Pasadena, CA 91011 USA. This work was done as an outside activity and not in the author’s capacity as an employee of the Jet Propulsion Laboratory, California Institute of Technology, USA.}
\corresp{Corresponding author: Turcan Tuna (tutuna@ethz.ch).}
\authornote{This work is supported in part by \\ the Sony Research Grant 2023, the EU Horizon 2020 programme grant agreements No. 852044, 101016970, and 101070405, EU Horizon 2021 programme grant agreement No. 101070596, the NCCR digital fabrication, the ETH Zurich Research Grant No. 21-1 ETH-27, the Swiss National Science Foundation
(SNSF) through project No. 227617, and the Max Planck ETH Center for Learning Systems. \looseness-1 \\[1em]  Project Page/Supp./Code/Datasets:       \url{https://sites.google.com/leggedrobotics.com/perfectlyconstrained}}

\begin{abstract}
The \acs{ICP} registration algorithm has been a preferred method for \acs{LiDAR}-based robot localization for nearly a decade. However, even in modern \acs{SLAM} solutions, \acs{ICP} can degrade and become unreliable in geometrically ill-conditioned environments. 
Current solutions primarily focus on utilizing additional sources of information, such as external odometry, to either replace the degenerate directions of the optimization solution or add additional constraints in a sensor-fusion setup afterward.
In response, this work investigates and compares new and existing degeneracy mitigation methods for robust \acs{LiDAR}-based localization and analyzes the efficacy of these approaches in degenerate environments for the first time in the literature at this scale.
Specifically, this work investigates \textit{i)} the effect of using active or passive degeneracy mitigation methods for the problem of ill-conditioned \acs{ICP} in \acs{LiDAR} degenerate environments, \textit{ii)} the evaluation of \acs{TSVD}, inequality constraints, and linear/non-linear Tikhonov regularization for the application of degenerate point cloud registration for the first time. Furthermore, a sensitivity analysis for least-squares minimization step of the ICP problem is carried out to better understand how each method affects the optimization and what to expect from each method.
The results of the analysis are validated through multiple real-world robotic field and simulated experiments. 
The analysis demonstrates that active optimization degeneracy mitigation is necessary and advantageous in the absence of reliable external estimate assistance for \acs{LiDAR}-\acs{SLAM}, and soft-constrained methods can provide better results in complex ill-conditioned scenarios with heuristic fine-tuned parameters. 
The code and data used in this work are made publicly available to the community.
\end{abstract}



\maketitle

\section{INTRODUCTION}\label{section:introduction}

\begin{figure*}[t]
\centering
    \includegraphics[ width=1\linewidth]{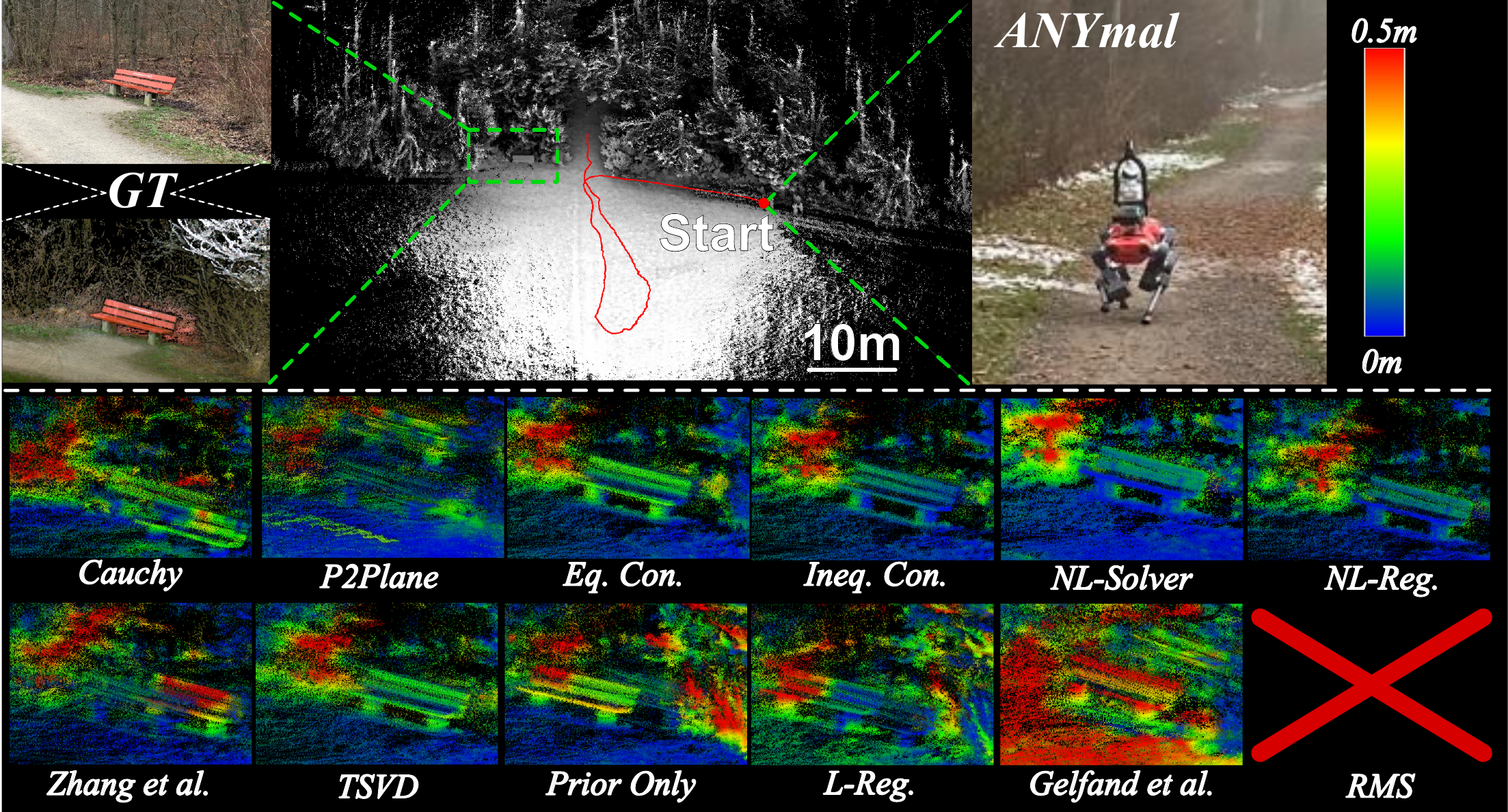}
    \caption{
    The mapping results of the ANYmal forest open-field degeneracy experiment (cf. \Cref{section:Experiments:forest}) are shown. \textbf{Top:} The map and robot trajectory generated by the Eq. Con. method is shown next to the ground truth bench and image of the region of interest. \textbf{Rest:} The point cloud error map for each method is shown. The color coding reflects the point-to-point distances to the ground truth map of the environment. The RMS method diverged due to the irregularities of the natural environment.}
\label{fig:forest_mapping_results}
\end{figure*}

\IEEEPARstart{M}{odern} sensors such as \ac{LiDAR} provide accurate distance measurements at high rates, making them suitable for reliable robot pose estimation and accurate mapping in large-scale environments. \ac{SLAM} is a crucial tool to operate robots in challenging and degenerate environments~\cite{slamOverview2}. While \ac{SLAM} and robot localization can also be achieved with other sensing modalities~\cite{visualSLAM,thermal2020,radar2024}, the field robotics community has been focusing on improving \ac{LiDAR}-based \ac{SLAM} techniques for better accuracy and robustness in large-scale environments~\cite{bavle2023slam, cadena2016past, slamOverview2, nair2024hilti}.

Most widely accepted \ac{LiDAR}-\ac{SLAM} frameworks leverage the efficiency of point cloud registration to provide an accurate environment map. These algorithms align pairs of point clouds with sufficient overlap by finding the six \ac{DoF} rigid transformation. A well-established point cloud registration method is the \ac{ICP} algorithm, often utilized in computer vision, 3D pose estimation, and robotics~\cite{qgore}. In a typical \ac{LiDAR}-\ac{SLAM} framework~\cite{compSLAM,jelavic2022open3d}, \ac{LiDAR} scan-to-map registration is performed, which is vital to align the new information to the already generated map of the environment.
\subsubsection{ALGORITHMIC LIMITATIONS} 
Despite the success of \ac{ICP} and the algorithm's broad applicability, there are still some limitations. These limitations include \textit{i)} the risk of convergence to local minima, \textit{ii)} the sensitivity to inaccurate initial transformation prior, \textit{iii)} being susceptible to point-wise feature extraction noise, and \textit{iv)} robustness against the lack of geometric constraints in the underlying optimization problem~\cite{censi, brossard}. Notably, while the first three sources of error can be directly addressed by modifying the \ac{LiDAR}-\ac{SLAM} framework, the underlying optimization problem's ill-conditioning stems directly from the environment's lack of information along specific directions and cannot be overcome easily.
 

%
\subsubsection{DEFINING LiDAR DEGENERACY} According to the literature~\cite{xicp, petracek2024rms, gelfand2003, kostas2024probabilistic}, the term LiDAR degeneracy refers to the ill-conditioning of the optimization within the ICP algorithm. If the LiDAR point cloud does not contain sufficient information, e.g., insufficient geometric constraints in a self-similar environment, a linear combination of some of the optimization states might become unobservable. In the presence of such unobservable optimization states, the ICP optimization is said to become \textit{ill-conditioned}, \textit{degenerate}, or \textit{non-localizable}.

\subsubsection{DEGENERATE POINT CLOUD REGISTRATION} 
While the robotic community has developed LiDAR-based mapping frameworks~\cite{kiss_icp, fastlio2, jelavic2022open3d} that function robustly in most environments, point cloud registration and, subsequently, mapping in degenerate environments (without a sufficient amount of distinctive features) still pose a challenge to these frameworks due to a lack of present information in the data. State-of-the-art methods~\cite{xicp, solRemap, solRemap_kaess} particularly targeted to those scenarios divide the problem into two steps: \textit{i)} analysis and detection of the degenerate directions and, \textit{ii)} mitigation of the adverse effects on the optimization. In particular, while the first point received considerable attention from the research community~\cite{nobili1,nobili2, LION, fisher1, nubert2022learning, zhen2017robust, tunnelLocalizability, xicp, solRemap}, the number of works salvaging the information in the best possible way given the detected degenerate directions is far more limited~\cite{rankDeficientSLAM, xicp, solRemap}. 
The latter methods try to mitigate the effect of the degeneracy on the optimization and share the ability to fully benefit from the pose updates, at least in the well-constrained optimization directions, while the former methods entirely skip the optimization and rely on external odometry to propagate the \ac{ICP} pose estimates. 
As discussed in~\cite{xicp}, skipping the optimization entirely, if ill-conditioned, is a plausible strategy only for short periods, particularly when the registration initial guess is unreliable and the additional fine-registration of the point cloud is beneficial.\looseness-1
\subsubsection{LiDAR DEGENERACY MITIGATION IN THE FIELD}
The robotics community has recently increased their attention to robotic operations in challenging and LiDAR-degenerate environments, as LiDAR-degenerate robotic datasets from field experiments~\cite{zhao2024subt, geode, nissov2024degradation, cerberus, enwide} become more readily available to researchers. 
The robotics community is still actively developing methods~\cite{xicp, kostas2024probabilistic, petracek2024rms, ji_pointtodistribution} to improve the robustness and accuracy of the LiDAR-SLAM and point cloud registration approaches. Despite this, there remains a fundamental research gap on the efficacy of degeneracy mitigation methods \textit{once LiDAR degeneracy is detected}. 
Many methods have been proposed to mitigate LiDAR degeneracy using external information, such as sensor fusion with factor-graph optimization~\cite{jin2024range, dareSLAM} and filters~\cite{chen2024relead}. However, addressing the optimization step in ICP-based point cloud registration~\cite{xicp} or mitigating LiDAR degeneracy within the ICP method remains an open question.
This work focuses on methods to mitigate LiDAR degeneracy within the ICP formulation to narrow the research gap mentioned above. 
The approaches for degeneracy mitigation in ICP are classified into two categories, namely \textit{Passive} and \textit{Active} degeneracy mitigation.\looseness-1 
\begin{leftitemize}
    \item \textbf{Active degeneracy mitigation}: A method is classified as an active degeneracy mitigation method if the output of any explicit LiDAR detection from the method (e.g., \Cref{subsubsection:degeneracy_detection}) is used to take action towards mitigating the degeneracy within the ICP cost function optimization.
    \item \textbf{Passive degeneracy mitigation}: In contrast to the previous item, a method is called to perform passive degeneracy prevention if the LiDAR degeneracy information is not explicitly used within the ICP formulation to mitigate the degeneracy.\looseness-1
\end{leftitemize}

\subsubsection{CONTRIBUTIONS} 
In response to the research gap mentioned above, this work studies and compares new and existing degeneracy mitigation methods with various simulated and real-world experiments, such as in~\Cref{fig:forest_mapping_results}, to investigate the efficacy of degeneracy mitigation in point cloud registration in complex degenerate environments. 
In addition to the study results, this work introduces three distinct methods to constrain the optimization problem for robust point cloud registration. %
Lastly, the study's results are put into context with theoretical derivations and the limitations of the state-of-the-art methods and validated through multiple robotic field experiments and simulated examples. The contributions of this work are as follows:\looseness-1
\begin{leftitemize}
    \item A thorough study on the effectiveness of different constraint types is performed for the degenerate point cloud registration task in the context of robot operation in geometrically ill-conditioned real-world environments such as open natural fields, urban tunnels, and complex construction sites.\looseness-1
    \item The introduction, investigation, and discussion of sub-space Tikhonov regularization, Truncated Singular Value Decomposition, and inequality-constrained \ac{ICP} methods for the field of degenerate point cloud registration for the first time in the literature.
    \item Comparison of the discussed methods in a fair and isolated fashion through simulated examples and multiple detailed robotic field experiments conducted with a legged robot, a walking excavator, and a hand-held setup with different LiDARs.\looseness-1
    \item The consolidation of open-source methods and all investigated methods are going to be open-sourced\footnote{\url{https://sites.google.com/leggedrobotics.com/perfectlyconstrained}} in a single framework to foster future research on developing degeneracy-aware \ac{LiDAR}-\ac{SLAM} systems. 
\end{leftitemize}


\section{RELATED WORK}\label{section:related_works}
\subsection{(ROBUST) POINT CLOUD REGISTRATION}\label{subsection:point_cloud_registration}

Fast and accurate pose estimation through point cloud registration has been a central research topic in robot perception over multiple decades~\cite{ICPderivation}. Different families of solutions have been explored in this field, such as \ac{ICP} and its variants~\cite{generalizedICP, point_to_gaussian, pointToLine, symmetricPointToPlane}, feature-based~\cite {loam, behley2018efficient}, probabilistic~\cite{NDT, cpd, bayesianICP}, learned~\cite{deepICP, pointdsc, selfSupervisedOdom} or robust global registration~\cite{fgr, lim2022single} methods.\looseness-1

Despite the majority of advancements, the simpler \ac{ICP} variants using point-to-point~\cite{pointToPointICP}, point-to-plane~\cite{point_to_plane}, symmetric point-to-plane~\cite{symmetricPointToPlane} and G-ICP~\cite{generalizedICP} are among the most widely used \ac{LiDAR}-based point cloud registration algorithms due to their simplicity and practicality~\cite{pomerleau2013applied, qgore}. 
All~\ac{ICP} variations have in common to find the refined six \ac{DoF} transformation between two point clouds by repeatedly finding sets of corresponding point pairs, followed by iterative minimization of a pre-defined alignment metric.


On the other hand, the susceptibility of the \ac{ICP} method to local minima encouraged researchers to investigate globally optimal and certifiable algorithms for robust point cloud registration~\cite{teaser, quasar, lagrangianDuality, qgore, lim2022single, lim2024quatro}.
These algorithms ensure that the estimated solution corresponds to the globally optimal solution, even in outlier dominant cases. Similarly, robust M-estimators have been proposed to deal with heavy noise-dominated point cloud registration problems. These estimators down-weigh the outliers to mitigate their influence on the cost function. The well-known M-estimators such as Welsch, Tukey, Cauchy, Huber, and Geman-McClure have been used by the robotics community~\cite{kiss_icp, robustICP, babin2019analysis}.
Recently, the authors in \cite{babin2019analysis} investigated the effectiveness of different robust norms on non-degenerate registration problems and concluded that, if tuned correctly, most robust norms perform similarly by increasing the signal-to-noise ratio in the optimization problem.
In addition, to further mitigate the need for manual hand-tuning, researchers proposed adaptive M-estimators~\cite{chebrolu} to be robust against different distributions of inputs and adaptive graduated non-convexity~\cite{graduated, teaser, convexGlobal, lim2022single, lim2024quatro} to improve robustness against large perturbations. 
Furthermore, driven by the success of the \ac{ICP} algorithm towards practical applications, researchers have focused on improving its robustness through statistical methods~\cite{bayesianICP, gaussianmm}. These improvements include better feature sampling~\cite{DNSS, kwok2016improvements, petracek2024rms}, adaptive matching techniques~\cite{kiss_icp}, and optimization acceleration methods~\cite{robustICP}.

However, none of these methods particularly tackle situations where insufficient information is contained in the underlying point cloud due to the symmetric or self-similar nature of the environment, which is often present in practice in environments such as tunnels, open planes, or narrow corridors~\cite{xicp,nubert2022learning,pfreundschuh2023coin}.


\subsection{OPTIMIZATION DEGENERACY DETECTION}
\label{subsubsection:degeneracy_detection}
\subsubsection{POINT CLOUD REGISTRATION UNCERTAINTY}
Previous methods have been proposed to model the uncertainty of the registration and capture the degeneracy of the optimization problem by estimating the covariance of the pose estimation process~\cite{censi, brossard}. Although this may be a good indicator of ill-conditioning, the obtained measures often provide overly optimistic estimates~\cite{bonnabel2016covariance}.
Recently, the authors in~\cite{talbot2023principled} provided additional insights on accurately modeling the uncertainty of \ac{LiDAR} point cloud registration information for use in a multi-modal sensor fusion framework as a loosely coupled measurement. More specifically, the authors evaluated the ICP uncertainty formulation in~\cite{cenci2} in the context of LiDAR SLAM by simplifying the formulation to omit the cross-covariance terms for compute limitation purposes.

Similarly, in LOG-LIO2~\cite{huang2024log}, authors provide an incremental methodology for efficient calculation of measurement uncertainty of \ac{LiDAR} measurements. Their proposed uncertainty model captures the range and bearing uncertainty as well as the uncertainty arising from LiDAR beam incident angle and surface roughness. Later, the authors integrated this uncertainty model in a \ac{LiDAR} inertial odometry framework to illustrate the benefits during deployment.
As one of the most recent works, the authors in~\cite{kostas2024probabilistic} proposed a novel probabilistic formulation to estimate the probability of optimization directions being degenerate by taking the measurement and surface normal estimation noise into account. {\color{black}As the increasing amount of recent work shows, the estimation of unified metric uncertainty of point cloud registration is a promising direction in quantifying optimization degradation. These methods further allow the fusion of the ill-conditioned poses into sensor-fusion frameworks through the metric uncertainty, as discussed in \Cref{subsubsection:sensorfusion}}.
%
\subsubsection{LEARNING-BASED APPROACHES}
Recent research has investigated data-driven approaches to identify degeneracy in \ac{LiDAR} information. The authors in~\cite{deepLocalizability} introduced a deep learning-based entropy metric by fusing covariance estimation with the information level of the environment. To avoid complicated Monte-Carlo sampling during robot operation, the authors in~\cite{nubert2022learning} suggested utilizing synthetic data for training while relying solely on the current \ac{LiDAR} scan to compute a $6$-\ac{DoF} localizability metric. Although data-driven methods have the potential to outperform traditional methods by modeling higher-degree dynamics of the problem, the interpretability of these methods still presents a challenge. 

\subsubsection{OPTIMIZATION STABILITY BASED APPROACHES}
Other works~\cite{planeRank, LION, dareSLAM} also propose to use the optimization condition number as a single combined degeneracy metric for the full $6$-\ac{DoF} pose estimation. Tangentially, CompSLAM~\cite{compSLAM} employs the D-optimality criterion~\cite{dOpt} as a metric to detect optimization degeneracy in a multi-modal sensor fusion framework. Following a similar formulation,~\cite{hinduja2019degeneracy} used the optimization's relative condition number as the degeneracy detection threshold in contrast to the minimum eigenvalue.
Among pioneering methods for degeneracy detection,~\cite{solRemap} introduced a detection metric called \textit{degeneracy factor}. This metric identifies the degenerate directions of the optimization by analyzing the minimum eigenvalue of the optimization's Hessian matrix.
Similarly, DAMS-LIO~\cite{han2023dams} follows the footsteps of~\cite{solRemap} and uses the Hessian of the optimization to acquire the degenerate directions by thresholding the eigenvalues for a filtering-based state estimation method.
Furthermore, the authors in~\cite{bai2021degeneration} argue that using only the minimum eigenvalue of the optimization Hessian is impractical due to the required tuning of thresholds. To mitigate this problem, the authors proposed simultaneously utilizing the minimum eigenvalue and the condition number.
%
Recently, the authors in~\cite{jin2024range} proposed a degeneracy metric called \textit{degenerate degree} by using the minimum eigenvalue in combination with the partial condition number of the optimization Hessian. 
Considering the challenges of setting a threshold for minimum eigenvalues, the authors in~\cite{switchslamlee} proposed to use the Chi-squared test to calculate a non-heuristic threshold of normalized eigenvalues. Later, the authors use these normalized eigenvalues to identify the degenerate directions for LiDAR as well as vision-based optimization problems.
In recent work, the authors of~\cite{ji_pointtodistribution} proposed a point-to-distribution ICP degeneracy detection method. Leveraging the probability distribution from their adaptive voxelization scheme, the authors show that their proposed metric is inherently correlated with the eigenvalues of the optimization Hessian.

\subsubsection{FINE-GRAINED APPROACHES}
Similarly,~\cite{tunnelLocalizability} proposed directly estimating an environment's localizability using the Hessian eigenspace by measuring the sensitivity of a point and surface-normal pair constraint with respect to the optimization states.
Using the formulation introduced in~\cite{tunnelLocalizability}, the authors in~\cite{zou2023lmapping} generated an information matrix from the LiDAR measurements to identify the degenerate directions of the optimization, which is later propagated to a factor-graph for accurate robot pose estimation.
Toward more fine-grained degeneracy handling, X-ICP~\cite{xicp} proposed a correspondence-based degeneracy detection method to constrain the optimization problem for point cloud registration. This method independently considered the individual observability contribution of each point-to-plane correspondence.
Following the formulation of X-ICP~\cite{xicp}, the authors in~\cite{chen2024relead} proposed to detect \ac{LiDAR} degeneracy using correspondence information in the registration step.\looseness-1

\subsection{DEGENERACY MITIGATION IN ROBOTICS}\label{subsubsection:degeneracyPreventionInRobotics}
For robot operation in more complex and unstructured environments, numerous approaches have been proposed to mitigate the effect of optimization ill-conditioning for LiDAR-based robot pose estimation methods.
These approaches can be categorized as either \textit{Passive} or \textit{Active}, based on how degeneracy mitigation is achieved. More specifically, the categorization is based on whether an approach changes the underlying optimization cost function or the immediate relative \ac{LiDAR} pose estimation of the registration, using the detected degeneracy information.
Technical details are provided in~\Cref{sec:methods}. 

\subsubsection{PASSIVE DEGENERACY MITIGATION}
\label{sec:related:passive}
Most passive degeneracy mitigation methods use a complementary sensor modality in a sensor fusion framework~\cite{nissov2024degradation, wen2024liver, bai2021degeneration} or rely on an external odometry source~\cite{compSLAM, pfreundschuh2023coin, dareSLAM, LION} to propagate the robot pose estimate in the case of LiDAR degeneracy. Uniquely, the authors of~\cite{lim2023adalio} employed an adaptive parameter setting strategy to mitigate the effects of \ac{LiDAR} degeneracy. In similar reasoning, the authors in~\cite{ferrari2024mad} focused on improving the correspondence search of the ICP algorithm to improve the quality of the ICP and, tangentially, the robustness to degeneracy.
Based on the analysis of the optimization variables, the authors in~\cite{gelfand2003} investigated the geometric stability of the point-to-plane \ac{ICP} by proposing a sampling-based method to extract the most informative scan points to improve the conditioning of the optimization process.
The authors in~\cite{petracek2024rms} proposed to minimize information redundancy in the input point cloud. To achieve this, the authors utilized an elaborate and degeneracy-aware sampling strategy and alleviated the adverse effects of the \ac{LiDAR} degeneracy.

\subsubsection{ACTIVE DEGENERACY MITIGATION}
\label{sec:related:active}
The work in~\cite{flory2009constrained} is among the first to perform active degeneracy mitigation by focusing on point cloud registration through additional constraints to reduce incorrect point cloud registration. Similarly,~\cite{constraintOptimization} concentrates on reducing the error originating from the linearization of the rotation variables by introducing non-linear equality constraints in the optimization. Recently, the authors in~\cite{solRemap} proposed \textit{Solution Remapping} to project the relative \ac{LiDAR} pose updates from the point cloud registration only along the well-constrained directions. In X-ICP~\cite{xicp}, the addition of hard constraints along the degenerate directions was proposed to reduce the robot pose estimate error. In~\cite{rankDeficientSLAM}, authors proposed using additional regularization terms in a non-linear least-squares problem to mitigate the effect of optimization degeneracy. Furthermore, to the best of the authors' knowledge,~\cite{rankDeficientSLAM} is the only existing work that compares different degeneracy mitigating methods.
Utilizing the constrained optimization from X-ICP~\cite{xicp}, the authors in~\cite{chen2024relead} proposed to use the detected degeneracy information in their error-state iterated extended Kalman filter to prevent constrain degenerate LiDAR measurement updates in degenerate directions.
However, the corresponding comparison is limited and does not generalize to the applications of \ac{LiDAR}-\ac{SLAM} and point cloud registration. Moreover, the method proposed in~\cite{rankDeficientSLAM} does not conduct field experiments and uses an RGBD camera instead of a 3D-LiDAR.
As discussed, a considerable body of work focuses on \ac{LiDAR} degeneracy. Despite this fact, the literature lacks a comprehensive analysis of how to effectively use the LiDAR degeneracy information. This work thoroughly analyzes and compares different methods in simulated and real-world settings to bridge this gap.

\subsubsection{SENSOR FUSION AND MULTI-MODAL DEGENERACY MITIGATION}\label{subsubsection:sensorfusion}

In addition to LiDAR-centric methods, other approaches have focused on utilizing complementary sensing modalities to alleviate the adverse effects of LiDAR degeneracy. COIN-LIO~\cite{pfreundschuh2023coin} proposed to complement the uninformative directions in the \ac{LiDAR} inertial framework with \ac{LiDAR} intensity image using visual feature detection techniques. Similarly, the authors in~\cite{jin2024range} complemented the ICP degenerate directions with constraints from an ultra-wideband positioning approach. Furthermore, the authors in~\cite{nissov2024degradation} used LiDAR and radar modalities in a factor-graph-based windowed smoother with explicit degeneracy information for each modality.
Recently, the authors in~\cite{switchslamlee} combined LiDAR, inertial, and visual information by actively switching between modalities in a pose graph optimization setting. The authors utilized separate detection of optimization degeneracy for LiDAR and camera modalities compared to more generic covariance estimation methods such as dynamic covariance scaling~\cite{dcs}, fallback graphs~\cite{nubertGraph} or global frame alignment~\cite{nubert2025holistic}.
In recent years, the DARPA Subterranean (SubT) challenge~\cite{slamOverview2} highlighted the importance of degeneracy-aware LiDAR-SLAM. Among the methods used in the challenge, CompSLAM~\cite{compSLAM} used the detected degeneracy to switch between sensor modalities. In contrast, DARE-SLAM~\cite{dareSLAM} used degeneracy detection to prevent the addition of LiDAR factors into the pose graph to prevent degradation. Another method, LION~\cite{LION}, utilized a measurement multiplexer to estimate the uncertainty of the LiDAR factors to robustly fuse with the IMU measurements in a sliding-window factor graph.
%
To minimize the dependency on a single sensing modality, the authors in \cite{superodometry} suggested running three parallel pose graphs: one for LiDAR inertial estimation, another for visual-inertial estimation, and a central IMU pose graph dedicated to sensor fusion. This approach, which depends on precise measurement uncertainties, highlights the effectiveness of factor-graph-based methods. Extending the sensor fusion framework~\cite{superodometry} to actively utilize degeneracy information, the authors in \cite{zhao2024superloc} proposed a framework to predict the observability of a LiDAR scan and incorporate it as a covariance (uncertainty) matrix.\looseness-1

%
%
%
Although these methods improve the robustness of the overall odometry or pose estimation of the system, they leave the original problem of LiDAR degeneracy untouched and rely on external sensor fusion for system-level robustness. This work focuses on investigating methods that improve the robustness and quality of a LiDAR-based pose estimation within the degenerate point cloud registration framework.

\begin{figure*}[t]
    \centering
    \includegraphics[width=1.0\linewidth]{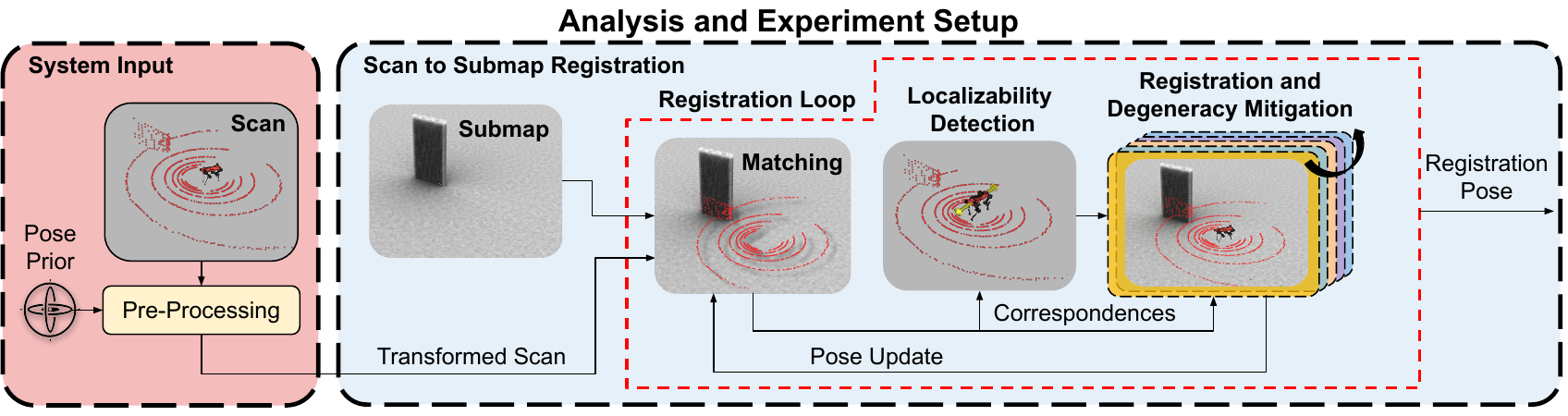}
    \caption{Overview of the analysis framework and experiment setup. The pose prior is used to transform and undistort the input point cloud, fed to the iterative registration optimization loop with the existing point cloud map. The localizability of the optimization at the current iteration is detected based on the method proposed in X-ICP~\cite{xicp} and fed to the analyzed degeneracy mitigation methods (explained in~\Cref{sec:methods}). The final registration pose within the registration loop is calculated iteratively as in \ac{ICP} fashion.}
    \label{fig:highLevelOverview}
\end{figure*}

\section{PROBLEM ANALYSIS AND PRELIMINARIES}\label{sec:problem_formulation}

This work investigates the advantages and disadvantages of different optimization degeneracy mitigation methods. 
Specifically, the problem of degenerate point cloud registration is considered under the scope of LiDAR-SLAM in geometrically challenging environments. This section presents the preliminaries on point cloud registration and degeneracy detection.\looseness-1

\subsection{POINT CLOUD REGISTRATION AND ICP}
\label{sec:problem_formulation:registration}
The point cloud registration problem aims to align two point clouds and obtain their relative rigid transformation as shown in \Cref{fig:highLevelOverview}. 
To achieve this, various techniques and metrics are proposed, as discussed in~\Cref{section:related_works}. 
In this work, the ICP algorithm with the point-to-plane~\cite{point_to_plane} cost function is utilized as it is widely adopted by the robotics community and
many point cloud registration frameworks~\cite{jelavic2022open3d, LION, compSLAM, ct_icp, libpointmatcher, fastlio2}.

Specifically, point cloud registration focuses on estimating the homogeneous transformation $\boldsymbol{T}_{\mathtt{M}\mathtt{L}} \in SE(3)$ between a \textit{source} and a \textit{reference} point cloud of sizes $N_p$ and $N_q$ respectively, given an initial guess $\boldsymbol{T}_{\mathtt{M}\mathtt{L},\text{init}}$. The \textit{source} point cloud $_{\mathtt{L}}\boldsymbol{P} \in \mathbb{R}^{3 \times N_p}$ is expressed in the LiDAR frame $\mathtt{L}$, while the \textit{reference} point cloud $_{\mathtt{M}}\boldsymbol{Q} \in  \mathbb{R}^{3 \times N_q}$ is expressed in the (sub)map frame $\mathtt{M}$.
The homogeneous transformation $\boldsymbol{T}_{\mathtt{M}\mathtt{L}} = \big[\boldsymbol{R}_{\mathtt{M}\mathtt{L}}\:|\: _\mathtt{M}\boldsymbol{t}_{\mathtt{M}\mathtt{L}}\big]$, consists of a rotation matrix $\boldsymbol{R} \in SO(3)$, and a translation vector $\boldsymbol{t} \in \mathbb{R}^3$, denoted as $\boldsymbol{t} = [t_x,\: t_y,\: t_z]^{\top}$. 
For the $N_p$ \textit{source} points $_{\mathtt{L}}\boldsymbol{p}_i \in \mathbb{R}^3$ in $_{\mathtt{L}}\boldsymbol{P}$, the closest reference point $_{\mathtt{M}}\boldsymbol{q}_i \in \mathbb{R}^3$ in $_{\mathtt{M}}\boldsymbol{Q}$, is found through a K-Nearest Neighbors correspondence search 
given initial transformation $\boldsymbol{T}_{\mathtt{M} \mathtt{L}, \text{init}}$ as shown in \Cref{fig:highLevelOverview} matching block. This transformation is often provided as an initial guess and applied to all points in $_{\mathtt{L}}\boldsymbol{P}$ to obtain $_{\mathtt{M},\text{init}}\boldsymbol{P}$ to improve the matching process and the optimization convergence characteristics~\cite{brossard} through the iterative process of \ac{ICP}.
In the remainder of the section, all variables are expressed in the (sub)map frame $\mathtt{M}$, and for the sake of brevity, the notation is simplified to omit the frame.
%
The result of the correspondence search can be summarized as a one-to-many matching $\{\boldsymbol{p}_i, \ \boldsymbol{Q} (\boldsymbol{p}_i)\}$ for $i \in \{1, \dots, N\}$, 
where $\boldsymbol{p}_i$ and $\boldsymbol{{q}}_i \coloneqq ~ \boldsymbol{Q} (\boldsymbol{p}_i)$ are the $N \leq N_p$ matched point pairs from \textit{source} and \textit{reference} point clouds, respectively. 
Moreover, $\boldsymbol{n}_i \in \mathbb{R}^3, \: \| {\boldsymbol{n}} \| =1$ is the surface normal of reference point $\boldsymbol{q}_j$, as needed for the point-to-plane~\cite{point_to_plane} \ac{ICP} cost function. 
The number of matched points $N$ indicates the size of the optimization problem. 
The cost function of the point-to-plane \ac{ICP} minimization problem is defined as\looseness-1
\begin{align}
\label{eq:point_to_plane_cost}
    \mathop{\text{min}}_{\boldsymbol{R},\: \boldsymbol{t}} 
    \sum_{i=1}^N \Big|\Big| \big((\boldsymbol{R} \boldsymbol{p}_i + \boldsymbol{t}) - \boldsymbol{ {q}}_i\big) \cdot {\boldsymbol{ {n}}}_i \Big|\Big|_2.
\end{align}
The problem shown in \Cref{eq:point_to_plane_cost} can be reformulated as a quadratic cost minimization problem following the derivation shown in \cite{ICPderivation} after introducing the scalar triple product and rotation matrix linearization
\begin{align}
\label{eq:huge_cost}
    \begin{split}
        \mathop{\text{min}}_{\boldsymbol{x}\in \mathbb{R}^{6}} 
        \boldsymbol{x}^T
        \underbrace{
        \Bigg(
          \sum_{i=1}^N
              \begin{bmatrix} \left(\boldsymbol{p}_i \times \boldsymbol{ {n}}_i\right) \\  \boldsymbol{ {n}}_i \end{bmatrix}
              \left[ (\boldsymbol{p}_i \times \boldsymbol{ {n}}_i)^T \  \boldsymbol{ {n}}_i^T \right]
        \Bigg)
        }_{\boldsymbol{A}}
        \boldsymbol{x}
        \\ -
        2\boldsymbol{x}^T
        \underbrace{
        \Bigg(
          \sum_{i=1}^N 
              \begin{bmatrix} \left(\boldsymbol{p}_i \times \boldsymbol{ {n}}_i\right) \\  \boldsymbol{ {n}}_i \end{bmatrix}
              \boldsymbol{ {n}}_i^T
              (\boldsymbol{ {q}}_i-\boldsymbol{p}_i)
        \Bigg)
        }_{\boldsymbol{b}} + \; \text{Constant}.
    \end{split}
\end{align}
Here, $\boldsymbol{x} = [\boldsymbol{r}^\top, \ \boldsymbol{t}^\top]^\top \: \in  \mathbb{R}^{6}$ are the optimization variables, where the translational part is $\boldsymbol{t} \in \mathbb{R}^3$. Moreover, the rotation vector $\boldsymbol{r} \in \mathbb{R}^{3}$ where $\boldsymbol{R} \approx \boldsymbol{r}^\wedge + \boldsymbol{I}$.   
The $^\wedge$ "hat operator"  transforms a vector to its skew-symmetric matrix representation, $\boldsymbol{r}^\wedge \in \mathfrak{so}(3)$. Here, $\mathfrak{so}(3)$ is the Lie algebra of $SO(3)$ group, which consists of skew-symmetric matrices of size $3 \times 3$.\looseness-1

Substituting $\boldsymbol{A}$ and $\boldsymbol{b}$ into~\Cref{eq:huge_cost} results in the regular linear least-squares optimization formulation
\begin{align}
    E(\boldsymbol{x}) = \mathop{\text{min}}_{\boldsymbol{x}\in \mathbb{R}^{6}} 
    \Big|\Big|\boldsymbol{A}
    \boldsymbol{x}- \boldsymbol{b}\Big|\Big|_2,
\label{eq:final_cost}
\end{align} 
with $E(\boldsymbol{x})$ indicating the residual error of the optimization and minimization of this cost function results in the Frobenius norm (2-norm) solution. 
Here, $\boldsymbol{A} \in \mathbb{R}^{6\times 6}$ denotes the Hessian of the optimization problem, and $\boldsymbol{b} \in \mathbb{R}^6$ contains the constraints imposed by the registration. 
%
The optimal translation $\boldsymbol{t}$ and rotation $\boldsymbol{r}$ vectors can be calculated at each linearization point.
Due to errors in linearization and correspondence matching, the matching and minimization operations of \ac{ICP} are repeated iteratively until convergence.

%
The minimization problem~\Cref{eq:point_to_plane_cost} can be solved with nonlinear solvers, such as Gauss-Newton or \ac{LM} optimization~\cite{ceres}, or using closed-form solvers, such as \ac{SVD}~\cite{svd} or \ac{LU}-decomposition to solve~\Cref{eq:final_cost}. This operation corresponds to the degeneracy mitigation block shown in~\Cref{fig:highLevelOverview}.
Except for~\Cref{sec:methods:active:nlreg} and~\Cref{sec:methods:passive:global}, this work uses \ac{SVD} to solve~\Cref{eq:final_cost} for the reasons introduced in the following.

The minimum of~\Cref{eq:final_cost} is found when the cost function gradient with respect to $\boldsymbol{x}$ is zero in all directions. Following the previous definitions,
\begin{align}
    \begin{split}
    \frac{\partial E(\boldsymbol{x})}{\partial \boldsymbol{x}} &= \frac{\partial}{\partial \boldsymbol{x}} \bigl[\boldsymbol{b}^{\top}\boldsymbol{b} - \boldsymbol{b}^{\top}\boldsymbol{A}\boldsymbol{x} - \boldsymbol{x}^{\top}\boldsymbol{A}^{\top}\boldsymbol{b} 
    + \boldsymbol{x}^{\top}\boldsymbol{A}^{\top}\boldsymbol{A}\boldsymbol{x}\bigr], \\
    &= -2\boldsymbol{A}^{\top}\boldsymbol{b} + 2\boldsymbol{A}^{\top}\boldsymbol{A}\boldsymbol{x} = 0,
    \end{split}
\label{eq:partial_derivative}
\end{align}
resulting in the normal equation after re-arranging the terms
\begin{align} 
    \left(\boldsymbol{A}^T \boldsymbol{A}\right)\boldsymbol{x}=\boldsymbol{A}^T \boldsymbol{b}.
\label{eq:normal_equation}
\end{align}
The solution of the least-squares problem with~\Cref{eq:normal_equation} can be interpreted geometrically as the projection of $\boldsymbol{b}$ onto the span of matrix $\boldsymbol{A}$, as shown in~\Cref{fig:geometric}. The residual of the optimization can be seen as the orthogonal distance of $\boldsymbol{b}$ to the span of $\boldsymbol{A}$.

\subsection{SENSITIVITY OF THE LEAST-SQUARES PROBLEM}
\label{sec:formulation:sensitivity}

This section contextualizes the previously introduced methods with the sensitivity and conditioning of the least-squares minimization problem. The sensitivity of a problem to small changes in the input is referred to as the conditioning of the problem. A problem is called ill-conditioned if small changes in the input lead to big changes in the output. Otherwise, the problem is called well-conditioned~\cite{og_book}.

Following the derivation~\cite{matrix_computations, og_book}, the Euclidian condition number of matrix $\boldsymbol{A}$ can be defined as 
\begin{align}
    \text{cond}(\boldsymbol{A}) = \frac{\sigma_{max}}{\sigma_{min}}= \|\boldsymbol{A}\|_2 \|\boldsymbol{A}^{+}\|_2.
    \label{eq:cond}
\end{align}
Here, $\boldsymbol{A}^{+}$ is the pseduo-inverse of $\boldsymbol{A}$. The pseudo-inverse can be calculated using the SVD of $\boldsymbol{A}$ as detailed in~\Cref{sec:formulation:solving_the_opt} and \Cref{eq:svd_of_hessian}. Furthermore, $\sigma_{max}$ and $\sigma_{min}$ refer to the maximum and minimum eigenvalue of $\boldsymbol{A}$, respectively. Similarly, eigenvalues can be retrieved using SVD as detailed in~\Cref{eq:svd_of_hessian}.

\begin{figure}[h!]
\centering
\includegraphics[width=1\linewidth]{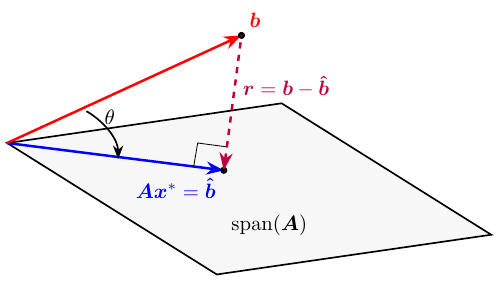}
\caption{
Geometric interpretation of the least-squares minimization problem. Here, $\theta$ is the angle between the approximated constraints $\boldsymbol{\hat{b}}$ and the original constraints $\boldsymbol{b}$. The minimization of~\Cref{eq:final_cost} minimizes residual the $\boldsymbol{r}$.
}
\label{fig:geometric}
\end{figure}

Previously, the authors in~\cite{solRemap} provided in-depth technical details of their degeneracy detection factor with the sensitivity analysis of a square linear system $\boldsymbol{A}\boldsymbol{x} = \boldsymbol{b}$. However, in reality, the ICP formulation does not use this equality and instead solves a least-squares minimization problem as described in~\Cref{sec:problem_formulation} and~\Cref{eq:normal_equation}. The conditioning of a least-squares problem $\boldsymbol{A}\boldsymbol{x} \approx \boldsymbol{b}$ depends not only on the matrix $\boldsymbol{A}$ but also on the right-hand-side vector $\boldsymbol{b}$, and thus $\text{cond}(\boldsymbol{A})$ alone does not suffice to characterize sensitivity.
As seen in \Cref{fig:geometric}, a constraint vector $\boldsymbol{b}$ near $\text{span}(\boldsymbol{A})$ would result in a smaller residual, thereby, perturbations on $\boldsymbol{b}$ would induce lesser variations in $\boldsymbol{x}$. It can be inferred that as $\theta \rightarrow \sfrac{\pi}{2}$ the least-squares solution becomes singular. This behavior can be characterized with $\cos{\theta} = \sfrac{\|\boldsymbol{A}\boldsymbol{x}\|}{\|\boldsymbol{b}\|}$.
%
%
\subsubsection[PERTURBATION ON b]{PERTURBATION ON $\boldsymbol{b}$} Here, an analysis of the sensitivity of $\boldsymbol{x}$ to a change in the right-hand-side vector $\boldsymbol{b}$ is provided. For brevity, the 2-norm notation is shown as $\|\cdot\|$ in the following. Given a perturbation $\delta\boldsymbol{b}$ and a resulting change in $\boldsymbol{x}$, $\delta\boldsymbol{x}$ the normal equation~\Cref{eq:normal_equation} can be rewritten as
\begin{align}
    \begin{split}
    \boldsymbol{A}^T \boldsymbol{A}\left(\boldsymbol{x} + \Delta\boldsymbol{x}\right)&=\boldsymbol{A}^T \left(\boldsymbol{b} + \Delta \boldsymbol{b} \right), \\
    \cancel{\boldsymbol{A}^T \boldsymbol{A}\boldsymbol{x}} + \boldsymbol{A}^T \boldsymbol{A}\Delta\boldsymbol{x}&=\cancel{\boldsymbol{A}^T\boldsymbol{b}} + \boldsymbol{A}^T \Delta \boldsymbol{b}.
    \end{split}
\label{eq:rewritten_normal}
\end{align}

Canceling the original components of the normal equation (\Cref{eq:normal_equation}) leaves
\begin{align} 
\begin{split}
    &\boldsymbol{A}^T \boldsymbol{A}\Delta\boldsymbol{x}=\boldsymbol{A}^T \Delta \boldsymbol{b}, \\ 
    &\Delta\boldsymbol{x} = \underbrace{(\boldsymbol{A}^T \boldsymbol{A} )^{-1}\boldsymbol{A}^T}_{\boldsymbol{A}^{+}}\Delta \boldsymbol{b}, \\
    &\Delta\boldsymbol{x} = \boldsymbol{A}^{+}\Delta \boldsymbol{b}.
\end{split}
\label{eq:substitute}
\end{align}
As this analysis investigates the amplitude of the change, the 2-norm of both sides is taken, which results in
\begin{align} 
    \|\Delta\boldsymbol{x}\| = \|\boldsymbol{A}^{+}\|\|\Delta \boldsymbol{b}\|.
\end{align}
To retrieve the relative amplitude of the change, both sides are normalized by $\|\boldsymbol{x}\|$. This results in an inequality bound instead of equality constraint as a consequence of the Cauchy–Schwarz inequality, $\text{cond}(\boldsymbol{A}) = \|\boldsymbol{A}\| \|\boldsymbol{A}^{+}\| \geq 1$ and  $\|\boldsymbol{A}^{+}\| \approx \frac{1}{\sigma_{min}}$. The inequality results in
\begin{align}
\begin{split}
    \frac{\|\Delta\boldsymbol{x}\|}{\|\boldsymbol{x}\|} &\leq \|\boldsymbol{A}^{+}\|\frac{\|\Delta \boldsymbol{b}\|}{\|\boldsymbol{x}\|}, \\
    \frac{\|\Delta\boldsymbol{x}\|}{\|\boldsymbol{x}\|} &\leq \frac{\text{cond}(\boldsymbol{A})}{\|\boldsymbol{A}\|} \frac{\|\boldsymbol{b}\|}{\|\boldsymbol{b}\|}\frac{\|\Delta \boldsymbol{b}\|}{\|\boldsymbol{x}\|}, \\
    \frac{\|\Delta\boldsymbol{x}\|}{\|\boldsymbol{x}\|} &\leq \text{cond}(\boldsymbol{A})\underbrace{\frac{\|\boldsymbol{b}\|}{\|\boldsymbol{A}\boldsymbol{x}\|}}_{(\cos{\theta})^{-1}} \frac{\|\Delta \boldsymbol{b}\|}{\|\boldsymbol{b}\|}.
    \end{split}
    \label{eq:perturb_b}
\end{align}
\Cref{eq:perturb_b} shows that the sensitivity of $\Delta\boldsymbol{x}$ is affected by $\text{cond}(\boldsymbol{A})$ and how well $\boldsymbol{b}$ is approximated with $\boldsymbol{A}\boldsymbol{x}$. Interestingly, when $\cos\theta \approx 1$ the effective condition number is approximately $\text{cond}(\boldsymbol{A})$ however, if $\cos\theta \approx 0$ the effective condition number would be much larger than $\text{cond}(\boldsymbol{A})$.
This finding implies that, despite perfect conditioning, a least-squares optimization could be ill-defined if $\boldsymbol{b}$ has noise and the residual is large.

\subsubsection[PERTURBATION ON A]{PERTURBATION ON $\boldsymbol{A}$} To fully understand the sensitivity, the perturbation of $\boldsymbol{A}$ should also be considered. Similarly to the perturbation of $\boldsymbol{b}$, the normal equation is performed after adding a perturbation of $\Delta\boldsymbol{A}$ to $\boldsymbol{A}$.
\begin{align} 
    \left({\boldsymbol{A} + \Delta\boldsymbol{A}}\right)^T\left({\boldsymbol{A} + \Delta\boldsymbol{A}}\right)\left(\boldsymbol{x} + \Delta\boldsymbol{x}\right)=\left({\boldsymbol{A} + \Delta\boldsymbol{A}}\right)^T\boldsymbol{b}.
\label{eq:rewritten_normal_perturb_A}
\end{align}
Expanding both sides of the above equation results to
\begin{align}
    \begin{split}
    &\cancel{\boldsymbol{A}^T \boldsymbol{A} \boldsymbol{x}}+ \boldsymbol{A}^T \boldsymbol{A} \Delta\boldsymbol{x} +  \boldsymbol{A}^T (\Delta\boldsymbol{A})\boldsymbol{x} + (\Delta\boldsymbol{A})^T \boldsymbol{A}\boldsymbol{x} \ +  \\ 
    &\underbrace{\boldsymbol{A}^T \Delta\boldsymbol{A} \Delta\boldsymbol{x} + (\Delta\boldsymbol{A})^T \boldsymbol{A} \Delta\boldsymbol{x} + (\Delta\boldsymbol{A})^T \Delta\boldsymbol{A}\boldsymbol{x} }_{\approx 0} + \\ & \underbrace{(\Delta\boldsymbol{A})^T \Delta\boldsymbol{A}\Delta\boldsymbol{x}}_{\approx 0}, \\ 
    &= \cancel{\boldsymbol{A}^{T}\boldsymbol{b}} + (\Delta\boldsymbol{A})^T\boldsymbol{b}.
    \end{split}
\end{align}
Higher-order terms are neglected to simplify the calculations, and the components of the original normal equation (\Cref{eq:normal_equation}) are canceled out. The above equation is rewritten after the simplifications as follows:
\begin{align*}
    \begin{split}
        \boldsymbol{A}^T \boldsymbol{A} \Delta \boldsymbol{x} +  \boldsymbol{A}^T (\Delta\boldsymbol{A})\boldsymbol{x} + (\Delta\boldsymbol{A})^T \boldsymbol{A}\boldsymbol{x} &\approx (\Delta\boldsymbol{A})^T\boldsymbol{b}, \\
         \boldsymbol{A}^T \boldsymbol{A} \Delta \boldsymbol{x} + \boldsymbol{A}^T (\Delta\boldsymbol{A})\boldsymbol{x}&\approx (\Delta\boldsymbol{A})^T \underbrace{\left(\boldsymbol{b} - \boldsymbol{A}\boldsymbol{x}\right)}_{\boldsymbol{r}},
    \end{split}
\end{align*}
\vspace{-1em}
\begin{align}
    \begin{split}
          \boldsymbol{A}^T \boldsymbol{A} \Delta \boldsymbol{x} &\approx (\Delta\boldsymbol{A})^T \boldsymbol{r} - \boldsymbol{A}^T (\Delta\boldsymbol{A})\boldsymbol{x}, \\
          \Delta \boldsymbol{x} &\approx(\boldsymbol{A}^T \boldsymbol{A})^{-1}(\Delta\boldsymbol{A})^T \boldsymbol{r} - \underbrace{(\boldsymbol{A}^T \boldsymbol{A})^{-1}\boldsymbol{A}^T}_{\boldsymbol{A}^+}(\Delta\boldsymbol{A})\boldsymbol{x}, \\
          \Delta \boldsymbol{x} &\approx (\boldsymbol{A}^T \boldsymbol{A})^{-1}(\Delta\boldsymbol{A})^T \boldsymbol{r} + \boldsymbol{A}^{+}(\Delta\boldsymbol{A})\boldsymbol{x}.
    \end{split}
\end{align}
In order to complete the sensitivity analysis, the relative changes are required. Hence, applying 2-norm and re-arranging the variables as
\begin{align}
\begin{split}
    \frac{\|\Delta \boldsymbol{x}\|}{\|\boldsymbol{x}\|} &\leq \frac{\text{cond}(\boldsymbol{A})^{2}}{\|\boldsymbol{A}\|^{2}}\|\Delta\boldsymbol{A}\| \frac{\|\boldsymbol{r}\|}{\|\boldsymbol{x}\|} + \frac{\text{cond}(\boldsymbol{A})}{\|\boldsymbol{A}\|}\|\Delta\boldsymbol{A}\|, \\
    \frac{\|\Delta \boldsymbol{x}\|}{\|\boldsymbol{x}\|} &\leq \text{cond}(\boldsymbol{A})^{2}\frac{\|\boldsymbol{r}\|}{\|\boldsymbol{A}\boldsymbol{x}\|}\frac{\|(\Delta\boldsymbol{A})\|}{\|\boldsymbol{A}\|} + \text{cond}(\boldsymbol{A})\frac{\|\Delta\boldsymbol{A}\|}{\|\boldsymbol{A}\|},\\
    \frac{\|\Delta \boldsymbol{x}\|}{\|\boldsymbol{x}\|} &\leq \left( \text{cond}(\boldsymbol{A})^{2}\underbrace{\frac{\|\boldsymbol{r}\|}{\|\boldsymbol{A}\boldsymbol{x}\|}}_{\tan\theta} + \text{cond}(\boldsymbol{A}) \right)\frac{\|\Delta\boldsymbol{A}\|}{\|\boldsymbol{A}\|}.
    \end{split}
    \label{eq:perturb_A}
\end{align}
Here, the fact that $\|(\Delta\boldsymbol{A})^{T}\| =  \|(\Delta\boldsymbol{A})\|$ and the definition of condition number (\Cref{eq:cond}) is used to bring the equation into an interpretable format. With the conclusion of \Cref{eq:perturb_A}, it becomes clear how the sensitivity of the least-squares problem is modeled against perturbations to $\boldsymbol{A}$. It is easy to see that two terms affect the propagation of the perturbation to $\boldsymbol{x}$, namely the condition number of $\boldsymbol{A}$ and a second term proportional to the residual $\boldsymbol{r}$, scaled with \textbf{square} of the condition number of $\boldsymbol{A}$. This second term is particularly important as any remaining residual $\boldsymbol{r}$ will be scaled by $\text{cond}(\boldsymbol{A})^2$, which is important for ill-conditioned problems. As a result of the above explanations, the effective conditioning of the least-squares optimization problem differs from the conditioning of $\boldsymbol{A}$. This finding shows the importance of mitigating the optimization degeneracy at the time of or before optimization so that the residual term does not interfere with the quality of the minimization.

\subsection{SOLVING THE OPTIMIZATION}
\label{sec:formulation:solving_the_opt}
In \Cref{sec:problem_formulation:registration}, the sensitivity of the normal equations (~\Cref{eq:normal_equation}) are analyzed for perturbations on $\boldsymbol{A}$ and $\boldsymbol{b}$; however, the retrieval of the optimal optimization variables is not discussed. Thus, this section introduces how the optimized variables that minimize \Cref{eq:final_cost} can be calculated with SVD.
\Cref{eq:normal_equation} can be solved with efficient factorization methods such as the Cholesky factorization, given that $ \boldsymbol{A}^\top \boldsymbol{A}$ is not ill-posed (near-singular). However, since this work focuses on \ac{LiDAR} degeneracy and, subsequently, an ill-conditioned optimization, using Cholesky factorization is not advantageous. 
Instead, \ac{SVD}, a factorization method that produces two orthonormal bases next to the diagonalized original matrix, is used to solve this equation, especially when the optimization problem is ill-conditioned, as SVD solution exists for all (non-square) matrices. 
For a matrix $\boldsymbol{A}$ of size $m \times n$, the decomposition is defined as
\begin{align}
    \boldsymbol{A} = \boldsymbol{U}\boldsymbol{\Sigma} \boldsymbol{V}^{\top}, \quad {\boldsymbol{A}}^{+} = \boldsymbol{V} \boldsymbol{\Sigma}^{-1} \boldsymbol{U}^{\top},
    \label{eq:svd_of_hessian}
\end{align}
with $\boldsymbol{U}$ a $m \times m$ rotation matrix, $\boldsymbol{\Sigma}$ a diagonal $m \times n$ scaling and projection matrix that contains the singular values of the matrix $\boldsymbol{A}^{\top}\boldsymbol{A}$, and lastly, $\boldsymbol{V}$ the second rotation matrix, with size $n \times n$. Moreover, $\boldsymbol{A}^{+}$ is the pseudo-inverse of $\boldsymbol{A}$ and in the full-rank case of $\boldsymbol{A}$, $\boldsymbol{A}^{+}\approx \boldsymbol{A}^{-1}$.
For symmetric and positive (semi)-definite   $\boldsymbol{A}$, \ac{SVD} becomes identical to eigenvalue-decomposition $\boldsymbol{A}=\boldsymbol{V}\boldsymbol{\Sigma} \boldsymbol{V}^\top$, and subsequently 
\begin{align}
    \begin{split}
    \boldsymbol{A}^{\top} \boldsymbol{A} & ={\left(\boldsymbol{V} \boldsymbol{\Sigma} \boldsymbol{V}^{\top} \right)}^{\top}(\boldsymbol{V} \boldsymbol{\Sigma} \boldsymbol{V}^{\top})
    =\boldsymbol{V} \boldsymbol{\Sigma}^2 \boldsymbol{V}^{\top},
    \end{split}
\end{align}
where $\boldsymbol{\Sigma}^2$ denotes a square diagonal matrix that has the squared singular values of $\boldsymbol{A}$ as entries. 
The right singular vectors $\boldsymbol{V}$ of $\boldsymbol{A}$ are equivalent to the eigenvectors of $\boldsymbol{A}^{\top} \boldsymbol{A}$, while the singular values of $\boldsymbol{A}$ are the square root of the eigenvalues of $\boldsymbol{A}^{\top} \boldsymbol{A}$.

After the \ac{SVD} decomposition, the resulting orthonormal bases and the diagonal scaling matrix allow for the pseudo-inverse calculation
\begin{equation}
    \left( \boldsymbol{A}^{\top} \boldsymbol{A}\right)^{+}  = \boldsymbol{V} {\boldsymbol{\Sigma}^{-2}} \boldsymbol{V}^{\top}.
    \label{eq:pseudo_inversion}
\end{equation}
Applying~\Cref{eq:pseudo_inversion} and~\Cref{eq:svd_of_hessian} to~\Cref{eq:normal_equation} results in
\begin{align}
    \begin{split}
    \boldsymbol{x}^{*}&= \left( \boldsymbol{A}^{\top} \boldsymbol{A}\right)^{-1}\boldsymbol{A}^T \boldsymbol{b}, \\
    \boldsymbol{x}^{*} &= \left(\boldsymbol{V} {\boldsymbol{\Sigma}^{-2}} \boldsymbol{V}^{\top}\right)  \boldsymbol{A}^T \boldsymbol{b}, \\
    \boldsymbol{x}^{*}&=  \left(\boldsymbol{V} {\boldsymbol{\Sigma}^{-2}} \boldsymbol{V}^{\top}\right)  \left( \boldsymbol{V} \boldsymbol{\Sigma} \boldsymbol{U}^{\top}  \right) \boldsymbol{b}, \\
    \boldsymbol{x}^{*}&= \: \boldsymbol{V}\boldsymbol{\Sigma}^{-1} \boldsymbol{U}^{\top}\boldsymbol{b}.
    \label{eq:svd_solving}
    \end{split}
\end{align}
As \ac{SVD} is based on the pseudo-inversion described in~\Cref{eq:pseudo_inversion}, the optimal $\boldsymbol{x}^{*}$ can be found regardless of the conditioning of $\boldsymbol{A}$. However, this optimal $\boldsymbol{x}^{*}$ would not be optimal along the ill-conditioned directions of the optimization as these directions will remain unobservable.

On a similar note, SVD is particularly effective against ill-conditioned or nearly rank-deficient problems, as the decomposition allows for omitted near-zero singular values. As a result, the least-squares solution is less sensitive to perturbations in the data~\cite{og_book}. This property is leveraged by one of the methods, Truncated SVD, detailed in~\Cref{sec:methods:active:tsvd}.\looseness-1

\subsection{LiDAR DEGENERACY DETECTION}\label{sec:localizability}
When robots are deployed in geometrically featureless and challenging environments, point cloud registration can become ill-conditioned. Hence, a method to detect the degenerate directions of the optimization is required.
As this work focuses on degeneracy mitigation, all works discussed here use the degeneracy detection method described in X-ICP~\cite{xicp} to ensure a fair comparison (for all methods that require localizability information). This localizability detection process is shown as the localizability detection block in~\Cref{fig:highLevelOverview}. {\color{black}For more details on LiDAR-based degeneracy detection, the readers are referred to  X-ICP(\cite{xicp}, Section V).}\looseness-1

%
Following the derivation from~\cite{xicp}, the optimization Hessian is divided into sub-matrices to identify the eigenvectors corresponding to the rotational and translational parts $\boldsymbol{r}$ and $\boldsymbol{t}$ of the optimization variables $\boldsymbol{x}$
\begin{align}
    {\boldsymbol{A}} = 
    \begin{bmatrix}
        {\boldsymbol{A}}_{rr}  & {\boldsymbol{A}}_{rt}      \\
        {\boldsymbol{A}}_{tr}  & {\boldsymbol{A}}_{tt}      \\
    \end{bmatrix}_{6\times6}. 
\end{align}
The factorized Hessians ${\boldsymbol{A}}_{rr} \in \mathbb{R}^{3\times 3}$ and ${\boldsymbol{A}}_{tt} \in \mathbb{R}^{3\times 3}$ correspond to rotational and translational information, respectively.
Furthermore, ${\boldsymbol{A}}_{rr}$ and ${\boldsymbol{A}}_{tt}$ are utilized for degeneracy detection as it is not trivial to treat $\boldsymbol{t}$ and $\boldsymbol{r}$ jointly due to differences in scale and physical meaning.
%
%
For the corresponding sub-matrices, the eigenvalue-decomposition can be written as
\begin{align}
    {\boldsymbol{A}}_{tt} = \boldsymbol{V}_{t}\Sigma_t \boldsymbol{V}_{t}^{\top}, \quad   {\boldsymbol{A}}_{rr} = \boldsymbol{V}_{r}\Sigma_r\boldsymbol{V}_{r}^{\top}.
\end{align}
The matrices $\boldsymbol{V}_t \in \mathbb{R}^{3\times3}$ and $\boldsymbol{V}_r \in \mathbb{R}^{3\times3}$ are the eigenvectors in matrix form,  
and $\Sigma_t \in \{\text{diag}(\boldsymbol{\boldsymbol{\sigma}_{tt}}): \boldsymbol{\sigma}_{tt} \in {\mathbb{R}^3} {\geq \boldsymbol{0}} \}$ and $\Sigma_r \in \{\text{diag}(\boldsymbol{\sigma}_{rr}): \boldsymbol{\sigma}_{rr} \in {\mathbb{R}^3} {\geq \boldsymbol{0}} \}$ are the scaling matrices for $\boldsymbol{A}_{tt}$ and $\boldsymbol{A}_{rr}$, respectively. Here, $\boldsymbol{\sigma}_{tt}$ and $\boldsymbol{\sigma}_{rr}$ are the eigenvalues of $\boldsymbol{A}_{tt}$ and $\boldsymbol{A}_{rr}$. 
For the remainder of the work, the null space approximation of $\boldsymbol{V}_t$ and $\boldsymbol{V}_r$ is assumed to be known through this analysis in combination with their associated eigenvalues from $\Sigma_t$ and $\Sigma_r$. 
If an eigenvector, $\boldsymbol{v}_j \in \{\boldsymbol{V}_t,\: \boldsymbol{V}_r\}$, is within the null space of $\boldsymbol{V}_t$ or $\boldsymbol{V}_r$, it is denoted as \textit{degenerate} (non-localizable), otherwise \textit{non-degenerate} (localizable).
In addition, the mapping $\boldsymbol{V} \Leftrightarrow \{ \boldsymbol{V}_r, \: \boldsymbol{V}_t \}$ is taken into account since a subset of methods described later in~\Cref{sec:methods} expect degenerate eigenvectors in $ \mathbb{R}^{3\times3}$, while others expect them in $\mathbb{R}^{6\times6}$. Moreover, the separate detection of translational and rotational degeneracy allows careful consideration of the scale differences of the entries in the Hessian $\boldsymbol{A}$. As a result of this consideration, the cross-terms $\boldsymbol{A}_{tr}$ and $\boldsymbol{A}_{rt}$ are omitted. However, since all methods that require degeneracy detection information have access to the same degeneracy information, the effect of this omission is alleviated.
\begin{figure*}[t]
    \centering
    \includegraphics[width=0.9\linewidth]{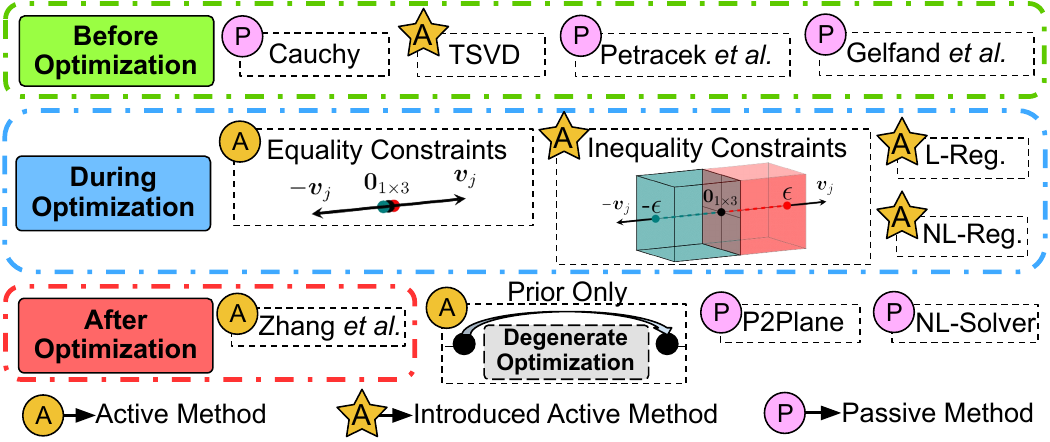}
    \caption{
    An overview of all the investigated methods, highlighting at which stage these methods affect the point cloud registration process. The letters \textbf{P} and \textbf{A} mark passive and active degeneracy mitigation methods, respectively. The $\bigstar$ symbol indicates that the method is investigated in the context of degenerate point cloud registration for the first time in this work. Zhang \textit{et al.} refers to work in~\cite{solRemap}, Petracek \textit{et al.} to~\cite{petracek2024rms} and Gelfand \textit{et al.} to~\cite{gelfand2003}.
    }
\label{fig:methods}
\end{figure*}

\section{METHODOLOGY AND IMPLEMENTATION}\label{sec:methods}
In this section, the theoretical and implementation details of the methods compared in Section~\ref{section:Experiments} are laid out.
Motivated by the discussion in~\Cref{section:introduction} and \Cref{section:related_works}, the methods are classified into two categories, namely, \textit{i) Active degeneracy mitigation methods} and \textit{ii) Passive degeneracy mitigation methods}, as shown in~\Cref{fig:methods}. The main difference between these two classes of methods is the utilization of the available degeneracy detection. Passive methods do not utilize this available information, while active methods utilize this information in different ways to mitigate the adverse effects of optimization degeneracy.

\subsection{ACTIVE DEGENERACY MITIGATION}
\label{sec:methods:active}
The active degeneracy mitigation class contains the methods that utilize the detected degeneracy information to alter the optimization cost function or the optimization variables. 
Often, these methods require accurate information on degeneracy detection to perform optimally. In addition, these methods perform additional computations to mitigate the effects of degeneracy or circumvent degeneracy by avoiding registration. 
In addition, active methods can be separated into further classes based on when the method affects the ICP,  as shown in~\Cref{fig:methods}.
%
%

\subsubsection{EQUALITY CONSTRAINTS}
Equality constraints (Eq. Con.), as recently discussed in X-ICP~\cite{xicp}, utilize constrained optimization techniques to add additional equality constraints along the degenerate directions, thus preventing optimization of the solution along these directions during iterative refinement. 
%
In the formulation of the equality constraints, the degenerate eigenvectors $ \boldsymbol{v}_j \in \{\boldsymbol{V}_t,\boldsymbol{V}_r\}$ of size $3\times1$ each for translation and rotation will be used as constraint directions. Here, where $ \boldsymbol{V_{{r},{t}}} \in \mathbb{R}^{3 \times 3}$ are the eigenvector matrices for rotational and translational parts, respectively.
Given the eigenvectors, the constraint formulation is
\begin{align}
   \begin{split}
       \boldsymbol{v}_j\cdot (\boldsymbol{t}-\boldsymbol{t}_{0})=0,\quad& \text{if } \boldsymbol{v}_j \in \boldsymbol{V}_t,\\
       \boldsymbol{v}_j\cdot (\boldsymbol{r}-\boldsymbol{r}_{0})=0,\quad& \text{if } \boldsymbol{v}_j \in \boldsymbol{V}_r,
   \end{split}
    \label{eq:constraint_form}
\end{align}  
where $\boldsymbol{t}_0$ and $\boldsymbol{r}_0$ are the translation and rotation constraint values, respectively. 
In this work, \ac{Eq. Con.} constraint values are set to $\boldsymbol{t}_0 = \boldsymbol{r}_0 = \boldsymbol{0}$ to prevent motion updates along the degenerate directions. 
In practice, these constraint values can be set to any pre-determined value to \textit{steer} the optimization to the correct minima to improve convergence. The determination of this value is an active research topic and is not covered in this work. As a consequence, the values of the constraints are set to $0$, which means that optimization is limited to having updates on the linear combination of optimization variables $\boldsymbol{x}$ in the degenerate directions  $ \boldsymbol{v}_j \in \{\boldsymbol{V}_t,\boldsymbol{V}_r\}$.

The ability to control the direction and amplitude of the motion updates is a key property of \ac{Eq. Con.}, directly benefiting the effectiveness against degenerate optimization problems. The underlying effect of these additional constraints is previously highlighted in~\Cref{sec:formulation:sensitivity}.

To utilize these constraints in the unconstrained cost function shown in~\Cref{eq:final_cost}, the equations are augmented with zero vectors to be applied to a $6$-\ac{DoF} optimization problem.
The resulting augmented constraints are
\begin{align}
    \begin{split}
        \displaystyle\big[\boldsymbol{0}_{1\times3},\boldsymbol{v}_j\big]\cdot \boldsymbol{x}=\boldsymbol{0},\quad& \text{if } \displaystyle\boldsymbol{v}_j \in \boldsymbol{V}_t,\\
        \displaystyle\big[\boldsymbol{v}_j, \boldsymbol{0}_{1\times3}\big]\cdot \displaystyle\boldsymbol{x}=\boldsymbol{0},\quad& \text{if } \displaystyle\boldsymbol{v}_j \in \boldsymbol{V}_r.
    \end{split}
     \label{eq:6d_constraint_form}
\end{align} 
The constraints are then stacked into a matrix of form $\boldsymbol{C}\boldsymbol{x}=\boldsymbol{d}$ and integrated into the \ac{ICP} problem formulation, via 
\begin{align}
    \displaystyle\underbrace{\left[
    \begin{array}{ccc}
        \boldsymbol{0}_{m_r\times3}&    & \boldsymbol{v}_j \\
        \vdots &    &  \vdots \\
        \boldsymbol{v}_j &    &\boldsymbol{0}_{m_t\times3} \\
    \end{array}\right]}_{\boldsymbol{C}_{(m_r + m_t) \times 6}}
    \displaystyle\boldsymbol{x}=
        \underbrace{\left[
    \begin{array}{c}
        \boldsymbol{0} \\
        \vdots\\
        \boldsymbol{0} \\
    \end{array}\right]}_{\boldsymbol{d}_{(m_r + m_t) \times 1}},
\end{align}
where the number of equality constraints is denoted as $m_t$ translational and $m_r$ rotational ones, resulting in a total number of $c=m_t+m_r$ constraints. 

The unconstrained \ac{ICP} cost function shown in~\Cref{eq:final_cost} can then be re-written as a constrained optimization
\begin{align}
    \begin{split}\mathop{\text{min}}_{\boldsymbol{x}\in \mathbb{R}^{6}}  \quad & \Big|\Big|\boldsymbol{A}
        \boldsymbol{x}- \boldsymbol{b}\Big|\Big|_2,\\
        \textrm{s.t.} \quad & \boldsymbol{C}\boldsymbol{x}-\boldsymbol{d}=0.
    \end{split}
\label{eq:constrained_final_cost}
\end{align}
Introducing Lagrangian multipliers can convert an equality-constrained optimization problem into an unconstrained optimization problem~\cite{xicp}.
The Lagrangian of the minimization problem of~\Cref{eq:constrained_final_cost} is defined as
\begin{align}
    \begin{split}
        \displaystyle\mathbb{L}(\boldsymbol{x}, \boldsymbol{\lambda}) = \:\boldsymbol{x}^{\top} \mathbf{A}^{\top} \mathbf{A} \boldsymbol{x} -2\boldsymbol{b}^{\top} \mathbf{A} \boldsymbol{x} 
        +\:\boldsymbol{\lambda}^{\top}(\mathbf{C} \boldsymbol{x}-\boldsymbol{d}) 
        + \boldsymbol{b}^{\top} \boldsymbol{b},
    \end{split}
\label{eq:lagrangian}
\end{align}

where $\boldsymbol{\lambda} \in \mathbb{R}^{c}$ are the Lagrangian multipliers. Here, $c$ indicates the number of constraints in the optimization.
Interestingly, the effect of the constraints can be inferred by the magnitude of the Lagrangian multiplier the constraint is paired with. 
As the effect of the constraint gets larger, the magnitude of the Lagrangian multiplier increases accordingly.
Similarly, if the magnitude of the Lagrangian multiplier is close to zero, the constraint has a minimal effect on the optimization.
%
%
The effect of the constraints can be seen by comparing~\Cref{eq:partial_derivative} and~\Cref{eq:lagrangian}, with the only difference being the added constraint terms.
Re-formating the Lagrangian into the matrix form reveals the familiar least-squares formulation
\begin{align}
    \mathop{\text{min}}_{\boldsymbol{x}^{\prime} \in \mathbb{R}^{6+c}} \left(
    \underbrace{
        \left[
            \begin{array}{cc}
            2\boldsymbol{A}^{\top} \boldsymbol{A} & \boldsymbol{C}^{\top} \\
            \boldsymbol{C} & \boldsymbol{0}
            \end{array}
        \right]
    }_{\boldsymbol{A^{\prime}}}
    \underbrace{
        \left[
            \begin{array}{c}
            \boldsymbol{x}^{*} \\
            \boldsymbol{\lambda}^{*}
            \end{array}
        \right]
    }_{\boldsymbol{x}^{\prime}}-
    \underbrace{
        \left[
            \begin{array}{c}
            2\boldsymbol{A}^{\top} \boldsymbol{b} \\
            \boldsymbol{d}
            \end{array}
        \right]
    }_{\boldsymbol{b}^{\prime}}\right).
\end{align}
Here, $\boldsymbol{C}\in\mathbb{R}^{c\times6}$ is the constraint matrix, relating the optimization variables $\boldsymbol{x}$ to the constraints $\boldsymbol{d}\in\mathbb{R}^{c\times1}$.
Furthermore, the augmented optimization vector $\boldsymbol{x}^{\prime}$ constitutes the optimization variables and the Lagrangian multipliers.
This augmented cost function can be solved using \ac{SVD}, similarly to the standard point-to-plane \ac{ICP} cost function in~\Cref{eq:final_cost}.
Theoretically, the activation of the degeneracy constraints can be investigated through the value of the Lagrangian multipliers $\boldsymbol{\lambda}$. The higher the amplitude of the multiplier, the more the constraint affects the optimization~\cite{lagrangianMultipliers}. This notion suggests that the redundancy of the degeneracy mitigation constraints can be analyzed
through $\boldsymbol{\lambda}$.
In the context of this work, this method adds hard constraints to the optimization and, through the generation
of the Lagrangian (\Cref{eq:lagrangian}), affects the error characteristics of the optimization. The relative error on
$\boldsymbol{b}$ and the residual $\boldsymbol{r}$ are bounded by the added constraints, also bounding the sensitivity of the least-squares
optimization problem to perturbations (\Cref{eq:perturb_A} and \Cref{eq:perturb_b}).

\subsubsection{INEQUALITY CONSTRAINTS}
Although equality constraints provide a straightforward way to deal with degenerate directions, the system's systematic errors, such as eigenvector calculation errors, sensor noise, and the iterative nature of \ac{ICP}, might require relaxation to escape local minima.
In this scenario, the \ac{Ineq. Con.} method provides a simple yet powerful alternative to equality constraints with a potentially larger convergence basin
\begin{align}
   \begin{split}
       \mathop{\text{min}}_{\boldsymbol{x}\in \mathbb{R}^{6}}  \quad & \Big|\Big|\boldsymbol{A}
       \boldsymbol{x}- \boldsymbol{b}\Big|\Big|_2,\\
       \textrm{s.t.} \quad & -\boldsymbol{\epsilon} \leq \boldsymbol{L}\boldsymbol{x} \leq \boldsymbol{\epsilon}.
   \end{split}
   \label{eq:ineq_least_squares}
\end{align}
Here, the constraint matrix $\boldsymbol{L}_{6\times6}$ is generated as a row-major matrix based on the optimization's degenerate directions, with rows indicating the degenerate directions. As discussed in \Cref{sec:localizability}, an X-ICP~\cite{xicp} based degeneracy detection method retrieves the degenerate directions.
The upper and lower bounds of the inequality constraints can differ, but for simplicity, both are set to the same magnitude vector, $\boldsymbol{\epsilon}$ of dimension $6\times1$. Importantly, to account for the difference in scale between translation and rotation components, the amplitude of the rotational and translational components is set to $\sfrac{\epsilon}{2}$ and $\epsilon$, resulting in
$\boldsymbol{\epsilon} = \left[\sfrac{\epsilon}{2},\sfrac{\epsilon}{2},\sfrac{\epsilon}{2},\epsilon,\epsilon,\epsilon  \right]^\top$. 
Here, $\epsilon$ is selected after heuristic tuning on the Ulmberg tunnel dataset, 
and set to $\epsilon = 0.0014$.

It is not trivial to solve inequality-constrained optimization problems as it is for equality constraints. Thus, to find the solution, the above formulation is converted to a quadratic program (QP)
\begin{align}
    \begin{split}
       \mathop{\text{min}}_{\boldsymbol{x}\in \mathbb{R}^{6}}   \quad &  \frac{1}{2} \boldsymbol{x}^{\top} \boldsymbol{F} \boldsymbol{x}+\boldsymbol{f}^{\top} \boldsymbol{x}, \\
        \textrm{s.t.} \quad & -\epsilon \leq \boldsymbol{L} \boldsymbol{x} \leq \epsilon.
    \end{split}
\end{align}
Here, $\boldsymbol{F} = 2{\boldsymbol{A}^{\top} \boldsymbol{A}}$ and $\boldsymbol{f} = -2{\boldsymbol{A}}^{\top} \boldsymbol{b} $. 
The QP problem formulation suits the cases where $\boldsymbol{F}$ might be ill-posed and can handle different types of constraint (equality and inequality at the same time); however, for the sake of clear comparison, only inequality constraints are used in the investigation presented in this work. 
The open-source QP library\footnote{\url{https://github.com/asherikov/qpmad}} \textit{QPmad}~\cite{qpmad_but_not} is utilized to solve this QP problem. QPmad internally employs the Goldfarb and Idnani dual primal method~\cite{goldfarb2006dual}. 
The minimum of the objective function, subject to the current active constraint set, is calculated at every iteration~\cite{qpmad_but_not}. 
The additional inequality constraints can be removed from the active constraint set if the constraint is no longer active, which relaxes the constrained optimization. The state of the inequality constraint can be inferred from its dual variables (Lagrangian multipliers). If the constraints are not activated, the bounds of the inequality constraint $\boldsymbol{\epsilon}$ are larger in magnitude than the motion update in that direction.
Similar to \ac{Eq. Con.}, this method adds additional degeneracy-aware constraints to the optimization to provide bounds for the relative error and the residual $\boldsymbol{r}$. Subsequently, the effect of the perturbations on the optimization variables is reduced, as discussed in~\Cref{sec:formulation:sensitivity}.


\subsubsection{SOLUTION REMAPPING~\label{sec:methods:solRemap}}
\textit{Solution Remapping} denotes the optimization degeneracy mitigation method proposed by the authors of \cite{solRemap}, shown in~\Cref{fig:methods} as Zhang \textit{et al.} method. This method is referred to as Zhang \textit{et al.} from now on.
In this method, a solution projection matrix is constructed from the eigenvalue decomposition of the optimization Hessian, which is used to project the degenerate optimization solution $\boldsymbol{x}$ only in well-constrained directions.
First, this method generates an augmented eigenvector matrix $\boldsymbol{V}_{{aug}}$ based on the degeneracy detection by setting the corresponding eigenvectors in $\boldsymbol{V}$ to $\boldsymbol{0}$.
Using the introduced definitions in~\Cref{sec:problem_formulation}, the reprojection matrix is calculated as
\begin{align}
   \boldsymbol{S} = \boldsymbol{V}_{aug}^{-1}\boldsymbol{V}.
\end{align}
After minimizing \Cref{eq:final_cost}, the reprojection matrix is applied to the solution $\boldsymbol{x^{*}}$ via
\begin{align}
   \boldsymbol{x^{*}_{aug}} = \boldsymbol{S} \boldsymbol{x}^{*}.
\end{align}
%
This algorithm relies on the notion that the optimization degrades only in the degenerate directions, and identification of the null-space of the Hessian is sufficient to reproject the solution on the well-constrained directions of the optimization. As in this work, the null space of the Hessian is assumed to be available; only the reconstruction of the projection matrix is required. In the context of this work, Zhang~\textit{et al.} does not interfere with $\text{cond}(\boldsymbol{A})$ or the residual $\boldsymbol{r}$ but projects the solution $\boldsymbol{x}$ to the well-constrained directions of the optimization. Consequently, the noise in the input can be amplified during the least-squares optimization (\Cref{sec:formulation:sensitivity}).
Interested readers are referred to the original work~\cite{solRemap} for more detailed derivation and explanations. 

\subsubsection{TRUNCATED SVD}
\label{sec:methods:active:tsvd}
\ac{TSVD} is a linear algebra method to handle ill-conditioned matrices with high-condition numbers to effectively reduce their rank and improve their computational stability~\cite{hansen1990truncated,draisma2018best}.
The \ac{TSVD} of a symmetric and semi-positive definite matrix involves truncating the smallest eigenvalues while retaining only the largest eigenvalues. As a result, producing a \textit{k}-rank approximation of a matrix where $k$ refers to the top-$k$ retained eigenvalues of the matrix.
%
\ac{TSVD} requires the eigenvalue decomposition for the Hessian defined in~\Cref{eq:final_cost}, which is given as 
\begin{align}
   \boldsymbol{A} = \boldsymbol{U} \boldsymbol{\Sigma} \boldsymbol{V}^{\top}.
\end{align}
Next, the diagonal scaling matrix $\boldsymbol{\Sigma}$ is truncated by overwriting the eigenvalues $ \sigma_i \forall~i \in \{1, \dots, 6\}$ in the diagonal entries, based on whether the corresponding eigenvector of the eigenvalue aligns with a degenerate direction. If an eigenvector is along a degenerate direction, then $\sigma^{tr}_i = 0$, and if not, then $\sigma^{tr}_i = \sfrac{1.0}{\sigma_i}$. Here, the truncated diagonal entries are denoted with $\sigma^{tr}_i \forall~i \in \{1 \dots 6\}$. Setting $\sigma^{tr}_i = 0$ effectively creates a rank deficiency which LAPACK SVD routines\footnote{\url{https://netlib.org/lapack/explore-html-3.6.1/d1/d7e/}}.\looseness-1
%
%

The truncated Hessian can directly be used to find the solution $\boldsymbol{x^*}$\looseness-1
\begin{align}
\begin{split}
    \boldsymbol{x^*} &= \boldsymbol{A}_{tr}^{-1} \boldsymbol{b},\\
   \boldsymbol{A}_{tr}^{-1} &= \boldsymbol{V} \boldsymbol{\Sigma}_{tr} \boldsymbol{U}^{\top}.
   \end{split}
   \label{eq:tsvd}
\end{align}
The $\boldsymbol{\Sigma}_{tr}$ the truncated diagonal scaling matrix is defined as follows\looseness-1
\begin{align}
\boldsymbol{\Sigma}_{tr} =
\begin{bmatrix}
  \sigma^{tr}_1 & \multicolumn{2}{c}{\text{\kern0.5em\smash{\raisebox{-1ex}{\Large 0}}}} \\
  & \ddots &  \\
  \multicolumn{2}{c}{\text{\kern-0.5em\smash{\raisebox{0.75ex}{\Large 0}}}} & \sigma^{tr}_6
\end{bmatrix}
\end{align}

As $\boldsymbol{\Sigma}_{tr}$ is constructed through the truncated eigenvalues, it omits the under-constrained directions from the decomposition.
By truncating the eigenvalues, the constraints to mitigate the motion along the degenerate directions are directly integrated into the Hessian. By design, $\boldsymbol{A}_{tr}$ is rank-deficient; however, as \ac{SVD} exists for all matrices, \Cref{eq:svd_of_hessian} to \Cref{eq:svd_solving} can be used to acquire the solution.
\paragraph*{\textbf{On effectiveness of TSVD}} The Eckart-Young theorem~\cite{draisma2018best} provides detailed insights for the \ac{TSVD} method.
According to the theorem, the best least-squares approximation of $\boldsymbol{A}$ by a rank $k$ is $\boldsymbol{A}_{tr}$, and the truncation approximation 2-norm error is defined as the sum of the truncated eigenvalues,
\begin{align}
    \|\boldsymbol{A} -\boldsymbol{A}_{tr}\|_F = \sqrt{\sum_{i=k+1}^6 \sigma_i^2}.
\end{align}
Here, $F$ indicates the Frobenius norm. In addition to this observation, a closer look at \Cref{eq:tsvd} shows the increased sensitivity to noise in $\boldsymbol{b}$ for small eigenvalues. By truncating these small eigenvalues, the noise amplification effect is mitigated by TSVD. An example of noise in $\boldsymbol{b}$ would be errors in surface normal calculation and correspondence search. The effect of perturbation on the quality of the output of the least-squares minimization problem is discussed in~\Cref{sec:formulation:sensitivity}; in this context, by truncating the near-singular eigenvalues, \ac{TSVD} changes the condition number $\text{cond}(\boldsymbol{A})$, the noise in $\boldsymbol{b}$,  the residual $\boldsymbol{r}$ and the relative error.

\subsubsection{TIKHONOV REGULARIZED LINEAR LEAST-SQUARES}\label{sec:methods:active:lreg}
Regularization, commonly employed in optimization techniques such as Levenberg Marquardt (\ac{LM}), offers an easy way to escape singularities and improve optimization characteristics. Although the characteristics and properties of Tikhonov regularization are well-established in the linear algebra community~\cite{golub1999tikhonov}, the benefits in the field of degeneracy aware point cloud registration have not yet been investigated.
Motivated by this idea, \ac{L-Reg.}, shown in~\Cref{fig:methods} as \ac{L-Reg.}, is employed for a sub-space of the linear least-squares minimization based on the null-space approximation provided by the localizability information. 
%
%
The previously defined cost function in~\Cref{eq:final_cost} can be re-written as a regularized linear least-squares problem as follows
\begin{align}
    \begin{split}
        \min _x \|\boldsymbol{A} \boldsymbol{x}-\boldsymbol{b}\|_2^2+\lambda\|\boldsymbol{L}\boldsymbol{x}\|_2^2
         \\
        = \min _x\left\|\binom{\boldsymbol{A}}{\sqrt{\lambda} \boldsymbol{L}} \boldsymbol{x}-\binom{\boldsymbol{b}}{\boldsymbol{0}}\right\|_2^2.
    \end{split}
\end{align}
Here, the scalar $\lambda$ is the linear regularization parameter, and $\boldsymbol{L} \in \mathbb{R}^{c \times 6}$ is the constraint matrix, with $c$ denoting the number of degenerate directions. The number of degenerate directions is provided by the localizability detection block discussed in~\Cref{sec:localizability}.
Following a similar derivation to~\Cref{eq:normal_equation}, the normal equations of the regularized minimization are
\begin{align}
    \left(\boldsymbol{A}^{\top} \boldsymbol{A}+\lambda \boldsymbol{L}^{\top} \boldsymbol{L}\right) \boldsymbol{x}=\boldsymbol{A}^{\top} \boldsymbol{b}.
\label{eq:regularized_normal_equations}
\end{align}
Immediately, $\lambda$ can be identified as the regularization scaling parameter to the additional cost matrix $\boldsymbol{L}^{\top} \boldsymbol{L}$ induced by the regularization term.
This term provides the penalization needed to minimize the projection of the solution onto the null-space approximation provided to the optimization (by improving the conditioning of the problem (cf.~\Cref{sec:problem_formulation})).
The regularization parameter $\lambda > 0$ is not known a priori and has to be determined based on the problem data. In this work, $\lambda$ is selected after heuristic tuning on the Ulmberg Tunnel dataset similar to \ac{Ineq. Con.} method
and set to $\lambda = 440$.
The constraint matrix $\boldsymbol{L}$ is set as a row-major matrix consisting of the degenerate directions of the optimization, $ \boldsymbol{v}_j \in \{\{\boldsymbol{V}_t,\boldsymbol{V}_r\} \in \mathbb{R}^{3 \times 3}\} $. 
\Cref{eq:regularized_normal_equations} can be solved with \ac{SVD} as described in~\Cref{sec:problem_formulation}.

\paragraph*{\textbf{Understanding the Tikhonov regularization}}
It is important to understand the effect of regularization on the underlying optimization. An analysis for the general case of $\boldsymbol{L} \neq \boldsymbol{I}$ requires the solution of the generalized \ac{SVD} of $(\boldsymbol{A}, \boldsymbol{L})$. Hence, for the brevity of understanding of the effects, the constraint matrix is set as $\boldsymbol{L}=\boldsymbol{I}$ for the following explanation.  
%
Similar to~\Cref{eq:svd_solving}, given the \ac{SVD} of $\boldsymbol{A}$, the steps to obtain the solution $\boldsymbol{x}_\lambda$ are as follows
\begin{align}
    (\boldsymbol{V} \boldsymbol{\Sigma}^{\top} \underbrace{\boldsymbol{U}^{\top} \boldsymbol{U}}_{=\boldsymbol{I}} \boldsymbol{\Sigma} \boldsymbol{V}^{\top}+\lambda \underbrace{\boldsymbol{I}}_{=\boldsymbol{V} \boldsymbol{V}^{\top}}) \boldsymbol{x}_\lambda=\boldsymbol{V} \boldsymbol{\Sigma}^{\top} \boldsymbol{U}^{\top} \boldsymbol{b}.
\end{align}
Similar terms are bundled together and re-arranged, which reveals how the regularization affects the Hessian.
Multiplying 
and setting $\boldsymbol{y}=\boldsymbol{V}^{\top} \boldsymbol{x}_\lambda$, the equation can be simplified to\looseness-1
\begin{align}
    \left(\boldsymbol{\Sigma}^{\top} \boldsymbol{\Sigma}+\lambda \boldsymbol{I}\right) \boldsymbol{y}=\boldsymbol{\Sigma}^{\top} \boldsymbol{U}^{\top} \boldsymbol{b}.
\end{align} 
Since $\boldsymbol{V}$ is orthogonal, the 2-norm of $\boldsymbol{y}$ is the same as $\boldsymbol{x}_\lambda$. The solution to the regularized problem can then be derived using the dyadic decomposition
\begin{align}
    \boldsymbol{x}_\lambda=\sum_{i=1}^{6} z_i\frac{\left(\boldsymbol{u}_i^{\top} \boldsymbol{b}\right)}{\sigma_i} \boldsymbol{v}_i \ , \quad z_i = \frac{{\sigma_i}^2}{\sigma_i^2+\lambda}.
    \label{eq:lreg_lambda}
\end{align}
Here $z$ is the filter coefficient damping the eigenvalues $\sigma_i < \lambda$ of $\boldsymbol{A}$ and subsequently eigenvalues $\sigma^{2}_i < \lambda$ of $\boldsymbol{A}^{\top}\boldsymbol{A}$. Hence, the regularization mainly affects the degenerate directions and effectively changes the Hessian's eigenvalues. 
Finally, as shown in~\Cref{eq:lreg_lambda}, this method alters the eigenvalues of the optimization Hessian and subsequently improves the condition number $\text{cond}(\boldsymbol{A})$ which in return improves the sensitivity characteristics of the least-squares minimization problem as shown in~\Cref{eq:perturb_A} and~\Cref{eq:perturb_b}.

\subsubsection{NON-LINEAR OPTIMIZATION WITH TIKHONOV REGULARIZATION}
\label{sec:methods:active:nlreg}
This section introduces the non-linear optimization-based regularization technique, \ac{NL-Reg.}, shown in~\Cref{fig:methods} as NL-Reg. Previous works such as~\cite{lmicp, rankDeficientSLAM} utilize non-linear cost functions to improve the \ac{ICP} behavior. In particular, the authors in~\cite{rankDeficientSLAM} proposed to incorporate the null space of the optimization as a regularization term for the non-linear cost function in the context of factor graph optimization.
Motivated by this, Tikhonov regularization is employed for point cloud registration with the non-linear cost function as
\begin{align}
     \begin{split}
     E&_{\text{NL}}(\boldsymbol{x})= 
    \sum_{i=1}^N \Big|\Big| \underbrace{\big((\boldsymbol{R} \boldsymbol{p}_i + \boldsymbol{t}) - \boldsymbol{ {q}}_i\big) \cdot {\boldsymbol{ {n}}}_i}_{{e_i}} \Big|\Big|_2 +\lambda_{D}\|\boldsymbol{L}\boldsymbol{x}\|_2^2,
    \end{split}
 \label{eq:nonlinear_cost}
\end{align}
where $\boldsymbol{x} = [\boldsymbol{r}^\top, \ \boldsymbol{t}^\top]^\top \: \in  \mathbb{R}^{6}$ are the optimization variables. Moreover, $\lambda_D$ is the degeneracy regularization parameter, and $\boldsymbol{L}$ is the row-major constraint matrix consisting of the degenerate directions of the optimization. Similarly to previous methods, the value of $\lambda_D$ is selected after heuristic tuning on the Ulmberg tunnel dataset, 
and set to $\lambda_D = 675$.
The well-known \ac{LM} algorithm is used for the minimization of the cost function in~\Cref{eq:nonlinear_cost}. The Jacobian, gradient and Hessian of cost $E(\boldsymbol{x})_{NL}$ with respect to $\boldsymbol{x}$ are defined as \looseness-1
\begin{align}
\begin{split}
      \boldsymbol{J_{\text{NL}}} &= \diffp[]{e}{\boldsymbol{x}},\\
     \nabla E_{\text{NL}}(\boldsymbol{x}) 
     &= 2\boldsymbol{J_{\text{NL}}}^{\top} e_i + 2\lambda_{D} \boldsymbol{L}^{\top} \boldsymbol{L} \boldsymbol{x}, \\
    \boldsymbol{H_{\text{NL}}} = \diffp[2]{E_{\text{NL}}(\boldsymbol{x})}{\boldsymbol{x}} &= 2\boldsymbol{J_{\text{NL}}}^{\top} \boldsymbol{J_{\text{NL}}} + 2\lambda_{D} \boldsymbol{L}^{\top} \boldsymbol{L}.
    \end{split}
\end{align}
Utilizing these definitions, the update rule of the \ac{LM} algorithm is derived as follows
\begin{align}
    \boldsymbol{x}_{\text{new}} = \boldsymbol{x} - ( \boldsymbol{H}_{\text{NL}} + \lambda_s \boldsymbol{I})^{-1} \nabla E_{\text{NL}}(\boldsymbol{x}).
    \label{eq:nonlinearupdate}
\end{align}
Here, $\lambda_s$ refers to the smoothness parameter of the \ac{LM} algorithm, ensuring the stability of the matrix inversion. As discussed in the Ceres documentation, the magnitude of the state correction might lead to a non-convergent algorithm. Hence, a trust region constraint is added to the solution. The reader is referred to the documentation of Ceres for more implementation details on robust non-linear least-squares. For clarity, the altered optimization termination tolerances are provided in Table~\ref{table:optconfig}.
\begin{table}[t]\centering
    \caption{Non-linear Optimizer configuration for the \ac{NL-Reg.} method.}
    \begin{threeparttable}

       \begin{tabular}{lcc}
       \toprule[1pt]
       & \makecell{Value} & \makecell{Default}\\
       \toprule[1pt]
       \makecell{Parameter Tolerance} & $10^{-3}$ & $10^{-8}$ \\[0.3em]
       \makecell{Function Tolerance} & $10^{-3}$ & $10^{-6}$ \\[0.3em]
       \makecell{Gradient Tolerance} & $10^{-6}$ & $10^{-10}$ \\[0.3em]
       \toprule[1pt]
       \end{tabular}
    \end{threeparttable}
    \label{table:optconfig}
    \vspace{-1.0em}
\end{table}

As this method utilizes the LM algorithm to solve the non-linear cost function \Cref{eq:nonlinear_cost}, the sensitivity analysis in \Cref{sec:formulation:sensitivity} does not apply directly. Nevertheless, \ac{NL-Reg.} method improves the conditioning of the Hessian $\boldsymbol{H_{\text{NL}}}$ by adding a degeneracy-aware regularization with $\lambda_D$. Besides, the LM algorithm is inherently more robust against ill-conditioned problems due to the regularization of the damping parameter $\lambda$.

\subsubsection{EXTERNAL PRIOR}
In this approach, if optimization ill-conditioning is detected, the initial guess $\boldsymbol{T}_{\mathtt{M} \mathtt{L}, \text{init}}$ is directly used for point cloud registration. 
Consequently, this operation renders point cloud registration ineffective by skipping the optimization process altogether. This method relies on external odometry estimates during degeneracy to propagate the solution; hence, it is denoted and shown in~\Cref{fig:methods} as \textit{Prior Only}. As this method completely skips the point cloud registration step in the event of optimization degeneracy, the least-squares optimization step of point cloud registration is skipped entirely. Since there are no pose updates from point cloud registration, the accuracy of this method depends on the accuracy of the external pose prior while LiDAR degeneracy is present.\looseness-1

%
\subsection{PASSIVE DEGENERACY MITIGATION}
\label{sec:methods:passive}
Passive degeneracy mitigation refers to the methods that are not \textit{actively} utilizing the detected degeneracy information to affect optimization. Instead, these methods use this information passively or rely on algorithms that are supposedly less affected by disturbances and degeneracy. 

\subsubsection{GLOBALLY OPTIMAL POINT CLOUD REGISTRATION}
\label{sec:methods:passive:global}In Quatro~\cite{lim2022single}, the authors define \textit{degeneracy} in globally optimal point cloud registration context as the \textit{absence of inlier correspondences} in the matching process. 
This definition differs from the LiDAR degeneracy definition used in this work (\Cref{sec:problem_formulation}), which focuses on the inherent lack of geometric information the environment provides along specific directions.
Quatro~\cite{lim2022single} is a robust globally optimal point cloud registration method estimating 4-\ac{DoF} transformation while requiring less than three correspondences. 
\ac{FGR}~\cite{fgr} is a well-known global registration method that performs a single-stage objective minimization to find the transform between two overlapping surfaces. Section~\ref{section:Experiments:global_sim} highlights that these globally optimal point cloud registration frameworks are not able to compensate for the lack of geometric constraints provided by the environment; inlier point correspondences also do not constrain the optimization along symmetric and self-similar directions of motion.\looseness-1

\subsubsection{M-ESTIMATORS}
\label{sec:methods:passive:m-estimator}
To analyze the efficacy of robust outlier rejection methods in point cloud registration, this work investigates the effectiveness of the Cauchy robust cost function, which is an M-Estimator, in combination with the \ac{MAD} scale estimator~\cite{babin2019analysis}. In the remainder of this work, this method is referred to as \textit{Cauchy} for simplicity and is shown in~\Cref{fig:methods} with the same name.
A non-convex robust cost function can be used in an \ac{IRLS}~\cite{irls} fashion, adapting the point cloud registration cost function defined in \Cref{eq:final_cost} as 
\begin{align}
    \mathop{\text{min}}_{\boldsymbol{x}\in \mathbb{R}^{6}} 
    \rho\left(\underbrace{\Big|\Big|\boldsymbol{A}
    \boldsymbol{x}- \boldsymbol{b}\Big|\Big|_2}_e\right).
    \label{eq:final_cost_with_robust}
\end{align}
As authors in~\cite{babin2019analysis} found the Cauchy robust cost function ${\rho}\left(\cdot\right)$ to be effective against the presence of outliers, it is selected as the cost function studied in this analysis. 
The corresponding function and the weight ${w\left(\cdot\right)}$ are formulated as follows\looseness-1
\begin{align}
    \rho\left(e\right) = \frac{\kappa^2}{2} \log \left(1+\left(\frac{e}{\kappa}\right)^2\right)\text{,} \quad {w\left(e\right)}=\frac{1}{1+{(e / \kappa)}^2}.
    \label{eq:robustnorm_and_weight}
\end{align}
Here, ${e}$ refers to the error term induced by each correspondence pair in the iterative process of \ac{ICP}, and $\kappa$ refers to the function tuning parameter. Here, the weight function ${w}\left(\cdot\right)$ is the decomposed weight of the non-convex ${\rho}\left(\cdot\right)$ cost function.
For details of the \ac{MAD} scale estimator and how to solve the IRLS problem, the readers are referred to~\cite{babin2019analysis}. 
As suggested in~\cite{babin2019analysis} the M-Estimator parameter ${\kappa}=1$ has been selected. Finally, using the weight calculated in~\Cref{eq:robustnorm_and_weight} the cost function~\Cref{eq:final_cost} is rewritten as
\begin{align}
    \mathop{\text{min}}_{\boldsymbol{x}\in \mathbb{R}^{6}} 
    {w}(e)\Big|\Big|\boldsymbol{A}
    \boldsymbol{x}- \boldsymbol{b}\Big|\Big|_2.
    \label{eq:weightedFinalCost}
\end{align}
In the context of this work, the Cauchy method improves the sensitivity of the least-squares problem by reducing the noise levels both in $\boldsymbol{b}$ and $\boldsymbol{A}$ and, subsequently, mitigates the adverse effects of noise amplification in ill-conditioned cases, however, do not alter the condition number of $\boldsymbol{A}$.

\subsubsection{GEOMETRICALLY STABLE SAMPLING}
As one of the seminal works on using sampling to improve the conditioning of \ac{ICP}, \cite{gelfand2003} proposed pair-wise information sampling from correspondence pairs $\left( \boldsymbol{p}, \boldsymbol{n} \right)$. In this method, point and surface normal pairs are sampled from the set of correspondences until a number of samples are collected or the conditioning threshold is met. Specifically, sampling is performed subsequently for all eigenvectors $\boldsymbol{v}_k, \ k \in \{1\dots 6\}$.\looseness-1
\begin{align}
\begin{split}
    {f}_i&= \left(\left[\boldsymbol{p}_i \times \boldsymbol{n}_i^\top, \boldsymbol{n}_i^\top \right]\cdot\boldsymbol{v}_k\right)^2, \quad i \in \{1\dots N\}, \\
    \hat{{\sigma}}_k&\approx \sum_{i=0}^{N}{f}_i.
    \end{split}
\end{align}
This approximation of the eigenvalues by summation of information allows for a graceful adjustment of the Hessian condition number. Specifically, the criterion is to resample the correspondences to construct a Hessian with a condition number as close to 1 as possible. This method provides an effective means of improving the condition number $\text{cond}(\boldsymbol{A})$ by resampling the information from the input data and consequently preventing the oversensitivity of the least-squares solution to perturbations as discussed in~\Cref{sec:formulation:sensitivity}.
This method is depicted as Gelfand \textit{et al.} in~\Cref{fig:methods}, and the open-source implementation in libpointmatcher~\cite{libpointmatcher} is used to replicate in this work.
\subsubsection{REDUNDANCY MINIMIZING SAMPLING (RMS)}
As discussed in~\Cref{section:related_works}, the authors in~\cite{petracek2024rms} proposed RMS as a means of minimizing redundancy in the input point cloud and, while, maximizing the entropy of the data in the same point cloud to retain information. Specifically, this method maximizes the entropy of information while keeping the Hessian well-conditioned by sampling points from the input point cloud in all directions of the optimization. The authors in this work introduce a heuristic called \textit{gradient flow}, which quantifies the uniqueness of a point with the flow of the geometric gradient. The gradient flow of each point $\boldsymbol{p}$ within a spherical neighborhood with radius $\lambda_p$ as:
\begin{align}
\begin{split}
\mathcal{N}_{\boldsymbol{p}} &= \{\boldsymbol{j} \mid \|\boldsymbol{j} - \boldsymbol{p}\|_2 < \lambda_p, \ \boldsymbol{j} \neq \boldsymbol{p}, \boldsymbol{j} \in \mathcal{P}\}, \\
\Delta_{\boldsymbol{p}} &= \frac{1}{|\mathcal{N}_{\boldsymbol{p}}|} \sum_{\boldsymbol{j} \in \mathcal{N}_{\boldsymbol{p}}} (\boldsymbol{j} - \boldsymbol{p}).
\end{split}
\end{align}
The gradient flow based data entropy maximization is defined as\looseness-1
\begin{align}
\hat{\mathcal{P}} = 
\operatorname*{arg\,max}_{\Omega \in \{\Theta \mid \Theta \subseteq \mathcal{P}_v, \Theta \neq \emptyset\}} 
H_{\Delta}(\Omega).
\end{align}
Here, $\mathcal{P}_v$ corresponds to the voxelized point cloud set and $\Omega$ to a subset of $\mathcal{P}_v$. Moreover, this maximization is conditioned on the normalized relative entropy rate to encourage early termination. Although this method does not balance the rotation-space observability, the authors specify that this is only relevant in high degrees of rotation. The readers are referred to the original work for a more detailed explanation and derivation of this method. Similarly to Gelfand \textit{et al.}, this method provides a means of sampling the input point cloud to maximize information gain and improve the condition number $\text{cond}(\boldsymbol{A})$ while minimizing the information redundancy to decrease possible noise in $\boldsymbol{b}$ and $\boldsymbol{A}$. As a result, this method mitigates the amplification effect of a sensitive least-squares optimization problem, as discussed in~\Cref{sec:formulation:sensitivity}.
In the context of this work, {RMS} is considered a passive degeneracy mitigation method, as it alters the input point cloud instead of actively using the detected degeneracy information. This method is depicted as Petracek \textit{et al.} in~\Cref{fig:methods} and in the rest of this work. The open-source implementation of the original work\footnote{\url{https://github.com/ctu-mrs/RMS}} is used to realize this work.

\subsubsection{POINT-TO-PLANE ICP}\label{sec:methods:p2plane}
Point-to-plane refers to the \ac{ICP} formulation using the point-to-plane~\cite {point_to_plane} cost function as previously detailed in~\Cref{sec:problem_formulation}. This method is referred to as P2Plane in the remainder of this work.
\ac{P2Plane}, is a passive baseline method and is shown in~\Cref{fig:methods} with the same name. The results of the baseline method show the adverse effects of \ac{LiDAR} degeneracy on point cloud registration with linear solvers and \ac{LiDAR} pose estimation if no degeneracy-aware actions are taken to mitigate the effects of optimization ill-conditioning.
The optimization cost function of this method is defined in~\Cref{eq:final_cost}. 
As this method does not change the sensitivity properties of the problem, it serves as a baseline in the evaluation.

\subsubsection{NON-LINEAR POINT-TO-PLANE ICP}\label{sec:methods:nlp2plane}
This baseline method, \ac{NL-Solver}, is introduced to understand the effect of the solver. Specifically, this method uses the cost function introduced in~\Cref{eq:nonlinear_cost} and \Cref{eq:nonlinearupdate} using $\lambda_D = 0$. Similar to the \ac{NL-Reg.} method, this baseline uses Ceres to implement the cost function and uses the LM algorithm with default regularization. The optimization termination criteria for this method are determined by the parameters defined in Table~\ref{table:optconfig}. This method is shown as \ac{NL-Solver} in~\Cref{fig:methods}. Compared to the \ac{P2Plane} baseline method, \ac{NL-Solver} uses the gradient-based LM solver. Finally, \ac{NL-Solver} does not interfere with the conditioning of $\text{cond}(\boldsymbol{A})$ with degeneracy-aware regularization; however, it uses the damping parameter $\lambda$ to improve stability during the update step of the LM algorithm.

\begin{figure}[h]
\centering
\includegraphics[width=0.95\linewidth]{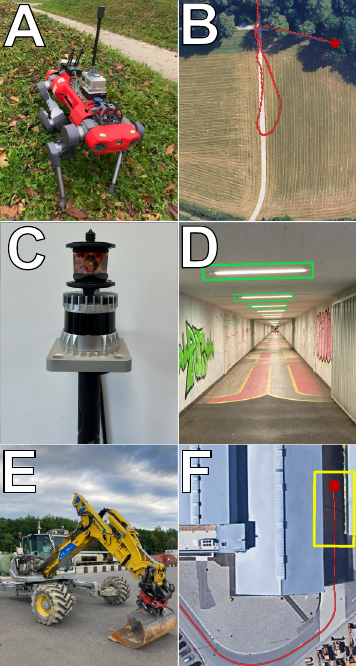}
\caption{An illustration of three real-world deployments with environmental LiDAR degeneracy. \textbf{A:} The ANYmal robot in the deployed environment. \textbf{B:} The robot's GPS trajectory (in red) over a satellite image of the deployed environment is shown. The shown open field is subject to 3-axes LiDAR degeneracy. \textbf{C:} The handheld sensor payload of the Ulmberg bicycle tunnel~\cite{enwide} experiment is shown. \textbf{D:} The degenerate bicycle tunnel is shown. In addition, the sunken LED light housings are highlighted in green, as these housings are the only identifiable geometric features along the tunnel direction. \textbf{E:} The HEAP~\cite{heap} walking excavator. \textbf{F:} The construction environment for the excavator experiment is shown with the trajectory of HEAP overlaid as a red line. The \ac{LiDAR} degenerate narrow-passage (highlighted in yellow) at the start (red dot) poses a significant challenge to LiDAR-SLAM solutions.}
\label{fig:Experiments:deployments}
\end{figure}

\subsection{IMPLEMENTATION DETAILS}\label{subsection:testingSetup}

All methods described in this section are implemented in the \textit{libpointmatcher}~\cite{libpointmatcher} open-source registration library and are integrated into the \textit{Open3d SLAM}~\cite{jelavic2022open3d} framework as the \mbox{scan-to-submap} registration module.
\textit{Open3d SLAM} runs at the \ac{LiDAR} rate of \SI{10}{\hertz}, and utilizes external odometry pose estimates as the registration prior $\boldsymbol{T}_{\mathtt{M}\mathtt{L}, \text{init}}$, enabling \mbox{scan-to-submap} nearest-neighbor search to identify the point correspondences.
The surface normal of a point is calculated using the k-nearest-neighbor search and principal component analysis, where the neighborhood is set as $k=10$. The LiDAR scans are cut off at a \SI{70}{\meter} maximum distance and processed without downsampling. Furthermore, all methods use the trimmed outlier filter to replicate a realistic system. In addition, the maximum number of iterations for the point-to-plane \ac{ICP} is set to 30 iterations. The ICP algorithm terminates early if the norm of the pose update drops to \SI{0.01}{\meter} for translation or if the angular distance between updates is less than \SI{0.001}{\radian}. These values are based on the default values used in \textit{libpointmatcher}.

Moreover, as discussed in~\Cref{subsubsection:degeneracy_detection}, a localizability module similar to X-ICP~\cite{xicp} is used as a localizability source for all methods requiring localizability information. Specifically, the X-ICP is simplified to do binary localizability classification, namely, localizable and non-localizable, to prevent giving an unintentional advantage to more complex methods. This method requires 3 localizability parameters, $\kappa_1$, $\kappa_2$, $\kappa_3$ and 1 filtering parameter $\kappa_f$. The localizability parameters are set based on the basin of convergence of the employed SLAM framework and the ICP algorithm. Furthermore, these parameters are set as \mbox{$\kappa_1 \geq \kappa_2 > \kappa_3$}. For all experiments in this work, the parameters are kept the same and set as $\kappa_1=300$, $\kappa_2=150$, and $\kappa_3=45$. These values are set only once according to the voxelization and filtering characteristics of \textit{Open3d SLAM}. Moreover, the filtering parameter is set as $\kappa_f=80$ for all experiments except the HEAP excavator experiment discussed in \Cref{section:Experiments:excavation}. For that experiment, it is set to $\kappa_f=60$ due to an older and noisy LiDAR variant Ouster OS0 Rev-1.

The parameter dependent methods \ac{L-Reg.}, \ac{NL-Reg.}, and \ac{Ineq. Con.} are tuned based on performance in the Ulmberg bicycle tunnel experiment~(\Cref{section:Experiments:tunnel}). In addition, the RMS method by \cite{petracek2024rms} and Gelfand \textit{et al.} also have a single tunable parameter each. The tunable parameter of RMS \textit{lambda} is set to $0.0042$ for experiments discussed in \Cref{section:Experiments:excavation} and \Cref{section:Experiments:tunnel}. For the remaining experiments, the value is set to $0.0036$. These values are obtained after testing RMS with different \textit{lambda} values and empirically selecting the most successful ones. For the method of \cite{gelfand2003}, when the number of points in a scan exceeds 1000, the maximum number of sampled points is set to \SI{60}{\percent} of the total points. The limit for scans with fewer points is set to \SI{90}{\percent} of the available points.
All evaluations were performed on a desktop computer with an \textit{Intel i9 13900K CPU}, and to guarantee repeatability and deterministic execution, all operations are performed in a single-threaded fashion.

\section{EXPERIMENTS AND EVALUATION} \label{section:Experiments}
In this section, the different degeneracy mitigation methods are analyzed in depth using simulation studies and real-world field experiments to identify the advantages and weaknesses of each method.\looseness-1

First, a simulated iterative degenerate point cloud registration experiment is performed in~\Cref{section:Experiments:simulation} to determine the feasibility and provide a deeper understanding of each method. Later, a simulated global point cloud registration study is discussed in Section~\ref{section:Experiments:global_sim}, showing the impact of global point cloud registration methods in mitigating LiDAR degeneracy. Finally, in Section~\ref{section:Experiments:static_sim}, the compared methods are tested on simulated data with \textit{Open3D-SLAM}~\cite{jelavic2022open3d} in the loop.\looseness-1

In addition, the methods are tested in different real-world robotic field deployments as shown in \Cref{fig:Experiments:deployments}, that include natural open fields (\Cref{section:Experiments:forest}), urban tunnels (\Cref{section:Experiments:tunnel}), and construction sites (\Cref{section:Experiments:excavation}). The real-world tests are also conducted using a diverse combination of robotic platforms and sensor payloads, namely, \textit{i)} the ANYmal legged robot~\cite{ANYmal} equipped with a Velodyne VLP-16 LiDAR (Section~\ref{section:Experiments:forest}), \textit{ii)} the HEAP~\cite{heap} excavator equipped with an Ouster OS0-128 LiDAR (Section~\ref{section:Experiments:excavation}), and \textit{iii)} the ENWIDE hand-held sensor pack\footnote{\url{https://projects.asl.ethz.ch/datasets/enwide}}~\cite{enwide} equipped with an Ouster OS1-128 LiDAR (Section~\ref{section:Experiments:tunnel}). 

Each experiment introduces a unique subset of real-world challenges for the existing point cloud registration frameworks, allowing for a detailed analysis of the compared methods. Importantly, some of these unique challenges render the possibility of recovering the true point cloud map of the environment infeasible. For example, in the Ulmberg tunnel experiment (\Cref{section:Experiments:tunnel}), the LiDAR degeneracy continues for more than 80\% of the dataset, making it infeasible for LiDAR-based solutions to provide a reasonable map. In this kind of scenario, the performance is mainly driven by better usage of the external motion prior to mitigate the effects of LiDAR degeneracy. In contrast, the experiments discussed in~\Cref{section:Experiments:forest} and \Cref{section:Experiments:excavation} either have mild LiDAR degeneracy where some geometric information is still present or have a short duration of LiDAR degeneracy, which the scan-to-submap registration can recover from. In accordance with the challenges mentioned above, the level of degeneracy and expected behavior of the investigated methods are summarized in~\Cref{table:dataset_summary}.
%
\begin{table}[t]\centering
     \caption{Challenge summary of the experiments presented in~\Cref{section:Experiments}. {\raisebox{-0pt}{{\large{\cmark}}}} indicates positivity and {\raisebox{-2pt}{{\large\xmark}}} indicates negativity.} 
    \resizebox{1\columnwidth}{!}{
    \begin{threeparttable}
    \begin{tabular}{ccccc}
    \toprule[1pt]
    & \large{\makecell{Good \\ Motion Prior}} & \large{\makecell{Severe \\ Degeneracy}} & \large{\makecell{Long Degeneracy \\ Duration}} & \large{\makecell{Is recovery \\ feasible?$^*$}} \\
    \toprule[1pt]\vspace{1mm}
    \Large{\makecell{Static Sim. (\Cref{section:Experiments:static_sim})}} & \Large\xmark         & \Large\cmark             &  \Large\textbf{-}& \Large\xmark \\[0.3em]
    \Large{\makecell{Global Sim. (\Cref{section:Experiments:global_sim})}} & \Large\xmark         & \Large\cmark             &  \Large\textbf{-} & \Large\xmark \\[0.3em]
    \Large{\makecell{Dynamic Sim. (\Cref{section:Experiments:dynamic_sim})}} & \Large\cmark         & \Large\cmark             &  \Large\cmark & \Large\cmark \\[0.3em]
    \Large{\makecell{Ulmberg Tunnel (\Cref{section:Experiments:tunnel})}} & \Large\cmark             & \Large\xmark             & \Large\cmark & \Large\xmark \\[0.3em]
    \Large{\makecell{ANYmal Forest (\Cref{section:Experiments:forest})}} & \Large\xmark              &  \Large\xmark             &  \Large\xmark & \Large\cmark \\[0.3em]
    \Large{\makecell{HEAP Excavator (\Cref{section:Experiments:excavation})}} & \Large\cmark         & \Large\cmark             & \Large\xmark & \Large\cmark \\[0.3em]
    \toprule[1pt]
    \end{tabular}\label{table:dataset_summary}
    \begin{tablenotes}
    \item[*] Here, recovery feasibility indicates the availability of sufficient information in the environment.
    \end{tablenotes}
    \end{threeparttable}
    }
\end{table}
%
\begin{figure}[b!]
    \centering
    \includegraphics[width=1\linewidth]{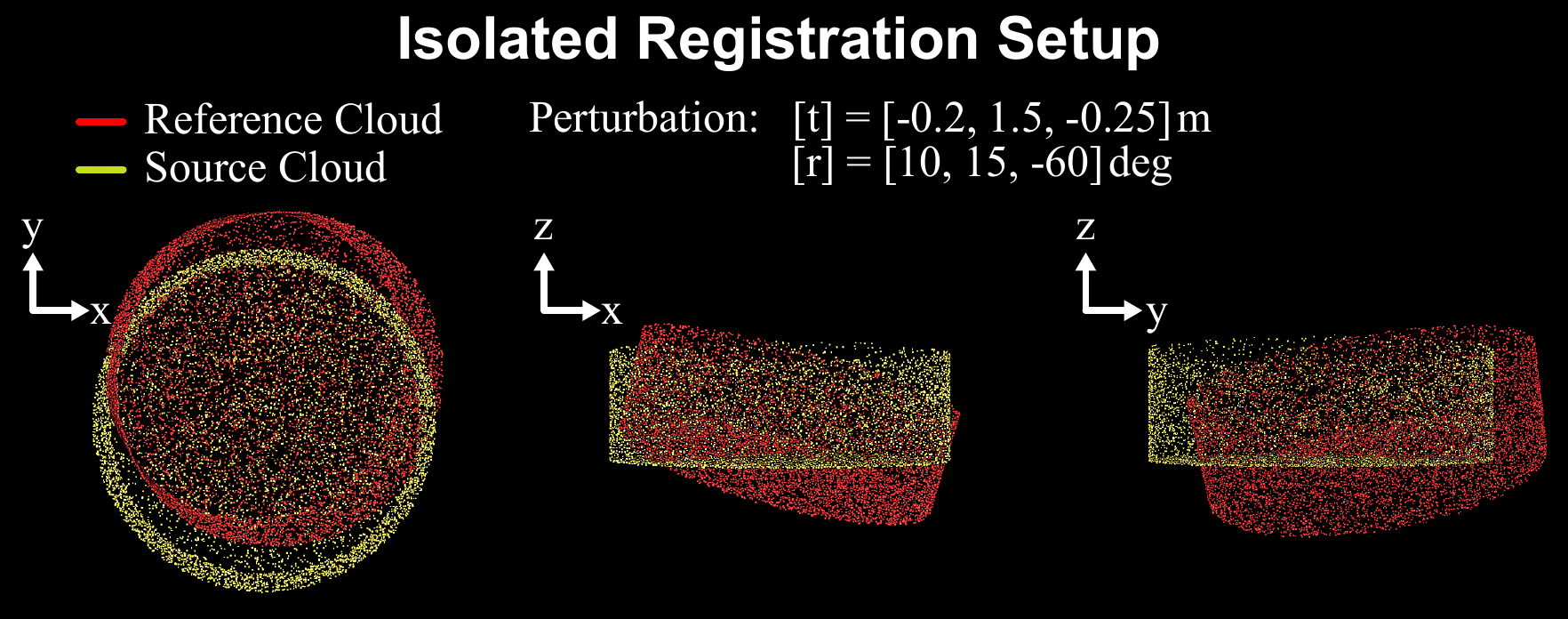}
    \caption{Setup of the static simulation experiment is shown. The red reference point cloud is shifted by $\boldsymbol{r}$ and $\boldsymbol{t}$ relative to the yellow source point cloud. The registration goal is to find the best-aligning transformation between the two point clouds.}
    \label{fig:Experiments:static_sim_setup}
\end{figure}

\begin{figure*}[t]
    \centering
    \includegraphics[width=1\linewidth]
    {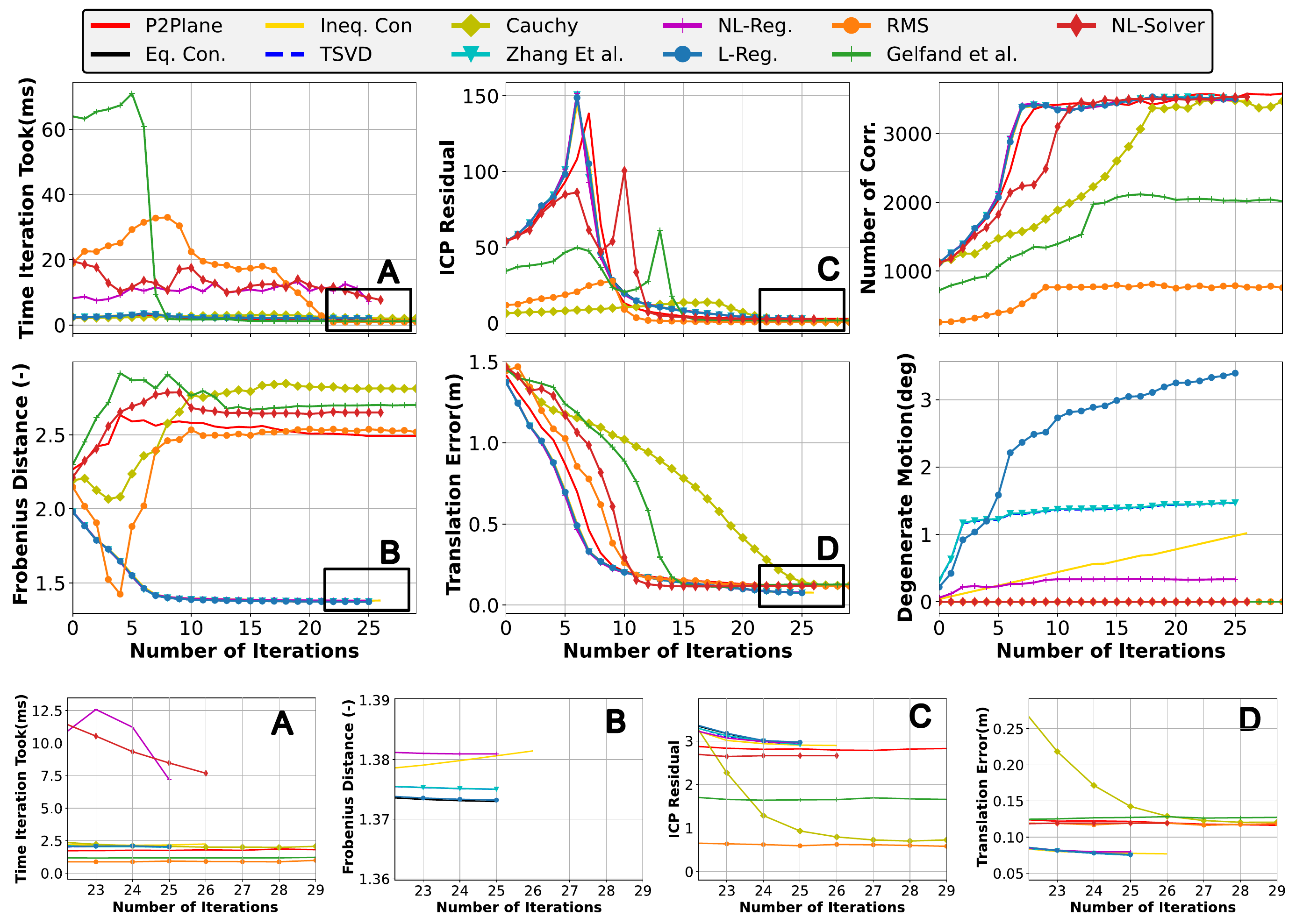}
    \caption{
    Analytics of the degenerate point cloud registration for the iterative registration experiment (\Cref{section:Experiments:static_sim}). \textbf{Top left:} Computational time for each iteration. \textbf{Top middle:} The residual cost of the \ac{ICP} formulation at the time of each iteration. \textbf{Top right:} The number of correspondences at the time of each iteration. \textbf{Center left:} The Frobenius distance to the true transform (the y-axis is unitless). \textbf{Center middle:} The translation errors of the registration. \textbf{Center right:} The motion residual along the degenerate direction during the registration process.}
    \label{fig:standalone_simulation_statistics}
\end{figure*}

\subsection{SIMULATION STUDIES} \label{section:Experiments:simulation}
A set of simulated experiments are conducted to understand the applicability and effectiveness of each method.
First, an iterative point cloud registration experiment is conducted in Section~\ref{section:Experiments:static_sim} to isolate the point cloud registration step from the mapping framework.
The registration step includes two point clouds with known perturbation and initial transformation guesses.
Second, Section~\ref{section:Experiments:global_sim} analyzes the feasibility of globally optimal point cloud registration methods using the same setup. 
Finally, a simulated walking ANYmal~\cite{ANYmal} robot experiment is conducted with \textit{Open3d-SLAM} in the loop to perform LiDAR-SLAM during a simulated robotic application.

\subsubsection{STATIC SIMULATION - ITERATIVE POINT CLOUD REGISTRATION}
\label{section:Experiments:static_sim}
In this experiment, each method is provided with source and reference point clouds, shown in~\Cref{fig:Experiments:static_sim_setup}, sampled from a simulated, perfectly cylindrical, and symmetric mesh. The reference point cloud is perturbed by a known transformation, and an identity initial guess associates these point clouds. 
The purpose of this experiment is two-fold: first, to analyze the performance of each method in a rotation-only (rotation around the Z-axis), one \ac{DoF}, degenerate setting and, second, to inspect the evolution of the solutions per iteration of the \ac{ICP} algorithm.
The Prior Only method is omitted, as this experiment focuses on the registration task, and the Prior Only method skips the registration step.
Similar to other experiments, the methods are employed for a \ac{ICP} registration task with a maximum iteration count of $30$. The results are shown in~\Cref{fig:standalone_simulation_statistics}.

First, as seen from the top left plot in~\Cref{fig:standalone_simulation_statistics}, the non-linear optimization based methods \ac{NL-Reg.} and NL-Solver have higher computational costs compared to the other methods. In addition, sampling-based methods took more time in the beginning, possibly due to a lower number of correspondences. Other methods are comparable in computational complexity and suited for real-time deployment.
Second, the ICP residual is minimized for all methods to a similar magnitude after a spike in the beginning (top middle) for all methods except Cauchy due to a rapid increase in correspondence inliers (top right).
However, this minimized cost does not necessarily reflect convergence to the actual transformation; it only indicates a successful residual minimization. 
In particular, this can be seen in the plot of the Frobenius distance as many methods that have reduced the ICP residual, in fact, have increased the Frobenius distance to the true transform. The reason is that every rotation along the axis of the degenerate rotation direction is infact a local-minima for the optimization.
However, all active degeneracy mitigation methods correctly utilize degeneracy information and prevent further increases in rotational error while decreasing the translation error (center middle).
Interestingly, the difference in internal dynamics of the active degeneracy mitigation methods is revealed by comparing the total motion residual in the degenerate direction (rotation around Z, center right).
As expected, \ac{Eq. Con.} does not produce any motion residual as the degeneracy constraints are utilized as equality constraints.
Among the soft-constrained methods, \ac{NL-Reg.} allows some motion until iteration $10$ while \ac{Ineq. Con.} allows a linear \textit{creep} motion as the optimization can move along the degenerate direction in every registration step. 
In contrast, \ac{L-Reg.} allows for motion in the degenerate direction, possibly due to a lower regularization parameter.
Finally, \ac{TSVD} and Zhang \textit{et al.} perform very similarly since these methods both manipulate the eigenvectors of the optimization through eigenvalues, either with \textit{k}-rank approximation or solution projection.
\begin{figure*}[t]
    \centering
    \includegraphics[width=1\linewidth]{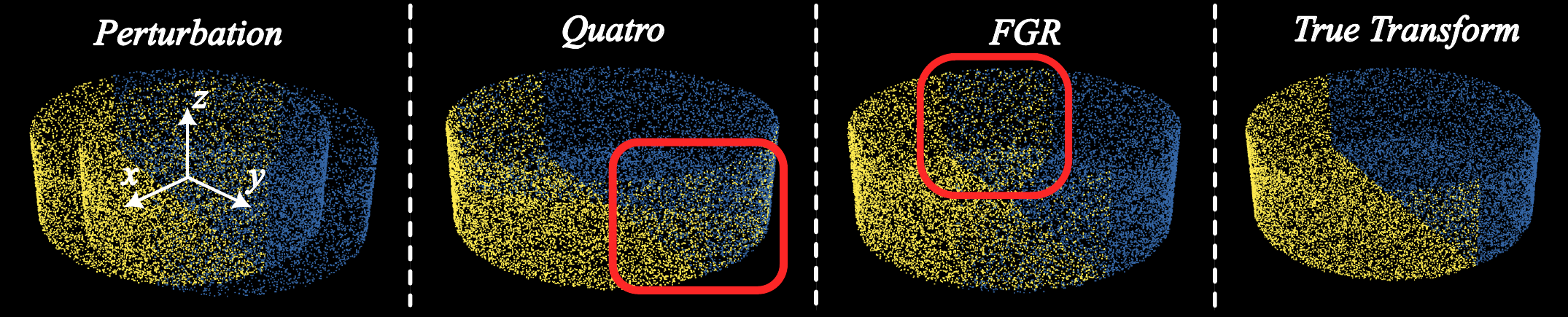}
    \caption{The results of Quatro~\cite{lim2022single} and FGR~\cite{fgr} are shown with the resulting transformation. The mismatch of the source and reference clouds is highlighted in red. The point clouds are colored to visualize the final transformation, with the ideal outcome depicted on the \textbf{right}. The perturbation is $\boldsymbol{t} = [0.2, 1.5, -0.25]$\SI{}{\meter} for translation and $\boldsymbol{r} = [0, 0, -60.00]$\SI{}{\deg} for rotation.
    }
    \label{fig:global_simulation_result}
\end{figure*}
\subsubsection{STATIC SIMULATION - GLOBAL POINT CLOUD REGISTRATION}
\label{section:Experiments:global_sim}

In this experiment, the global point cloud registration methods Quatro~\cite{lim2022single} and FGR~\cite{fgr}, discussed in \Cref{sec:methods:passive:global}, are employed for a similar task as in~\Cref{section:Experiments:static_sim}. Moreover, similar to \Cref{section:Experiments:static_sim}, a 1-axis rotational degenerate point cloud is used, as shown in \Cref{fig:global_simulation_result}, but with a different perturbation from the reference point cloud for rotation around the Y and X axes, since Quatro can only estimate the $4$-\ac{DoF} transforms. Multiple configurations have been employed for both methods to ensure objective comparison, and the best performing configuration is used in the generation of~\Cref{fig:global_simulation_result}.
The purpose of this experiment is to highlight the fact that the degenerate point cloud registration problem is not a correspondence outlier problem but, in fact, an absence of geometric information within correspondences, i.e. absence of true inliers problem. 
Specifically, this experiment underlines the importance of degeneracy detection and mitigation and shows that in the absence of inliners, even globally optimal point cloud registration methods are unable to retrieve the true transformation.
As this is an absence of data problem, the expected output from these methods is not the recovery of the true transform but, in fact, mitigation of further adverse effects of the degeneracy.

As shown in~\Cref{fig:global_simulation_result}, both methods recovered translation very well and, as expected, but failed to retrieve the rotation around the degenerate axis (i.e., rotation around the Z axis) correctly, as illustrated by the mismatched colors of the point clouds. This result is expected since the rotation around the degenerate axis is unobservable, and termination of the optimization after convergence of all well-conditioned directions is considered a successful optimization.
This experiment supports the notion that global registration methods are not \textit{drop-in} solutions to the problem of degenerate point cloud registration, as for the LiDAR point cloud this problem is inherently ill-defined, and there are many minima for the optimization solution. 
Many global registration methods are robust to outliers and can recover the true minima from noisy data \textit{given the presence of inliers}. The assumption of the presence of the inliers no longer holds for the problem of degenerate point cloud registration, and as a result, these methods are unable to provide additional benefits to the other methods described in~\Cref{sec:methods}.
%
\begin{figure*}[t!]
    \centering
    \includegraphics[width=0.95\linewidth]{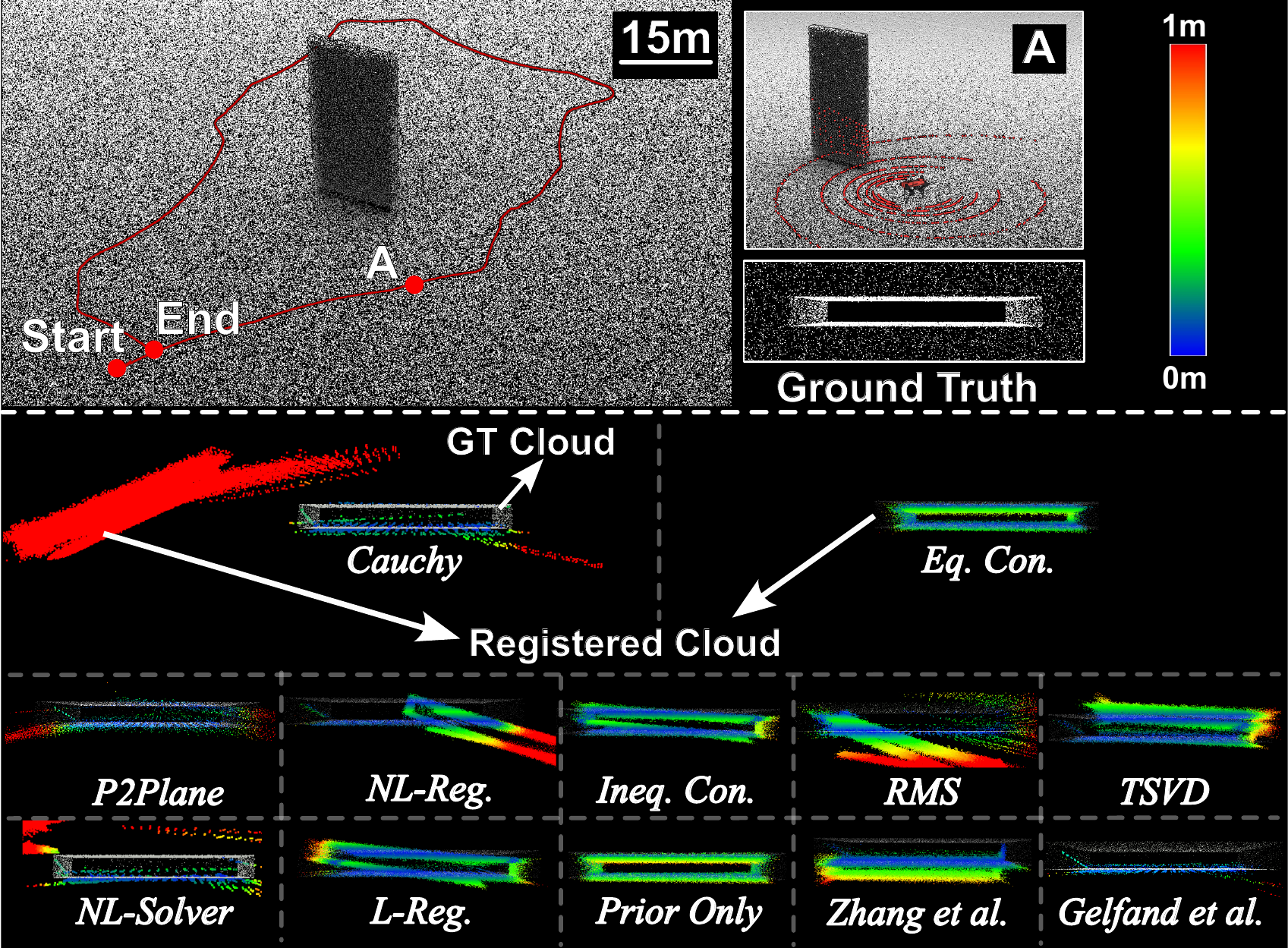}
    \caption{The mapping results of the dynamic ANYmal simulation experiment with \textit{Open3d SLAM} in the loop are shown. \textbf{Top:} The robot trajectory is overlaid with the simulated environment ground truth map. \textbf{Bottom:} The registered maps (with the ground removed) for each method are shown next to an error color bar. Points in the maps are colored according to the point-to-point distances to the ground truth map of the environment. Results of Cauchy and Eq. Con. methods are shown from a far-away perspective to illustrate the scale of the registration error. For visual clarity, the remaining methods are cropped around the pillar. }
    \label{fig:Experiments:dynamic_sim}
\end{figure*}

\subsubsection{ANYmal SIMULATION}
\label{section:Experiments:dynamic_sim}

In this experiment, a simulated ANYmal robot equipped with a Velodyne~\mbox{VLP-16} LiDAR is simulated to navigate within a 3-axis LiDAR degenerate environment as shown in~\Cref{fig:Experiments:dynamic_sim}.
The robot traverses around a rectangular pillar on a planar surface, challenged with degeneracy in translation on the ground plane and rotation around the ground plane normal.
During the experiment, the body velocities of the robot reach \SI{0.85}{\metre\per\second} and \SI{40}{\degree\per\second}, adding an additional layer of difficulty.\looseness-1

To leverage the simulated environment and analyze the robustness of each algorithm, a uniformly distributed noise is added to the otherwise perfect external pose prior, which is used as an initial guess for the point cloud registration process in all methods. 
The noise is sampled from normal distributions $\mathcal{N}(\mu_t,\,{\sigma_t}^{2})$ and $\mathcal{N}(\mu_r,\,{\sigma_r}^{2})$ where the distribution variables are $\mu_t=\SI{0}{\cm}, \ \sigma_t=\SI{0.05}{\meter}$, $\mu_r=\SI{0}{\radian}, \ \sigma_r=\SI{0.01}{\radian}$. 
The mapping results for each method and the ground truth point cloud of the pillar are shown in~\Cref{fig:Experiments:dynamic_sim}. To highlight the degradation of the mapping quality due to \ac{LiDAR} degeneracy, the ground plane is cut from the results of each method. 
As seen in~\Cref{fig:Experiments:dynamic_sim}-A, the sparsity of the \ac{LiDAR} scan and the size of the environment create another layer of complexity since only a limited number of LiDAR scanlines fall onto the pillar.
Uniquely, in this experiment, the robot starts in a spot where the LiDAR is already degenerate; hence, degenerate scan-to-submap registration can be thought of as degenerate scan-to-scan registration at the beginning of the sequence. 
This further challenges the optimization process as the sensitivity to outliers increases since the number of correspondences is limited at this stage.

The mapping results shown in~\Cref{fig:Experiments:dynamic_sim} indicate a good performance of \ac{Eq. Con.} and Prior Only methods.
As expected, a misaligned map was generated by the baseline methods \ac{P2Plane}, NL-Solver, and the Cauchy method. This is due to the severe LiDAR degeneracy introduced by the experiment setup, where the robot starts and stops the simulated experiment in the presence of multi-directional LiDAR degeneracy. In addition, the sampling methods RMS method of \cite{petracek2024rms} and Gelfand \textit{et al.} also fail to register the point clouds correctly. The RMS method is described as not resilient against high-rate rotations~\cite{petracek2024rms}, providing a reason for the observed behavior. Besides, the limited number of features in the environment affects the efficacy of the sampling methods.
Morever, the \ac{TSVD}, \ac{L-Reg.} and \ac{Ineq. Con.} methods generate a similar map, showing a moderate amount of drift in rotation and translation.
The drift of \ac{L-Reg.} and \ac{Ineq. Con.} can be attained to the usage of parameters tuned for the Ulmberg tunnel experiment. The environment based tuning might improve the performance of these methods. Performance of \ac{TSVD} might indicate the importance of mitigating the degeneracy during or after the optimization as this is the only active degeneracy mitigation method acting before the optimization.\looseness-1
Lastly, the \ac{NL-Reg.} method performs worse than the other active degeneracy mitigation methods but better than the passive counterparts; similar reasoning with parameter selection might explain the performance.

\begin{table}[ht]\centering
    \caption{APE and RPE (per \SI{1}{\meter} distance traveled) metrics are provided for the dynamic ANYmal simulation experiment where the best is in \textbf{bold} and the second best is \underline{underlined}.}
    \resizebox{1\columnwidth}{!}{
    \begin{tabular}{lcccc}
    \toprule[1pt]
    \multirow{2}[3]{*}{} & 
    \multicolumn{2}{c}{\makecell{APE}} & \multicolumn{2}{c}{RPE} \\
    \cmidrule(lr){2-3}  \cmidrule(lr){4-5}&
    \makecell{Translation \\ $\mu(\sigma)[m]$ } & \makecell{Rotation \\ $\mu(\sigma)[\deg]$} & \makecell{Translation \\ $\mu(\sigma)[m]$ } & 
    \makecell{Rotation \\ $\mu(\sigma)[\deg]$} \\[0.3em]
    \toprule[1pt]
    \makecell{Eq. Con.} &
    \underline{0.083}(\underline{0.055}) & 
    \textbf{0.374}(\underline{0.206}) & 
    \underline{0.022}(\underline{0.018}) & 
    \textbf{0.362}(\textbf{0.241}) \\[0.3em]
    \makecell{InEq. Con.} &
    {0.651}({0.665}) & 
    {1.900}({1.610}) & 
    {0.046}({0.032}) & 
    {0.494}({0.286}) \\[0.3em]
    \makecell{Zhang \textit{et al.}} &
    {0.658}({0.347}) & 
    \underline{0.521}({0.325}) & 
    {0.041}({0.040}) & 
    {0.428}({0.329}) \\[0.3em]
    \makecell{TSVD} &
    {2.05}({1.23}) & 
    {1.224}({0.642}) & 
    {0.039}({0.035}) & 
    {0.439}({0.289}) \\[0.3em]
    \makecell{P2Plane} &
    {2.485}({1.299}) & 
    {8.31}({635}) & 
    {0.155}({0.209}) & 
    {0.940}({1.451}) \\[0.3em]
    \makecell{Prior Only} &
    \textbf{0.018}(\textbf{0.009}) & 
    {0.583}(\textbf{0.201}) & 
    \textbf{0.017}(\textbf{0.007}) & 
    {0.77}({0.295}) \\[0.3em]
    \makecell{NL-Reg.} &
    {1.169}({0.918}) & 
    {3.811}({2.502}) & 
    {0.046}({0.055}) & 
    {0.471}({0.886}) \\[0.3em]
    \makecell{NL-Solver} &
    {2.832}({1.71}) & 
    {10.43}({4.696}) & 
    {0.133}({0.185}) & 
    {1.140}({3.255}) \\[0.3em]
    \makecell{L-Reg.} &
    {1.105}({1.114}) & 
    {3.127}({2.434}) & 
    {0.043}({0.042}) & 
    {0.586}({0.801}) \\[0.3em]
    \makecell{Cauchy} &
    {1.210}({0.802}) & 
    {4.627}({4.196}) & 
    {0.077}({0.132}) & 
    {0.732}({1.757}) \\[0.3em]
    \makecell{RMS} &
    {2.863}({1.423}) & 
    {9.855}({4.01}) & 
    {0.13}({0.197}) & 
    {1.042}({2.347}) \\[0.3em]
    \makecell{Gelfand \textit{et al.}} &
    {4.273}({3.24}) & 
    {14.30}({10.24}) & 
    {0.173}({0.222}) & 
    {1.495}({3.60}) \\[0.3em]

    \toprule[1pt]
    \end{tabular}
    }
    \label{table:numeric_results_dynamic_sim}
\end{table}

The quantitative comparison of methods is presented in \Cref{table:numeric_results_dynamic_sim} as relative and absolute error \mbox{metrics~\cite{ate_rpe}} and calculated using the EVO evaluation package\footnote{\url{https://github.com/MichaelGrupp/evo}\label{foot:evo}}.
The metrics show a comparable performance between \ac{Eq. Con.}, Prior Only and Zhang \textit{et al.} with a lead from \ac{Eq. Con.} method.
Since the Prior Only method utilizes the noisy pose prior in the presence of degeneracy, the observed errors originate from the numeric errors of registration and mapping and the added motion prior noise.
Interestingly, none of the tunable methods allowing for constraint relaxation, \ac{Ineq. Con.}, \ac{L-Reg.} and \ac{NL-Reg.} perform comparably to the best-performing method.
Intuitively, constraint relaxation indicates reliance more on the point cloud registration rather than the external pose prior, and since the experiment consists of multiple snippets of severe degeneracy, relying on the registration does not result in better performance.
Instead, methods that do not allow for constraint relaxation perform better. 
On the other hand, passive degeneracy mitigation methods and baselines show LiDAR slip and incorrectly registered maps.\looseness-1

\subsection{ANYmal FOREST EXPERIMENT: OPEN FIELD DEGENERACY} \label{section:Experiments:forest}

During this experiment, the ANYmal robot, shown in~\Cref{fig:Experiments:deployments}-A, starts next to a forest and navigates to an open field, does rapid rotations, and returns to the same passage it previously observed after walking for \SI{107}{\meter}.
In the open field, shown in~\Cref{fig:Experiments:deployments}-B, the optimization is expected to have 3-axes degeneracy. That is, XY translations along the ground plane and rotation around the normal of this plane. 
However, as the robot is close to a forest region, the tree canopy provides unstructured yet sufficient geometric information for the optimization to converge. 
As a result, \ac{LiDAR} degeneracy only occurs when the \ac{LiDAR} loses line of sight to the forest canopy during the experiment.
In this experiment, the ANYmal's kinematic leg odometry estimator (TSIF)~\cite{tsif} propagates the previous scan-to-submap registered pose along the degenerate direction.
The ground truth trajectory is acquired through \ac{GNSS} positioning and used to analyze the global consistency of the estimated robot translation, as the \ac{GNSS} measurements only provide position information. 
Moreover, the ground truth point cloud of the region of interest is collected by the hand-held Leica BLK2GO~\cite{blk2go} scanner. 
Interestingly, even this commercial scanner failed to register in the degenerate open field; hence, only the points around the re-visited area are kept for error metrics computation.

The mapping results of this experiment and the ground truth map of the re-visited are provided in~\Cref{fig:forest_mapping_results}.
Qualitatively, the methods \ac{NL-Reg.}, \ac{Ineq. Con.}, \ac{Eq. Con.}, \ac{TSVD} and NL-Solver produced a good map, inferred by inspecting the man-made bench present in the environment.
All these methods, except NL-Solver, perform active degeneracy mitigation and act on degeneracy before (\ac{TSVD}) or during the optimization (\ac{NL-Reg.}, \ac{Ineq. Con.}, \ac{Eq. Con.}). NL-Solver performs well without awareness of degeneracy, which supports the notion that the degeneracy in this experiment is mild and can be recovered from. Moreover, the sampling methods RMS and Gelfand \textit{et al.} did not register the bench correctly. The main reason is the presence of large amounts of leaves and vegetation, which hinder the correct calculations of surface normals due to their unstructured nature.
It is important to note that the methods \ac{NL-Reg.} and \ac{Ineq. Con.} have been fine-tuned for the Ulmberg bicycle tunnel experiment, yet have also performed adequately in this experiment.
\begin{table}[h]\centering
    \caption{Error metrics and ICP-loop computational time for the ANYmal forest experiment, best in \textbf{bold} and the second best is \underline{underlined}.}
    \resizebox{1\columnwidth}{!}{
    \begin{tabular}{lccc}
    \toprule[1pt]
    & \makecell{RTE \\ $\mu(\sigma)[m]$ } & \makecell{ATE \\ $\mu(\sigma)[m]$ } & \makecell{Computational \\ Cost $\mu(\sigma)[ms]$} \\
    \toprule[1pt]
    \makecell{Eq. Con.}     & 0.049(0.048)          &0.490(0.403)              &9.67(3.71) \\[0.3em]
    \makecell{Ineq. Con.}   & 0.045(\textbf{0.031}) &0.405(0.302)              &11.92(4.29)\\[0.3em]
    \makecell{Zhang \textit{et al.}} & 0.083(0.094)          &0.372(\textbf{0.224})     &10.45(4.76)\\[0.3em]
    \makecell{TSVD}         & 0.068(0.066)          &0.542(0.423)               &{10.43}(4.83)\\[0.3em]
    \makecell{P2plane}      & 0.212(0.281)          &2.62(1.99)                 &{11.45}(6.53) \\[0.3em]
    \makecell{Prior Only}   & \underline{0.044}(0.045)         &0.662(0.436)               &{9.02}(5.39) \\[0.3em]
    \makecell{NL-Reg}       & \textbf{0.039}(\underline{0.037}) &\underline{0.364}(0.290)      &{{59.86}(30.20)}  \\[0.3em]
    \makecell{NL-Solver}    &0.048(0.047)          &\textbf{0.342}(\underline{0.263})               &45.32(19.93) \\[0.3em]
    \makecell{L-Reg}        & 0.062(0.086)          &0.843(0.517)               &{11.91}(4.93)  \\[0.3em]
    \makecell{Cauchy}       & 0.051(0.086)          &0.882(0.738)               &\underline{8.82}(\textbf{3.98}) \\[0.3em]
    \makecell{RMS}          & 0.362(0.415)          &3.70(3.48)               &\textbf{8.28}(\underline{4.22}) \\[0.3em]
    \makecell{Gelfand \textit{et al.}}          &0.258(0.269)          &2.61(2.44)               &15.20(6.42) \\[0.3em]
    \toprule[1pt]
    \end{tabular}
    \label{table:RTE_ATE_results_anymal_forest}
    }
\end{table}
Out of the methods that do not require parameter tuning, \ac{Eq. Con.} and \ac{TSVD} perform better, possibly thanks to the degeneracy mitigation before or during the optimization.
As the P2plane and Cauchy methods are blind to LiDAR degeneracy, the map generated by these methods is broken, as seen by the duplication.
The Prior Only method relies on the odometry prior, and since the leg odometry estimator does not perform well on soft terrain, the generated map is also misaligned.
Differently, Zhang \textit{et al.} and \ac{L-Reg.} methods, despite actively mitigating degeneracy, show LiDAR drift in the generated maps.
The good performance of Zhang \textit{et al.} might be explained by the difference in the way the method addresses degeneracy, since this is the only method to act on optimization degeneracy after optimization. This method performs among the best until the $t=\SI{180}{\unit{\second}}$ mark, where the payload of the robot blocks the view of the unique features, leading to LiDAR drift.
Moreover, the performance of \ac{L-Reg.} can be explained by a sensitivity to parameter tuning, as the performance of this method relies on the value of the degeneracy regularization parameter.
\begin{figure*}[t!]
\centering
    \includegraphics[width=1\linewidth, ]{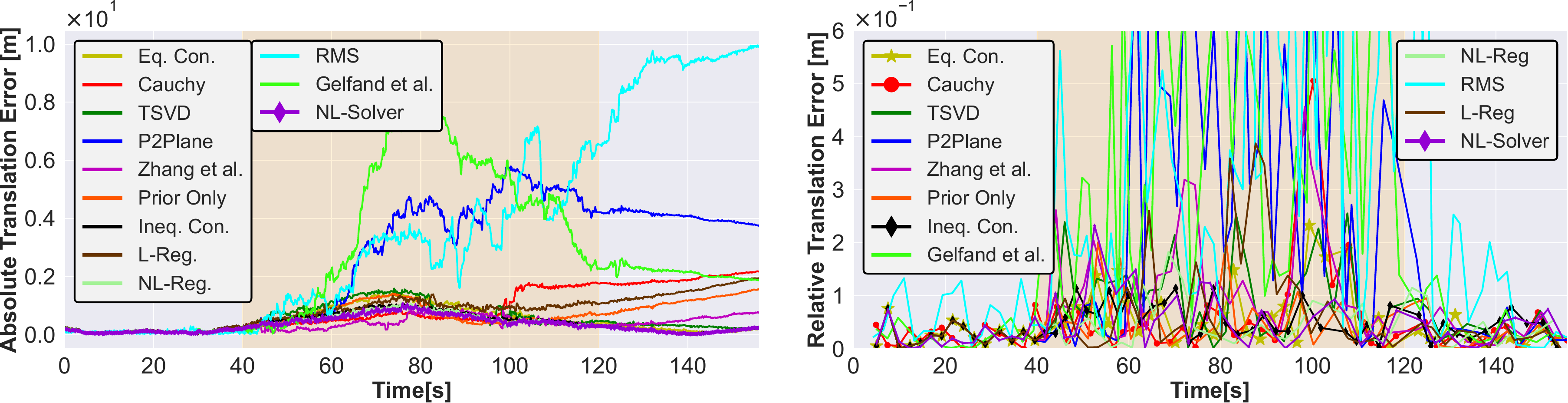}
    \caption{\ac{ATE} (left) and \ac{RTE} (right) over time of each method for the duration of the ANYmal forest experiment. The approximate LiDAR degenerate region is highlighted in orange. For \ac{RTE}, the per \SI{1}{\meter} traveled distance error increases on average in the degenerate region.}
    
    \label{fig:Experiments:ape_rpe_forest_analytics}
\end{figure*}
\ac{RTE} per \SI{2}{\meter} traversed distance and \ac{ATE} metrics are provided along with the ICP computational cost in~\Cref{table:RTE_ATE_results_anymal_forest}.
Quantitatively, the best method in terms of \ac{RTE} and \ac{ATE} mean error is \ac{NL-Reg.} followed by \ac{Ineq. Con.} in RTE and Zhang \textit{et al.} in \ac{ATE}.
It should be noted that the good performance of Zhang \textit{et al.} is shown in statistical metrics, but the misaligned map shown in~\Cref{fig:forest_mapping_results} indicates a sub-optimal estimation in rotation as the provided metrics only measure the translation errors.
To ensure an in-depth understanding of the performance differences of the compared results, the \ac{RTE} and \ac{ATE} errors of each method are also provided throughout the robot trajectory as shown in~\Cref{fig:Experiments:ape_rpe_forest_analytics} and \Cref{fig:Experiments:ape_rpe_forest_analytics}.
In both plots, at time $t=100$\SI{}{\second}, multiple methods' performance considerably changes.
This point in time corresponds to the 3-axes degeneracy, including the rotation around the ground plane normal.
The \ac{ATE} performance of Cauchy and Zhang \textit{et al.} is better than that of the other methods until the degeneracy at time $t=100$\SI{}{\second} is reached.
This suggests that these methods might be sensitive to the severity of the degeneracy.\looseness-1

Regarding computational cost, all methods, except the best performing \ac{NL-Reg.} and \ac{NL-Solver} methods, behave similarly to the baseline and Prior Only taking the lead, as this method skips the registration step. 
However, \ac{NL-Reg.} and \ac{NL-Solver} are approximately 5 times more computationally expensive as the underlying non-linear optimization takes more time to solve despite the same degeneracy constraints provided as with other methods. Moreover, the fastest method is RMS by \cite{petracek2024rms}; this result is expected as RMS samples, partially to reduce the redundancy in the point cloud. However, ATE and RTE suggest that this method might not be suitable for unstructured natural environments where rapid rotations are performed. The RTE error plot also suggests unstable behavior without LiDAR degeneracy, which supports the challenges of the environment in this method.\looseness-1

\begin{figure}[t!]
\centering
    \includegraphics[width=1\linewidth, ]{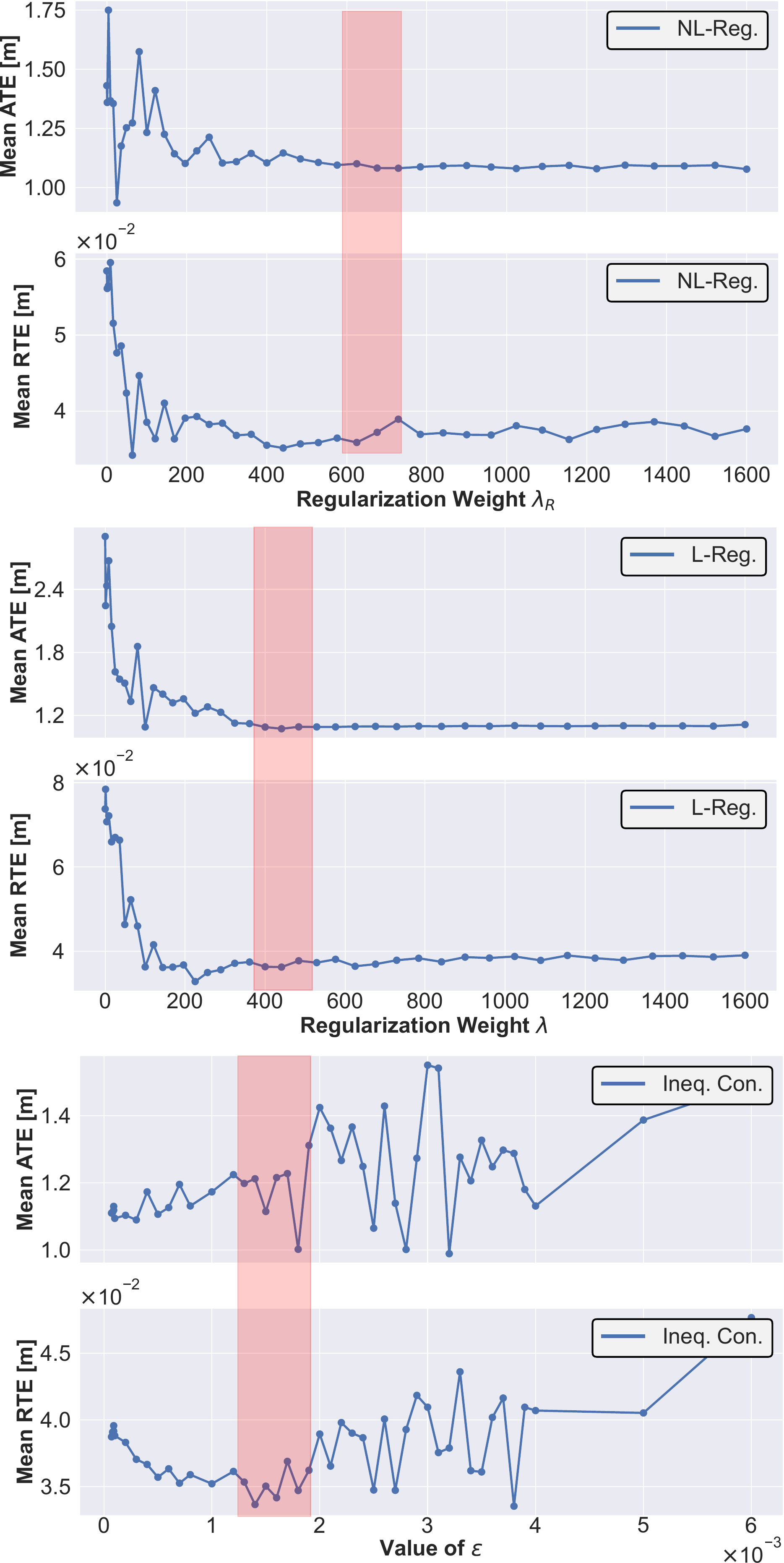}
    \caption{ The effect of parameter tuning is illustrated for \ac{NL-Reg.} (Top), \ac{L-Reg.} (Middle) and \ac{Ineq. Con.} (Bottom) methods for the Ulmberg Bicycle tunnel experiment. Selected regularization weight and $\epsilon$ values with low errors are highlighted in red for each method. }
    
    \label{fig:weights}
\end{figure}

\subsection{ULMBERG BICYCLE TUNNEL EXPERIMENT: ONE DIRECTIONAL DEGENERACY}\label{section:Experiments:tunnel}

One of the common degenerate cases for \ac{LiDAR}-SLAM systems is the one-directional degeneracy, typical in corridor or tunnel-like structures. 
Ulmberg bicycle tunnel dataset~\cite{pfreundschuh2023coin}, depicted in~\Cref{fig:Experiments:deployments}-D, represents an example of such an environment. 
The handheld sensor payload used in this experiment, shown in~\Cref{fig:Experiments:deployments}-C, is equipped with an Ouster OS0-128 \ac{LiDAR}.
Importantly, as the handheld payload does not have kinematic measurements that can be used to provide a pose prion, the intensity-based odometry estimates from COIN-LIO~\cite{pfreundschuh2023coin} are used to propagate the previous scan-to-submap registered pose during optimization degeneracy for all methods. COIN-LIO can successfully localize in such environments since it utilizes the visual cues of the environment in the intensity spectrum and does not rely on the geometric features of the degraded environment.

\subsubsection{PARAMETER TUNING}
\label{section:parameter_tuning}
The methods \ac{Ineq. Con.}, \ac{L-Reg.}, and \ac{NL-Reg.} are sensitive to parameter selection. To address this problem, the performance of these methods is analyzed using the data from this experiment.
\Cref{fig:weights} shows the effect of the upper and lower bound selection for the \ac{Ineq. Con.} method on the \ac{RTE} and \ac{ATE} metrics. The RTE metric is the measure of local consistency and stability of the method; on the other hand, the ATE metric measures the absolute error and is the measure of consistency over longer horizons.
When the bounds are numerically close to 0, the \ac{RTE} error increases, indicating local errors; however, the decreasing \ac{ATE} suggests that the constraint's increased tightness helps prevent large-scale drift.
Consequently, the bound parameter is set to $\epsilon = 0.0014$, indicating a lower bound of $-\epsilon$ and an upper bound of $\epsilon$. This value is selected as it falls to the saddle region where the error metrics are, on average, low.

Similarly, the effect of the regularization parameter for the methods \ac{L-Reg.}, and \ac{NL-Reg.} are provided in \Cref{fig:weights}, respectively, in relation to the \ac{RTE} and \ac{ATE} errors.
Interestingly, as the regularization weight increases, both methods reach a minimum error region for the \ac{RTE} error; however, further increments increase the \ac{RTE} error. On the other hand, increments in the regularization weight generally decrease the \ac{ATE} error.
Based on this analysis, $\lambda=440$ for \ac{L-Reg.} and $\lambda_{R}=675$ for \ac{NL-Reg.} are selected and used for all the experiments in this work.

Intuitively, the impact of the regularization is positively related to the amplitude of the regularization weight. Consequently, as the weight increases, these methods approximate the behavior of hard-constrained methods. Since the strength of regularization-based methods originates from the delicate balance between degeneracy-aware regularization and the point-to-plane cost function, as a consequence, the first regularization parameter that results in low RTE and ATE is selected.\looseness-1
%
\begin{figure*}[t]
    \centering
    \includegraphics[width=1\linewidth]{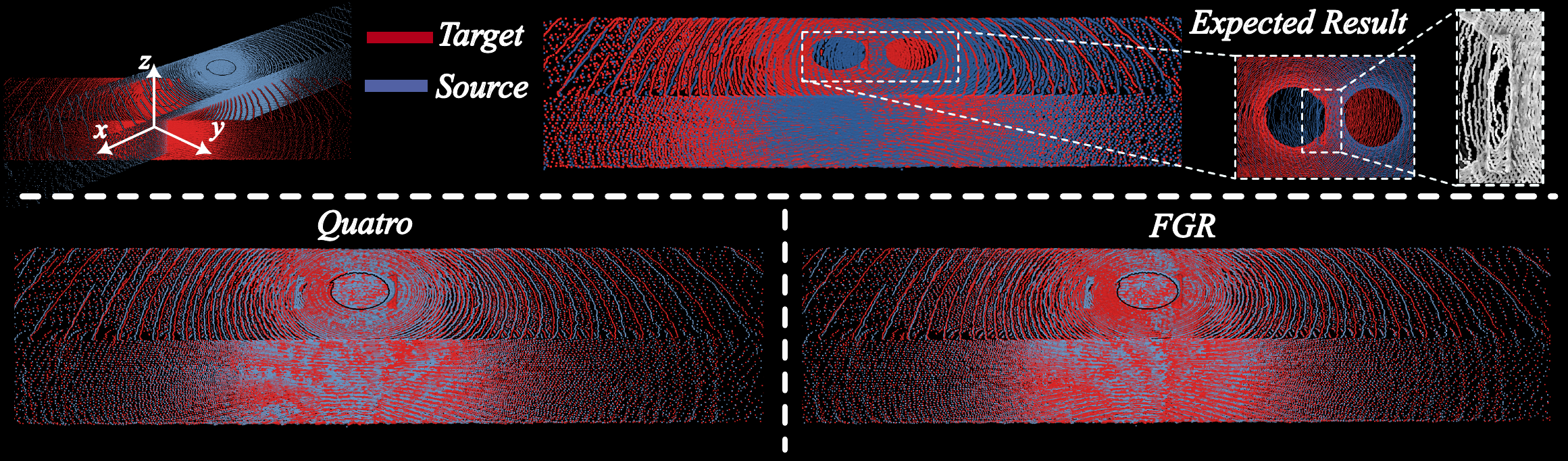}
    \caption{ \textbf{Top-left:} The point cloud registration setup is provided with the perturbed reference cloud. \textbf{Top-right:} The expected registration result is shown alongside the highlighted light housing of the tunnel; for better visualization of the edges, the points are in shaded gray. \textbf{Bottom:} The registration results from the Quatro~\cite{lim2022single, lim2024quatro} and FGR~\cite{fgr} methods are shown.  The perturbation is $\boldsymbol{t} = [-1, -1, -0.5]$\SI{}{\meter} for translation and $\boldsymbol{r} = [0, 0, -30.00]$\SI{}{\deg} for rotation.}
    \label{fig:tunnel_global}
\end{figure*}

\subsubsection{GLOBAL POINT CLOUD REGISTRATION}
A standalone registration analysis is performed to evaluate the feasibility of global registration methods in real-world LiDAR degenerate scans.
The registration setup consists of two point clouds extracted from the experiment data while maintaining approximately \SI{70}{\percent} overlap between the scans.
As shown in the upper row of~\Cref{fig:tunnel_global}, the source point cloud is transformed and provided as input along with the reference cloud to the global registration methods, Quatro~\cite{lim2022single} and FGR~\cite{fgr}.
Compared to the experiment described in \Cref{section:Experiments:global_sim}, where only a rotational direction is degenerate, in this experiment, only a translational direction is degenerate.
As seen in the bottom row of~\Cref{fig:tunnel_global}, both methods were unable to find the true global minima in the absence of distinguishable geometric features. However, both methods successfully retrieved the correct transform for all well-constrained axes, illustrating a correct usage of the method.
Furthermore, for this problem setup, the empty spots in the scan act as false global minima, which the methods cannot distinguish from the true global minima, which can be inferred from the roof light structures illustrated in the upper-right part of \Cref{fig:tunnel_global}.
This degenerate scan-to-scan point cloud registration experiment furthermore strengthens the notion that, in the absence of observable inliers, global point cloud registration methods also suffer from the problem of LiDAR degeneracy and might not bring additional benefits compared to other approaches.

\subsubsection{SLAM IN THE LOOP REGISTRATION}

\begin{figure*}[t]
\centering
    \includegraphics[width=1\linewidth]{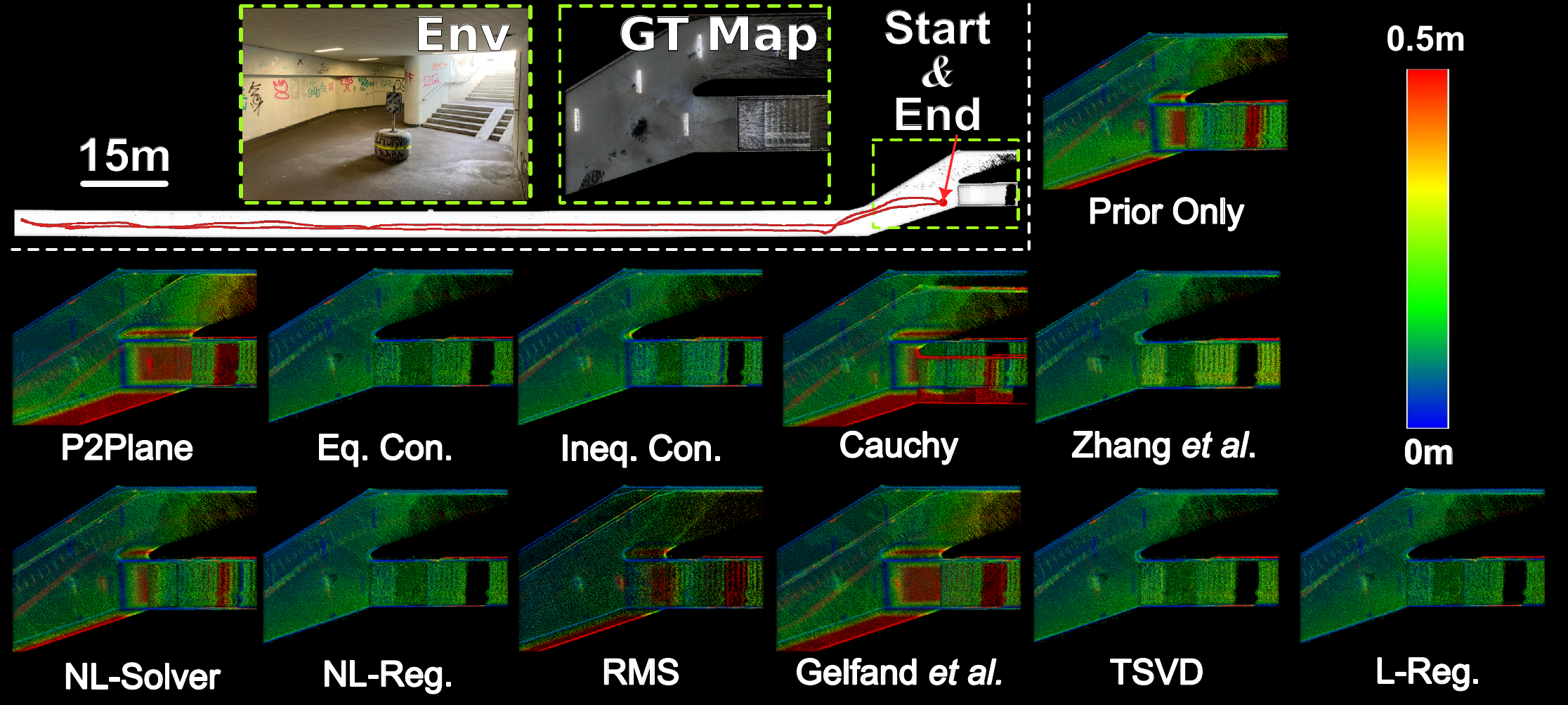}
    \caption{The mapping results from the Ulmberg bicycle tunnel experiment are shown. \textbf{Top-left:} The ground truth map of the tunnel and the real-world image of the region of interest are shown next to the point cloud map of the Eq. Con. method. \textbf{Rest:} The registered maps for each method are shown next to an error color bar. Points in the maps are colored according to the point-to-point distances to the ground truth map of the environment.}
    \label{fig:experiments:tunnel_results}
\end{figure*}

For each method, the point cloud map of the twice-observed (upon return) stairs region is provided in~\Cref{fig:experiments:tunnel_results}. 
Moreover, the ground truth map of the non-degenerate section, collected again with the vision-aided Leica BLK2GO scanner~\cite{blk2go}, and the real-life picture of the environment are provided in~\Cref{fig:experiments:tunnel_results}.
The mapping results show the effectiveness of active degeneracy mitigation methods, as all methods of this category generate a point cloud map with minimal or no drift. 

\begin{figure*}[t]
\centering
    \includegraphics[width=1\linewidth, ]{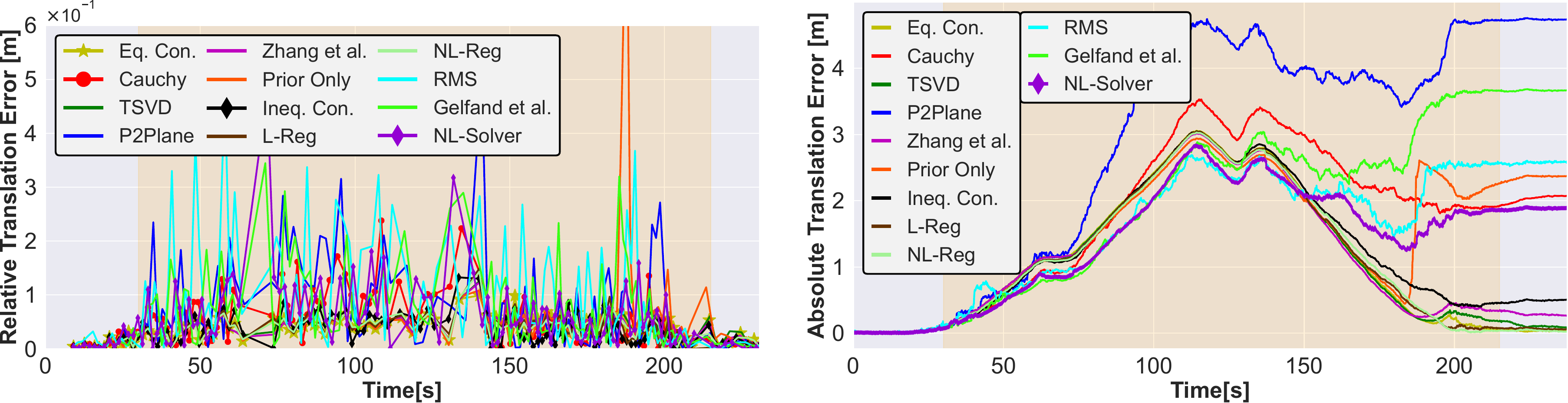}
    \caption{The \ac{RTE} (left) and \ac{ATE} (right) for each method is provided over the duration of the experiment. The LiDAR degenerate region is highlighted in orange. For \ac{RTE}, the active degeneracy mitigation methods consistently prevent higher errors. For \ac{ATE}, the common increase in error at the middle of the trajectory is mainly due to the bending of the map originating from the errors in surface normal extraction. }
    
    \label{fig:combined_ape_rpe_tunnel}
\end{figure*}

Importantly, \ac{Ineq. Con.}, \ac{L-Reg.}, and \ac{NL-Reg.} methods are tuned for this environment; hence, these methods are expected to perform well. The P2Plane and Prior Only methods show map duplication along the direction of the tunnel, and the Cauchy method shows the most drift among all methods. Moreover, both sampling methods, the RMS and Gelfand \textit{et al.}, improve their performance compared to the baseline P2Plane; however, they perform worse compared to the non-linear baseline NL-Solver. The underlying reason is the severity and duration of the degeneracy in this experiment which, limits the performance of sampling based methods. The performance difference between the two baseline methods might suggest the effectiveness of the non-linear solver in mitigating the adverse effect of an ill-conditioned Hessian.
The change in \ac{RTE} per \SI{1}{\meter} traversed distance and \ac{ATE} are shown in \Cref{fig:combined_ape_rpe_tunnel} and \Cref{fig:combined_ape_rpe_tunnel}, respectively. As seen in \Cref{fig:combined_ape_rpe_tunnel}, the active degeneracy mitigation methods have a lower error variation compared to the baseline P2Plane and Cauchy methods. In comparison, the Prior Only method has a bad estimate peak around $t=\SI{180}{\unit{\second}}$, which is sufficient to generate a misaligned map. Interestingly, a comparison of \ac{Ineq. Con.} and \ac{NL-Reg.} reveals that despite both methods performing well, the error variation in \ac{Ineq. Con.} is higher, resulting in a duplication of the map at the end of the trajectory. This result might indicate that stability and robustness are preferable to better local accuracy for large-scale mapping.\looseness-1

On the other hand, as seen in \Cref{fig:combined_ape_rpe_tunnel}, all methods accumulate errors over the trajectory of the platform. The source of this error is the aleatoric uncertainty of the data, and the epistemic uncertainty of \textit{Open3D-SLAM}~\cite{jelavic2022open3d} as the map curves throughout the trajectory due to the high incident angle of the observed points. 
\begin{table}[b!]\centering
    \caption{Error metrics and ICP registration computational time for the Ulmberg bicycle tunnel experiment are shown (best in \textbf{bold}), and the second best is \underline{underlined}.}
    \resizebox{1\columnwidth}{!}{
    \begin{tabular}{lccc}
    \toprule[1pt]
    & \makecell{RTE \\ $\mu(\sigma)[m]$ } & \makecell{ATE \\ $\mu(\sigma)[m]$ } & \makecell{Computational \\ Cost $\mu(\sigma)[ms]$} \\
    \toprule[1pt]
    \makecell{Eq. Con.} &0.039(\textbf{0.020})     & 1.10(1.04) &18.95(3.18) \\[0.3em]
    \makecell{Ineq. Con.} & \textbf{0.033}(0.026) & 1.21(0.97) &19.13(2.98)\\[0.3em]
    \makecell{Zhang \textit{et al.}} & 0.038(\underline{0.021})    & 1.125(\underline{0.983}) &18.90(3.03)\\[0.3em]
    \makecell{TSVD} & 0.039(0.021)    &1.11(1.01) &{18.95}(3.00)\\[0.3em]
    \makecell{P2plane} & 0.073(0.082)    &2.90(1.78) &{16.96}(3.50) \\[0.3em]
    \makecell{Prior Only} & 0.056(0.107)    &1.54(\textbf{0.947}) &\textbf{10.24}(9.28) \\[0.3em]
    \makecell{NL-Reg} &  \underline{0.035}(0.022)    &\textbf{1.097}(1.031) &{80.14}(29.04)  \\[0.3em]
    \makecell{NL-Solver} & 0.058(0.063)    &1.43(0.846) &64.70(17.05) \\[0.3em]
    \makecell{L-Reg} & 0.036(0.024)    &\underline{1.099}(1.030) &{22.17}(4.18)  \\[0.3em]
    \makecell{Cauchy} & 0.050(0.046)    &1.76(1.090) &18.33(3.52) \\[0.3em]
    \makecell{RMS} & 0.09(0.1)    &1.62(0.922) &\underline{10.942}(\textbf{1.06}) \\[0.3em]
    \makecell{Gelfand \textit{et al.}} & 0.073(0.075)    &1.97(1.28)      &24.321(4.54) \\[0.3em]
    \toprule[1pt]
    \end{tabular}\label{table:RTE_ATE_results_ulmberg_tunnel}
    }
\end{table}

%
It is interesting to observe that between $t=\SI{40}{\unit{\second}}$ and $t=\SI{80}{\unit{\second}}$, the Cauchy method outperforms all other methods until $t=\SI{90}{\unit{\second}}$, where it starts accumulating errors due to misregistration of the scans. Indicating that robust norms are powerful tools that might benefit accuracy; however, they still suffer from the absence of inliers. In addition, the sampling methods RMS, Gelfand \textit{et al.} and the non-linear baseline NL-Solver all outperform the active degeneracy methods until the severe degeneracy between $t=\SI{120}{\unit{\second}}$ and $t=\SI{150}{\unit{\second}}$ takes its effect. The intuition behind this behavior is possibly twofold: \textit{i)} active methods depend on motion prior quality and, \textit{ii)} sampling methods actively select scarce unique features.\looseness-1

Furthermore, the mean and standard deviation of these metrics are provided along with the ICP computational cost of the ICP registration in~\Cref{table:RTE_ATE_results_ulmberg_tunnel}. 
In a comparison of all methods, \ac{Ineq. Con.} methods produce the best mean error in \ac{RTE} while \ac{Eq. Con.} has the least variation. Among the not fine-tuned methods, Zhang \textit{et al.} performs the best in terms of mean RTE error while \ac{Eq. Con.} method has the least ATE error. 
Moreover, \ac{NL-Reg.} performs the best in terms of \ac{ATE} error, and Prior Only has the least variation. The performance is unsurprising as \ac{NL-Reg.} is fine-tuned to this environment. However, the computational cost of this method makes it less applicable to compute limited platforms. In addition, the RMS sampling method results in the most stable computational cost variation, thanks to its sampling strategy to reduce redundancy in the point cloud.
\begin{figure*}[t]
    \centering
    \includegraphics[ width=1\linewidth]{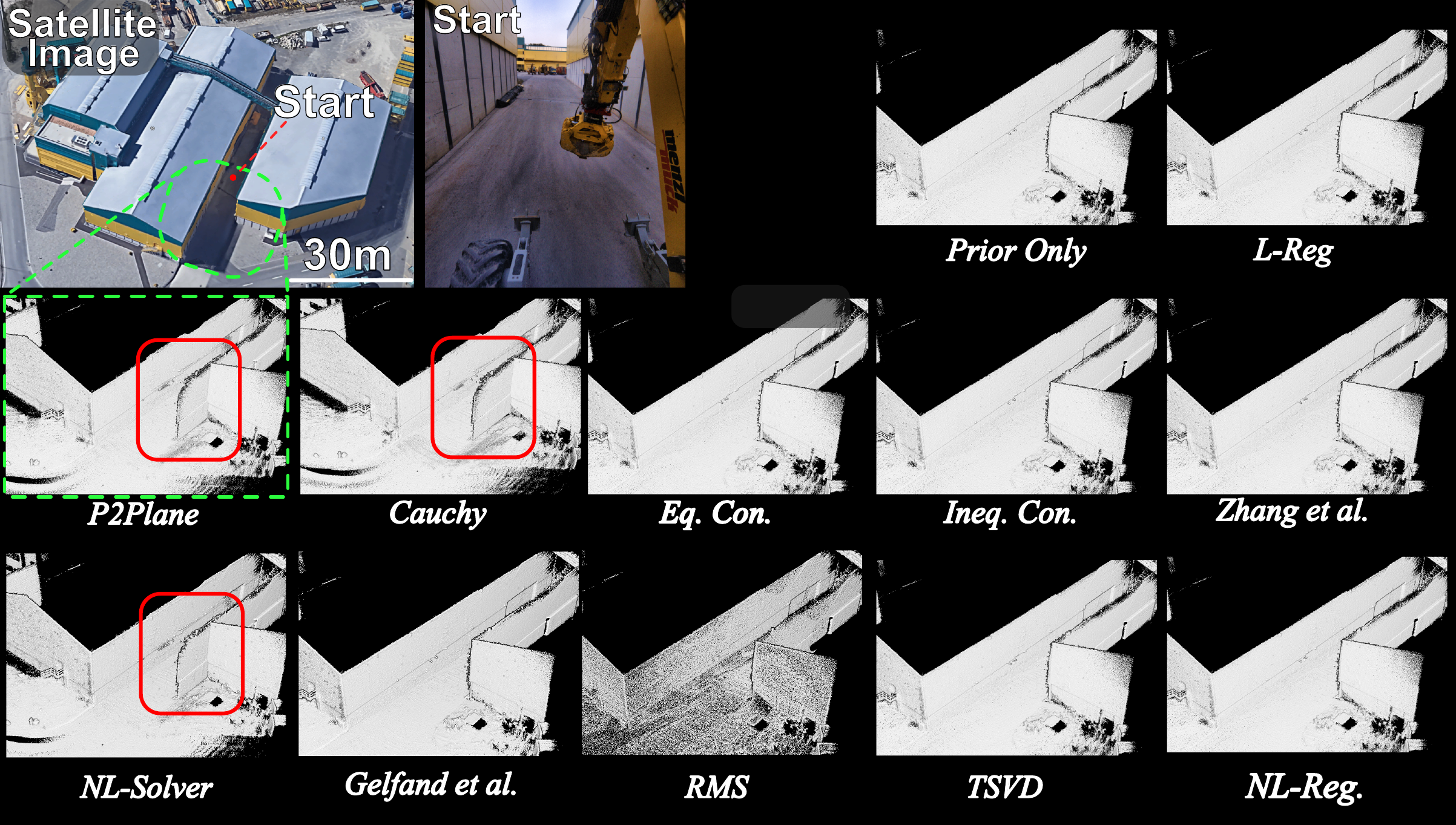}
    \caption{The mapping results of the excavator narrow-passage experiment are shown. \textbf{Top-left:} The experiment site's satellite imagery is shown alongside HEAP's start location. \textbf{Rest:} The maps from each method are shown. The adverse effect of the \ac{LiDAR} degeneracy is outlined in red for the \ac{P2Plane}, NL-Solver, and Cauchy methods. }
    \label{fig:excavator_results}
\end{figure*}
\subsection{HEAP EXCAVATOR EXPERIMENT: SHORT BURST DEGENERACY}
\label{section:Experiments:excavation}
In this experiment, a construction site in Oberglatt, Switzerland, is traversed by a total of \SI{170}{\meter} by the HEAP~\cite{heap} autonomous excavator. 
As previously shown in \cite{nubertGraph}, the environment contains a \ac{GNSS}-denied narrow passage between two buildings. 
HEAP and the construction environment are shown in~\Cref{fig:Experiments:deployments}.
The purpose of this experiment is to analyze the performance of the methods against short and severe LiDAR degeneracy while providing a good registration initial guess through COIN-LIO~\cite{pfreundschuh2023coin}.
As shown in \Cref{fig:Experiments:deployments}-F, only the between-building section of the environment is \ac{LiDAR} degenerate. However, the global consistency of the \ac{LiDAR} map is still compromised if the degeneracy is not mitigated.
The mapping result of this experiment is shown in~\Cref{fig:excavator_results}.
The results indicate that all active degeneracy prevention methods perform comparably, given a good initial guess, regardless of the severity of the degeneracy.
The accuracy of the registration initial guess can be inferred from the performance of the Prior Only method. 
Since the map from the Prior Only method is consistent with the real-world geometry, the odometry from COIN-LIO is inferred to be good.
On the other hand, the \ac{P2Plane}, {Cauchy} and NL-Solver methods show \ac{LiDAR} slip alongside the degenerate direction as these methods do not actively use the degeneracy information in the registration process.
Lastly, RMS and Gelfand \textit{et al.} sampling methods perform marginally better than the rest of the methods, which are good up to the accuracy of the motion prior. This improvement is due to the lines and corners present in the environment, which the sampling methods can actively identify and utilize.

\section{PRACTICAL MATTERS AND LESSONS LEARNED}
In this section, the authors provide insights and lessons learned on robust point cloud registration with a focus on field deployments in LiDAR degenerate environments.

\subsection{BEST PRACTICES FOR REGISTRATION IN FIELD ROBOTICS}
Often, field robotics implies a functional set of modules interacting with each other. 
These modules should be customized with respect to each other for the best performance, such as communication delays, sensor calibrations, and computation.
From these modules, the availability of an external odometry module (e.g. visual-inertial odometry~\cite{pfreundschuh2023coin}, kinematic-inertial odometry~\cite{tsif}, radar odometry~\cite{nissov2024robust}, etc.) is crucial, as the authors found that reliable external odometry prior is crucial to ensure the robustness of the robot pose estimate through a subsequent degenerate point cloud registration process. This has been illustrated in the experiments presented in~\Cref{section:Experiments:excavation}, as all degeneracy-aware methods performed equally given a good motion prior. Multi-modal tightly-coupled degeneracy-aware solutions could mitigate this need. However, this is a topic for future research.
Moreover, hardware and, subsequently, the deployed sensor setup play a crucial role. For example, a limited field of view of a LiDAR can create LiDAR degeneracy in one environment, while a different sensor placement successfully localizes in the same environment. This phenomenon is realized in~\Cref{section:Experiments:forest} as the VLP-16 LiDAR of ANYmal was partially blocked. Similarly, the HEAP excavator experiment described in~\Cref{section:Experiments:excavation} showed that a large field of view LiDAR, such as the Ouster OS0-128, is sufficient to create a degeneracy-aware solution to traverse in a self-similar corridor between two buildings. This is inferred through the performance of the sampling methods  Gelfand \textit{et al.} and RMS.

\subsection{IMPORTANCE OF THE SOFTWARE SETUP}
Different than the robot hardware configuration, the \ac{SLAM} backend and degeneracy-aware registration framework are similarly important.
Degeneracy mitigation alone is ineffective unless coupled with the correct degeneracy detection, as this facilitates degeneracy-aware point cloud registration.
The detection method should be robust and generalizable to changes in environments. Furthermore, excellent and reliable spatial feature extraction (cf. surface normals) is necessary for correct degeneracy detection.
Although this work does not focus on selecting the LiDAR degeneracy detection method, it is vital to mitigating its adverse effects.
Solutions such as robust norms, statistical filters, and semantic filters (e.g. RMS, Gelfand \textit{et al.}) can help mitigate noise in feature extraction. 
This is observed in~\Cref{section:Experiments:tunnel} and shown in~\Cref{fig:combined_ape_rpe_tunnel} where Cauchy, RMS and Gelfand \textit{et al.} performed better than other methods until errors generated from LiDAR degeneracy accumulated past a certain range.  

Besides the data and detection quality, the mapping back-end plays a critical role in robust scan-to-submap point cloud registration, as the submap acts as the reference point cloud and the system's understanding of the world. 
As discussed in~\Cref{section:Experiments:tunnel}~\Cref{fig:combined_ape_rpe_tunnel}, all active degeneracy prevention methods drift on the global scale, as common epistemic errors originating from the mapping back-end affect all the methods.
Finally, the computational complexity of the system must be real-time-capable, ideally matching the rate of the LiDAR sensor with minimal latency, to ensure usability for downstream tasks such as locomotion and navigation. Throughout~\Cref{section:Experiments}, the computational cost of \ac{NL-Reg.} is highlighted as, while accurate, this method would have to skip measurements in computationally limited systems.

\subsection{HOW TO MITIGATE LiDAR DEGENERACY}
Experiments in various real-world scenarios showed that active degeneracy mitigation methods perform more accurately and reliably than passive degeneracy methods in the presence of LiDAR degeneracy.
Among the active degeneracy mitigation methods, \ac{Eq. Con.}, \ac{TSVD} and \ac{NL-Reg.} demonstrate robust performance in all experiments, as shown in~\Cref{table:summary_table}, and show good mapping performance with minimal final \ac{ATE} error as shown in the results in~\Cref{section:Experiments}.
As a common factor among these methods, the constraints or the regularization terms are added \textit{before} or \textit{during} the optimization but not after.
%
\begin{table}[t]\centering
     \caption{Performance summary of the compared degeneracy mitigation methods. {\raisebox{-0pt}{{\large{\cmark}}}} indicates no visible drift,  {\raisebox{-2pt}{{\large\textbf{\sphere}}}} indicates minimal drift, and {\raisebox{-2pt}{{\large\xmark}}} indicates a broken point cloud map.} 
    \resizebox{1\columnwidth}{!}{
    \begin{tabular}{ccccc}
    \toprule[1pt]
    & \large{\makecell{ANYmal \\ Forest}} & \large{\makecell{Ulmberg \\ Tunnel}} & \large{\makecell{HEAP \\ Excavator}} & \large{\makecell{ANYmal \\ Simulation}} \\
    \toprule[1pt]\vspace{1mm}
    \large{\makecell{\underline{Hard-Constrained}}}\\[0.15em]
    \large{\makecell{Eq. Con.}} & \large\textbf{\sphere}   & \large\cmark             & \large\cmark & \large\cmark \\[0.3em]
    \large{\makecell{Zhang \textit{et al.}}} & \large\xmark  & \large\textbf{\sphere} & \large\cmark & \large\textbf{\sphere} \\[0.3em]
    \large{\makecell{TSVD}} & \large\textbf{\sphere}       & \large\textbf{\sphere} & \large\cmark & \large\textbf{\sphere} \\[0.3em]
        \cdashline{1-5}\\[-0.9em]
        \large{\makecell{\underline{Soft-Constrained}}}\\[0.15em]
    \large{\makecell{Ineq. Con.}} & \large\cmark             & \large\textbf{\sphere} & \large\cmark & \large\textbf{\sphere}\\[0.3em]
    \large{\makecell{NL-Reg}} &  \large\cmark                & \large\cmark             & \large\cmark & \large\xmark\\[0.3em]
    \large{\makecell{L-Reg}} & \large\xmark                  & \large\cmark             & \large\cmark & \large\textbf{\sphere}\\[0.3em]
        \cdashline{1-5}\\[-0.9em]
    \large{\makecell{\underline{Unconstrained}}}\\[0.15em]
    \large{\makecell{P2plane}} & \large\xmark                & \large\xmark             & \large\xmark & \large\xmark \\[0.3em]
    \large{\makecell{NL-Solver}} & \large\cmark                 & \large\xmark             & \large\xmark & \large\xmark \\[0.3em]
    \large{\makecell{Cauchy}} & \large\xmark                 & \large\xmark             & \large\xmark & \large\xmark \\[0.3em]
    \large{\makecell{\mbox{Prior Only}}} & \large\xmark             & \large\xmark             & \large\cmark & \large\cmark \\[0.3em]
    \large{\makecell{RMS}} & \large\xmark                 & \large\xmark             & \large\cmark & \large\xmark \\[0.3em]
    \large{\makecell{Gelfand \textit{et al.}}} & \large\xmark                 & \large\xmark             & \large\cmark & \large\xmark \\[0.3em]
    \toprule[1pt]
    \end{tabular}\label{table:summary_table}
    }
\end{table}
Furthermore, the heuristic parameter dependent \ac{L-Reg.} and \ac{Ineq. Con.} methods are found to be more sensitive to parameter tuning than \ac{NL-Reg.} method.
For example, the best parameters identified for the Ulmberg tunnel experiment do not necessarily perform similarly in other environments, as shown in~\Cref{table:summary_table}.\looseness-1
It is crucial to note that the conclusions drawn for these tunable methods are subject to the empirical parameter selection, as described in~\Cref{section:parameter_tuning}. An automatic and optimal selection of these parameters is subject to future research. Between different constraint types, it is hard to designate one method better than the other. \ac{Eq. Con.} (equality constraints) method is more consistent in all environments while \ac{Ineq. Con.} (inequality constraints) can be more accurate, albeit with lower estimation consistency. 
The \ac{TSVD} method provides a good alternative solution due to the ease of implementation, minimal computation overhead, and overall consistent performance across all environments with minimal error, as seen in~\Cref{table:summary_table}.
\begin{figure}[h!]
    \centering
    \includegraphics[width=1\linewidth]{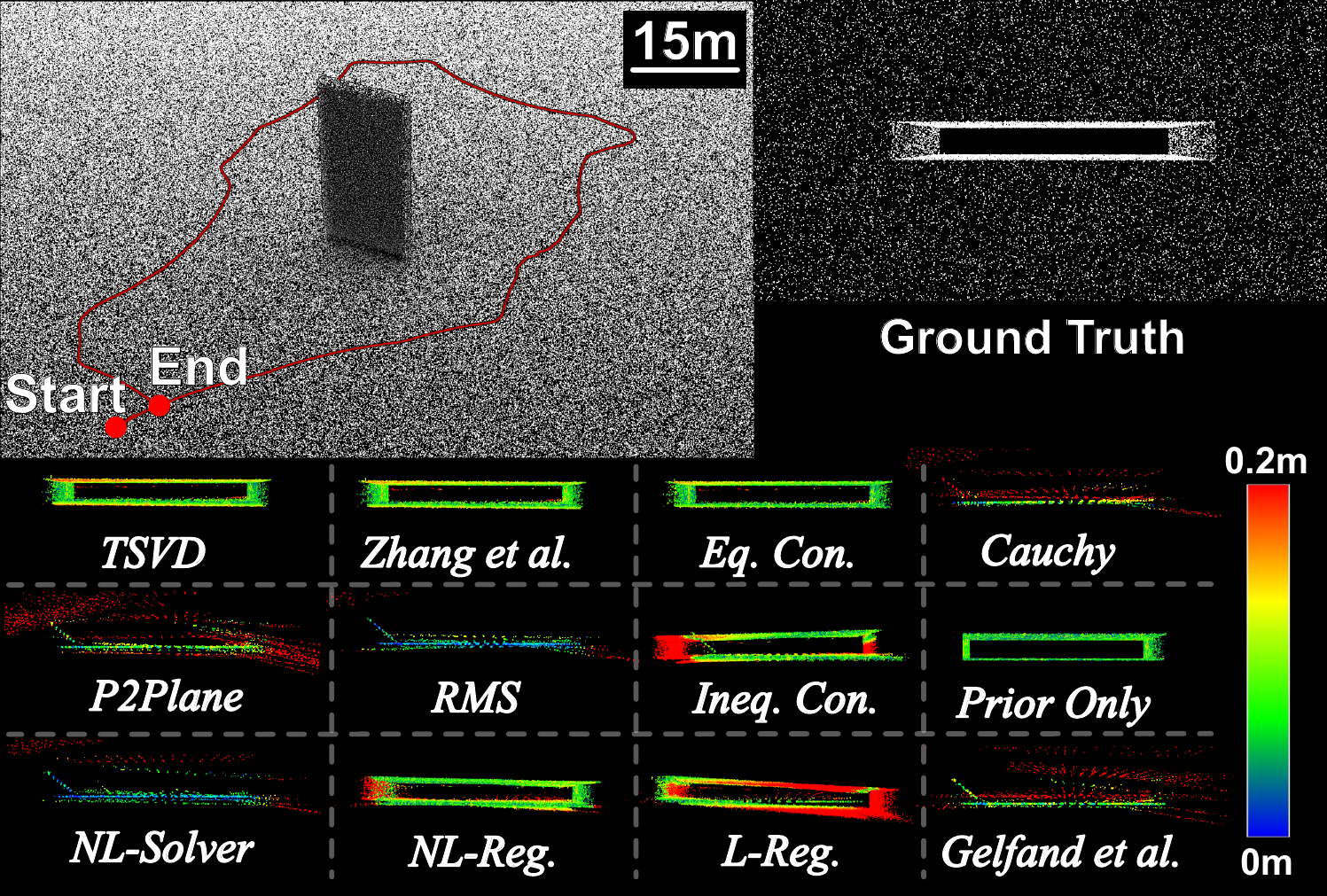}
    \caption{ 
    \textbf{Top:} The ground truth point cloud and trajectory of ANYmal simulation experiment. \textbf{Rest:} The mapping results of the experiment (\Cref{section:Experiments:dynamic_sim}) with ground truth poses as motion prior. The registered maps (with the ground removed) for each method are shown next to an error color bar. Points in the maps are colored according to the point-to-point distances to the ground truth map of the environment. The scale of the color bar is adjusted to make errors visually distinguishable.
    }
    \label{fig:Experiments:anymal_sim_gt}
\end{figure}

\begin{figure}[h!]
    \centering
    \includegraphics[width=1\linewidth]{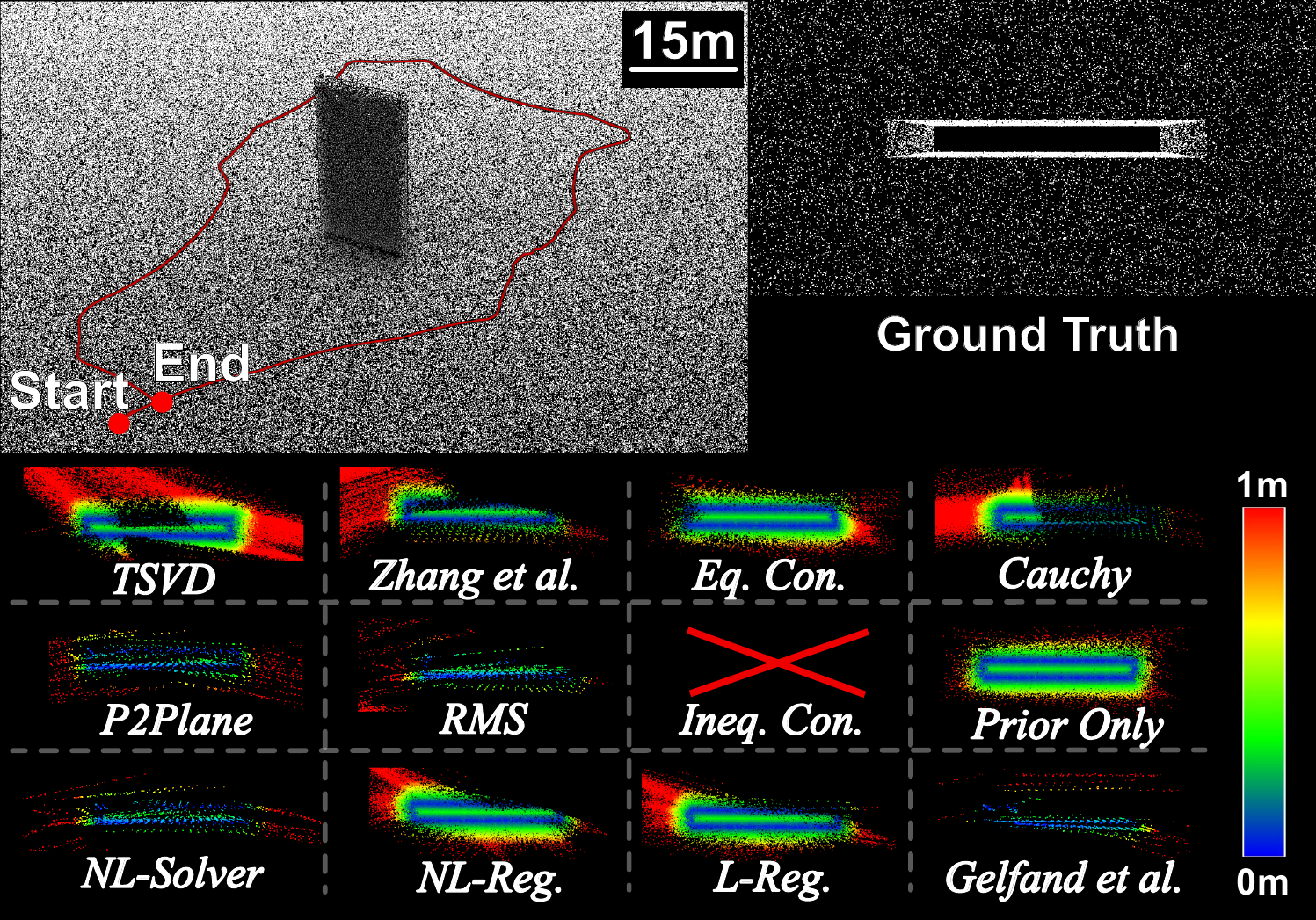}
    \caption{\textbf{Top:} The ground truth point cloud and trajectory of ANYmal simulation experiment, repeated for clarity. \textbf{Rest:} The mapping results of the repetition of the dynamic ANYmal simulation experiment with heavily noised motion prior. The point cloud maps for each method are shown next to an error color bar. Points in the maps are colored according to the point-to-point distances to the ground truth map of the environment.
    }
    \label{fig:Experiments:anymal_sim_heavy}
\end{figure}
Importantly, the overall performance of the \mbox{Prior Only} method indicates that when the motion prior is good (ANYmal simulation experiment) or degeneracy is short in duration (HEAP excavator experiment), skipping the point cloud registration entirely can be a viable solution. 
Using \mbox{Prior Only} can also avoid additional computations while also removing the adverse effects of LiDAR degeneracy from the final solution. 
However, it should be considered that skipping the entirety of the registration would also mean not doing registration in the well-constrained directions, hence losing the LiDAR accuracy. 
Moreover, the sensor-fusion methods described in~\Cref{section:related_works} can be used if the robotic platform has additional sensing modalities such as a camera and a radar. 
This effectively makes the \mbox{Prior Only} much more effective, yet this solution would only offload the problem to other modalities instead of addressing the core problem of optimization instability.

\subsection{IMPORTANCE OF MOTION PRIOR QUALITY}
Accuracy of the motion prior is an important element for the successful mitigation of LiDAR degeneracy. To highlight its importance, an additional experiment is performed as a variation of the ANYmal simulation experiment (\Cref{section:Experiments:dynamic_sim}). First, instead of adding noise to the simulated robot pose, this perfect pose is used as the motion prior. The results in \Cref{fig:Experiments:anymal_sim_gt} show that all active methods perform similarly, with soft-constrained methods (Ineq. Con., NL-Reg. and L-Reg.) being marginally worse. This observation implies that in this severely degenerate scene, degenerate point cloud registration can degrade an otherwise perfect pose prior estimate. It should be noted that here, the point clouds are voxelized at \SI{0.1}{\meter}, which implies that due to the effect of voxelization, the point-to-point error is not expected to be exactly zero.\looseness-1

Second, to highlight the adverse effect of noisy motion prior, increased noise is added to the ground truth robot poses compared to before \Cref{section:Experiments:dynamic_sim}. This noise is sampled from the normal distributions $\mathcal{N}(\mu_t,\,{\sigma_t}^{2})$ and $\mathcal{N}(\mu_r,\,{\sigma_r}^{2})$ where the distribution variables are $\mu_t=\SI{0.05}{\meter}, \ \sigma_t=\SI{0.05}{\meter}$, $\mu_r=\SI{0}{\radian}, \ \sigma_r=\SI{0.02}{\radian}$. 
The results of this variation are shown in \Cref{fig:Experiments:anymal_sim_heavy}. It can be seen through the performance of the \mbox{Prior Only} method that the effect of noise on the quality of the registration is detrimental. Almost all the methods except Eq. Con. diverged, which can be seen through the misaligned point cloud sections of the wall in red. Moreover, Ineq. Con. method also diverged before registering sufficient points on the wall of the environment and, as a result, not visualized. As motion prior is a key element for all the downstream tasks such as feature extraction, correspondence search, and degeneracy detection, a thorough analysis is required to draw any conclusions about which method might be more robust to variations in the motion prior.\looseness-1

\subsection{CONSIDERATIONS AND RECOMMEDATIONS}
Regardless of the performance of LiDAR degeneracy mitigation methods, some additional aspects, such as computational complexity and ease of use, are equally important to consider. In terms of computation, all methods except \ac{NL-Reg.} are close to each other and capable of performing the optimization at the LiDAR sensor rate. 
As the \ac{NL-Reg.} method uses non-linear optimization techniques; it takes more computing power, it can fall behind the LiDAR rate, and consequently drop measurements in exchange for increased accuracy and robustness.
Another consideration to take into account is the need for parameter tuning. 
The methods \ac{NL-Reg.}, \ac{L-Reg.} and \ac{Ineq. Con.} require parameter tuning.
As discussed in~\Cref{section:parameter_tuning}, the performance of these methods changes according to the set parameters.
%
There are applications such as continuous inspection and surveillance in which a robot would need to operate in the same environment multiple times. %
For such applications, the ability to fine-tune and overfit to an environment can be a benefit instead of a drawback.
Similarly, in some applications, such as the generation of metric-semantic digital twins, the accuracy and robustness of the registration can have priority over ease of use, parameter tuning, or computational requirements.\looseness-1

The summary provided in~\Cref{table:summary_table} shows that \ac{Eq. Con.}, \ac{TSVD} and \ac{Ineq. Con.} did not fail in any of the experiments. Meanwhile, the \ac{P2Plane} and Cauchy methods always generated an incorrect map.
Despite failing in the ANYmal dynamic simulation experiment, \ac{NL-Reg.} method generated accurate results. The method of Zhang \textit{et al.} performed comparably with misregistration in the ANYmal simulation and the ANYmal forest experiments.

As a recommendation, if more computational resources and the ability to fine-tune are available, \ac{NL-Reg.} method is a good option. If deployed on a computationally limited system, with fine-tuning availability \ac{Ineq. Con.} method has great potential since this method (thanks to the formulation of the QP problem) can also accept equality constraints. On the other hand, if fine-tuning is not an option, \ac{Eq. Con.} method performed reliably throughout the experiments with good accuracy however, the performance of this method is tied to the accuracy of the motion prior. If the freedom to choose an optimization is desired, Zhang \textit{et al.} and the \ac{TSVD} method showed the same reliability in exchange for accuracy but without the addition of hard-constraints to the optimization problem (e.g., no need for QP formulation, Lagrangian, Ceres, and sampling). As a sampling approach, RMS~\cite{petracek2024rms} provided a speed-up to the framework throughout all experiments and performed the best in the urban setting. Although this method did not perform well in other experiments, if the environment is geometrically feature-rich, it could be a strong candidate for many frameworks to mitigate the effect of mild degeneracy while providing substantial speed-up.\looseness-1

The results and the recommendations done in this work are to provide an objective view of the problem of degenerate point cloud registration. However, the reader should beware of the merits of the system used in this work (\textit{Open3D-SLAM}~\cite{jelavic2022open3d}), which uses a voxel-based mapping backend, external pose prior, and scan-to-submap registration as described in~\ref{subsection:testingSetup}.\looseness-1

\section{CONCLUSION AND FUTURE WORK}\label{section:conclusion}
This work analyzed the efficacy of degeneracy mitigation methods in various real-world scenarios and simulated setups.
New methods such as \ac{Ineq. Con.}, \ac{TSVD}, and \ac{L-Reg.} are proposed and introduced to the field of degeneracy-aware point cloud registration.
In addition, previously introduced methods such as \ac{Eq. Con.}~\cite{xicp}, Solution Remapping~\cite{solRemap}, Cauchy robust norm~\cite{babin2019analysis}, RMS~\cite{petracek2024rms} and Gelfand\textit{et al.}~\cite{gelfand2003} are compared.\looseness-1

{\color{black}As discussed in~\Cref{section:Experiments} and summarized in~\Cref{table:summary_table}, active degenerate mitigation is crucial and beneficial for severe LiDAR degeneracy, in mild cases of LiDAR degeneracy, passive methods, such as sampling methods, can provide the necessary robustness. As a result, selecting the LiDAR degeneracy mitigation method is a problem that depends, at least, on the severity of the LiDAR degeneracy and the availability of a good motion prior.} 

Among the active degeneracy mitigation methods, solutions such as \ac{TSVD}, \ac{Eq. Con.}, and \ac{Ineq. Con.} perform more consistently. Compared to methods such as Zhang \textit{et al.}, \ac{L-Reg.} and \ac{NL-Reg.} methods. {\color{black}On the other hand, among the passive degeneracy methods RMS~\cite{petracek2024rms} performed the best in the mild degeneracy case while providing a speed-up.}
The robust norm method, Cauchy, improved the performance of the SLAM system in the absence of LiDAR degeneracy as previously highlighted~\cite{babin2019analysis}; however, it consistently failed among all datasets when LiDAR degeneracy was present. The positive effect of non-linear optimization, in exchange for the additional computational cost, is shown through the comparison between the P2Plane and NL-Solver baseline methods.\looseness-1

{\color{black}As the focus of this work is on robust point cloud registration, multi-modal sensor-fusion solutions are not covered in this work; however, in practice, sensor-fusion is often employed in robotic deployments complimentary modalities can resolve the necessity of the LiDAR degeneracy mitigation. The fusion of ill-conditioned poses in sensor-fusion frameworks is still an open-resource question.}
%
Lastly, globally optimal point cloud registration methods are studied for degenerate point cloud registration problem. As shown in~\Cref{fig:tunnel_global} and~\Cref{section:Experiments:global_sim}, global registration methods are not compatible drop-in solutions for LiDAR degeneracy, as the assumption of the existence of informative inliers is often violated. These results supported the claim that LiDAR degeneracy is not an outlier removal problem (e.g. Cauchy method did not perform well) but rather an absence of an informative inliers problem.\looseness-1

{\color{black}An interesting future direction is to utilize active and passive methods together while acknowledging the strengths and weaknesses of each category. An example would be to combine NL-Reg. and RMS methods to compensate for the computational load of NL-Reg. with the speed-up provided by RMS.}\looseness-1

Similarly, detailed analysis of the effect of fine-tuning methods \ac{L-Reg.}, \ac{NL-Reg.}, and \ac{Ineq. Con.} and ways to automate the selection of these parameters is still an open research question. These methods have the potential to perform better if manual environment specific parameter selection requirement is removed or mitigated.

{\color{black}As an outlook, ICP is still a dominant solution in point-cloud registration, and many researchers focus on addressing the weaknesses that arise from optimization degradation. The problem originates from the lack of sufficient inlier information provided to the least-squares optimization through steps such as surface-normal extraction and K-Nearest Neighbors. A one-fits-all solution does not seem to be possible without quantifying the errors in these steps. As a result, the future steps in ICP-based robust point-cloud registration in degenerate cases might lie in metric uncertainty estimation for each step or reduction of the importance of these steps with novel research (cf. new loss formulation, a new optimization technique, data-driven correspondence search method).}
%

\begingroup
\small
\setlength{\itemsep}{0pt}
\setlength{\parsep}{0pt}
\renewcommand{\baselinestretch}{0.95}\selectfont
\bibliographystyle{IEEEtran}
\bibliography{IEEEabrv,REFERENCES}
\endgroup

\begin{IEEEbiography}[{\includegraphics[trim={22 0 22 0},clip,height=1.15in,keepaspectratio]{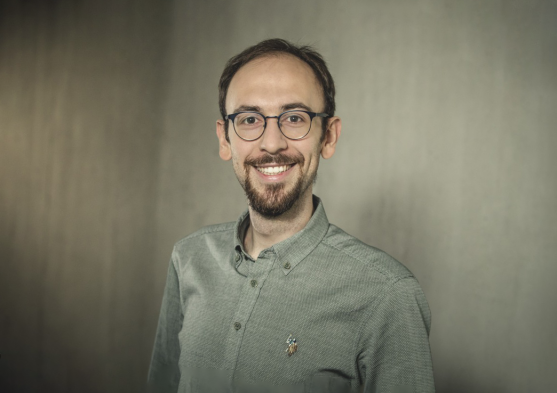}}]%
{Turcan Tuna} \: is a PhD student in the Robotic Systems Lab at ETH Zurich. He received his M.Sc. in Robotics, Systems \& Control in 2022 from ETH Zurich. Previously, he completed a double major, B.Sc in Mechanical Engineering and Control \& Automation Engineering, at Istanbul Technical University. He graduated from both of his B.Sc majors with distinction. His research interests include developing and deploying robust localization, perception, and mapping frameworks on robotic systems.
\end{IEEEbiography}

\begin{IEEEbiography}
[{\includegraphics[trim={236 0 236 0},clip,height=1.10in,keepaspectratio]{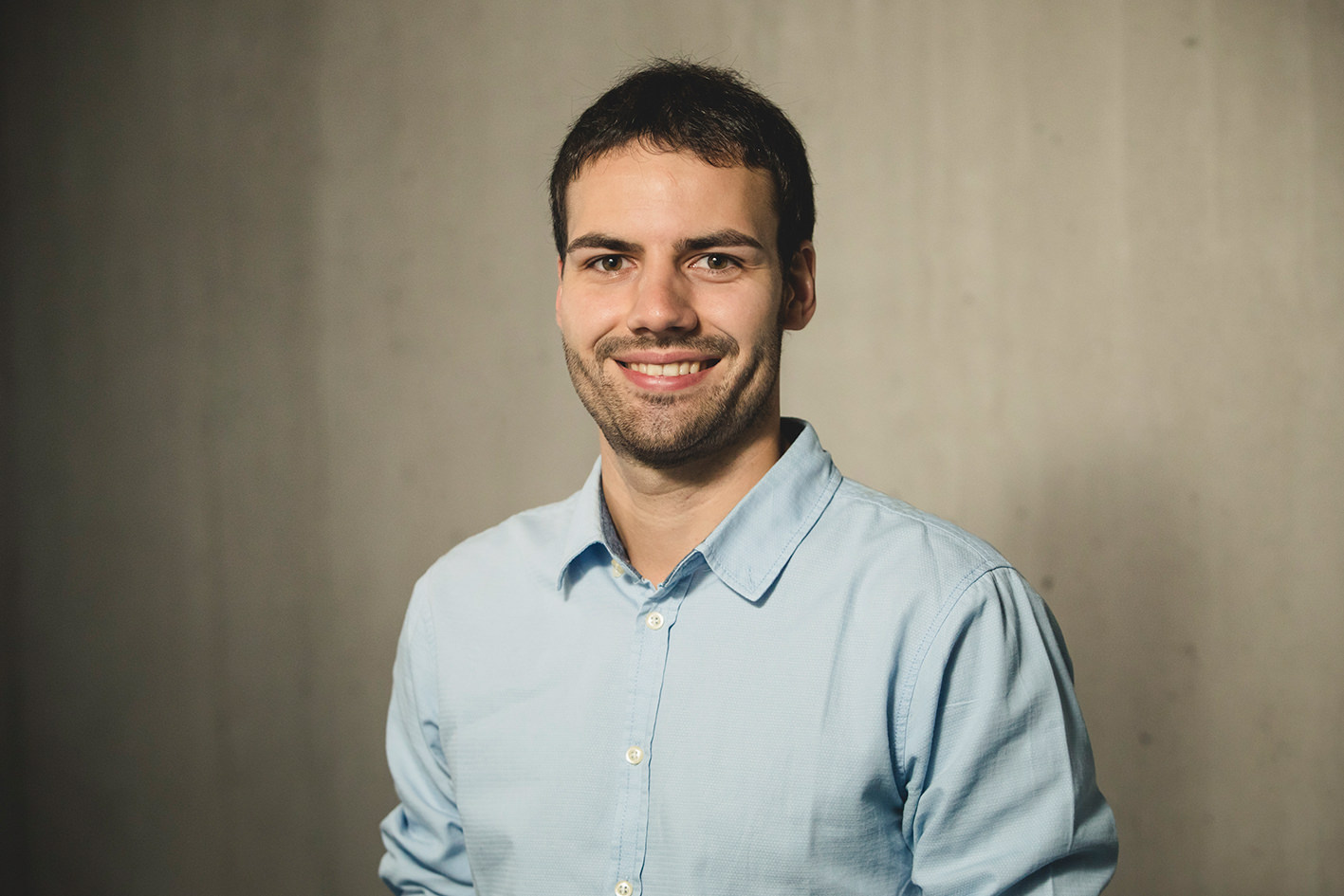}}]%
{Julian Nubert} \: is a PhD student in the Robotic Systems Lab at ETH Zurich. He received his M.Sc. in Robotics, Systems \& Control in 2020 from ETH Zurich. He is also affiliated with the Max Planck Institute through the MPI ETH Center for Learning Systems. His research interests lie in the field of robust robot perception, and how it can be used for the deployment of mobile robotic systems. Julian received the ETH silver medal and was awarded the Willi-Studer-Price for his achievements during his master studies.
\end{IEEEbiography}

\begin{IEEEbiography}
[{\includegraphics[trim={0 0 0 0},clip,height=1.1in,keepaspectratio]{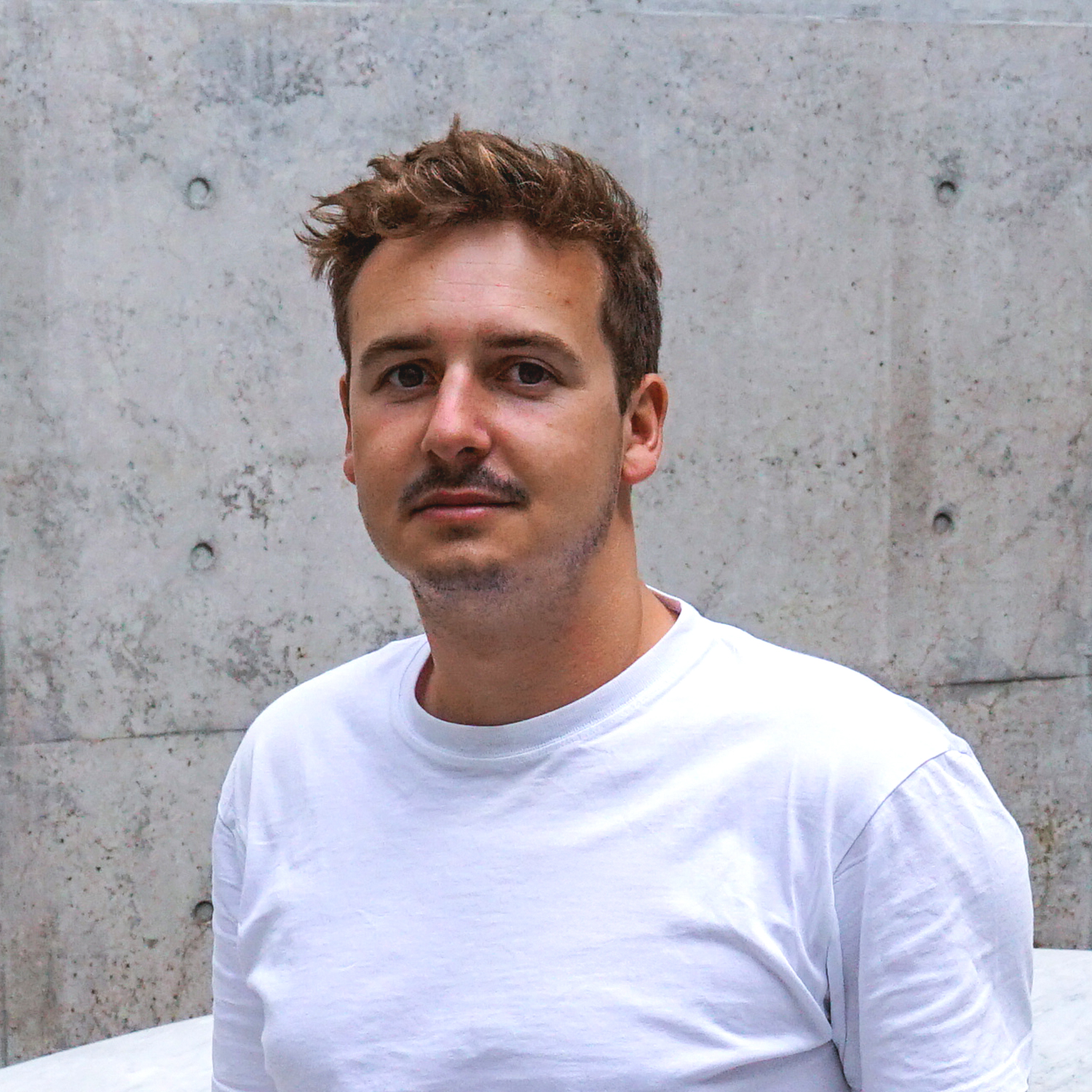}}]%
{Patrick Pfreundschuh} \: is a PhD student in the Autonomous Systems Lab at ETH Zurich. He received his M.Sc. in Robotics, Systems \& Control in 2020 from ETH Zurich. In his research, he focuses on robust LiDAR odometry in challenging environments. Previously, he worked as a member of team CERBERUS, which won the DARPA SubT Challenge (2021).
\end{IEEEbiography}

\begin{IEEEbiography}[{\includegraphics[trim={236 0 236 0}, clip,height=1.1in,keepaspectratio]{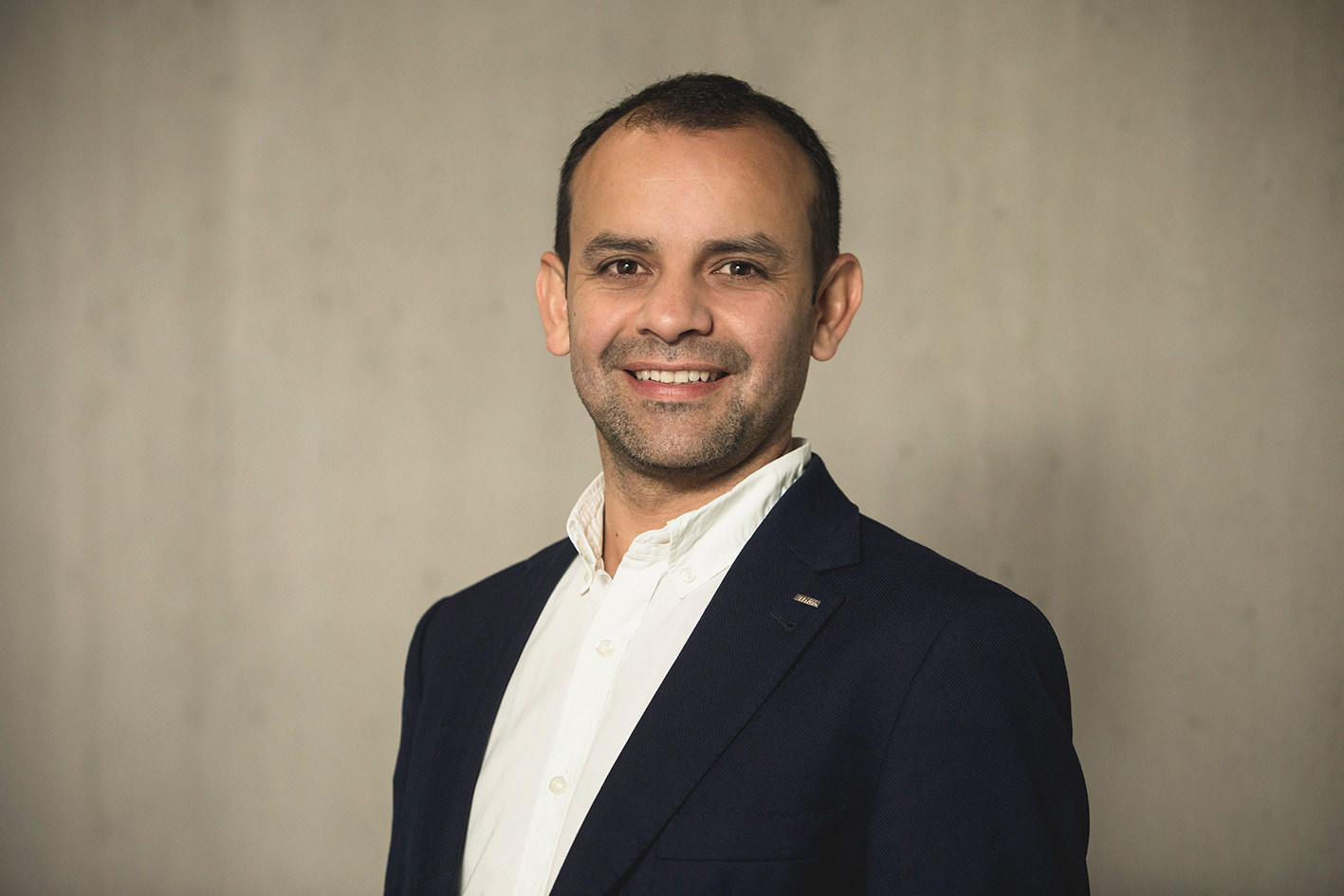}}]%
{Cesar Cadena} \: received his PhD (2011) in computer science from the University of Zaragoza, Spain. He is a Senior Scientist at ETH Zurich. His research interests are in providing machines the ability to understand and interact with our ever-changing world. His work covers robotic scene understanding, both geometry and semantics, covering semantic mapping, data association, and place recognition tasks, simultaneous localization and mapping, as well as persistent mapping in dynamic environments and robot navigation.
\end{IEEEbiography}

\begin{IEEEbiography}[{\includegraphics[trim={236 0 236 0}, clip,height=1.1in,keepaspectratio]{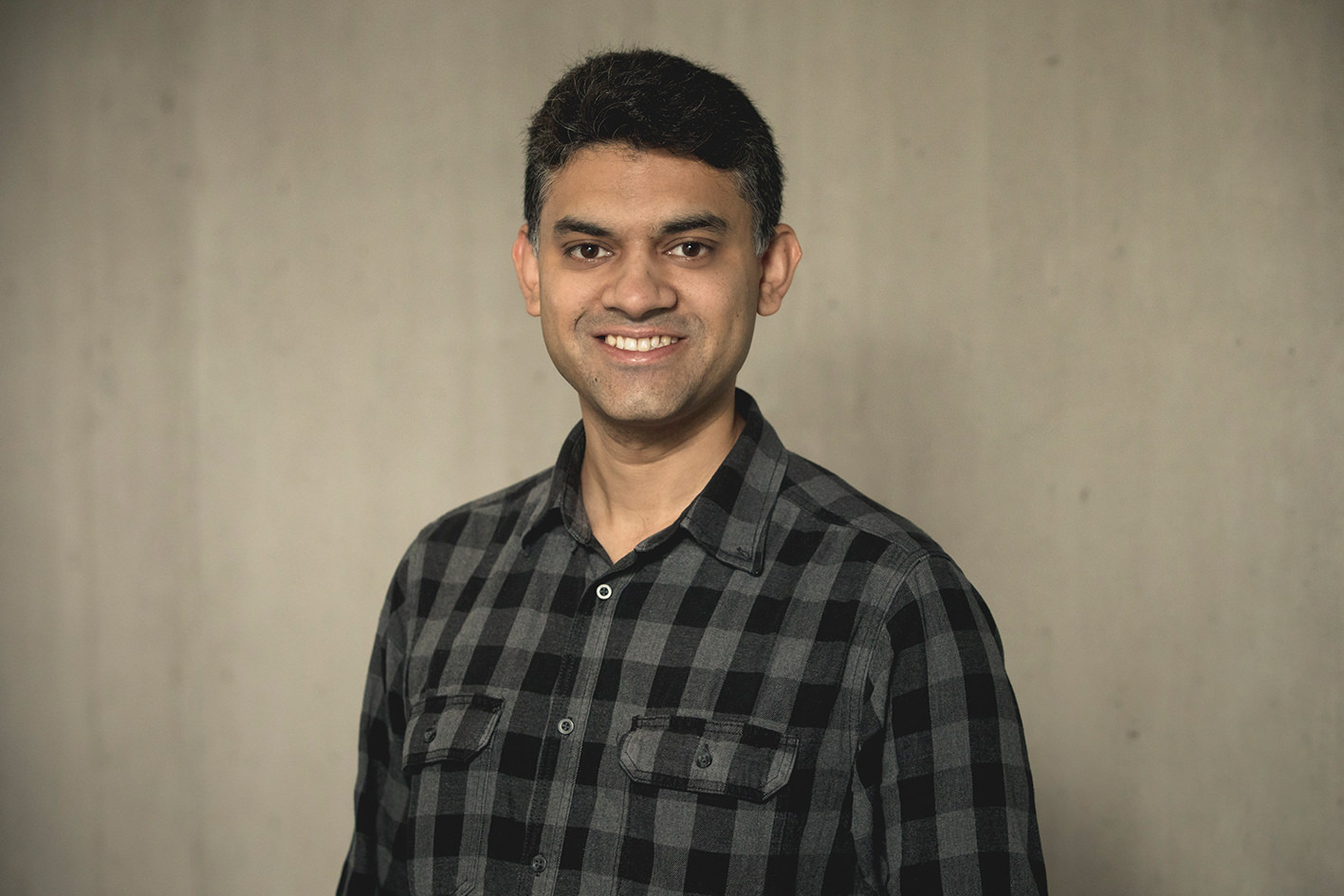}}]%
{Shehryar Khattak}  \: is a Robotics Technologist in the Perception Systems Group at NASA's Jet Propulsion Laboratory (JPL). His work focuses on enabling resilient robot autonomy in complex environments through multi-sensor information fusion. Currently, he is the Principal Investigator (PI) for the Multi-robot Autonomous Intelligent Search and Rescue project at JPL. He previously served as the perception lead for JPL's team in the DARPA RACER project and for Team CERBERUS, winners of the DARPA Subterranean Challenge. Before joining JPL, Shehryar was a postdoctoral researcher at ETH Zurich. He earned his Ph.D. (2019) and M.S. (2017) in Computer Science from the University of Nevada, Reno. Additionally, he holds an M.S. in Aerospace Engineering from KAIST (2012) and a B.S. in Mechanical Engineering from GIKI (2009).
\end{IEEEbiography}

\begin{IEEEbiography}[{\includegraphics[trim={236 0 236 0}, clip,height=1.1in,keepaspectratio]{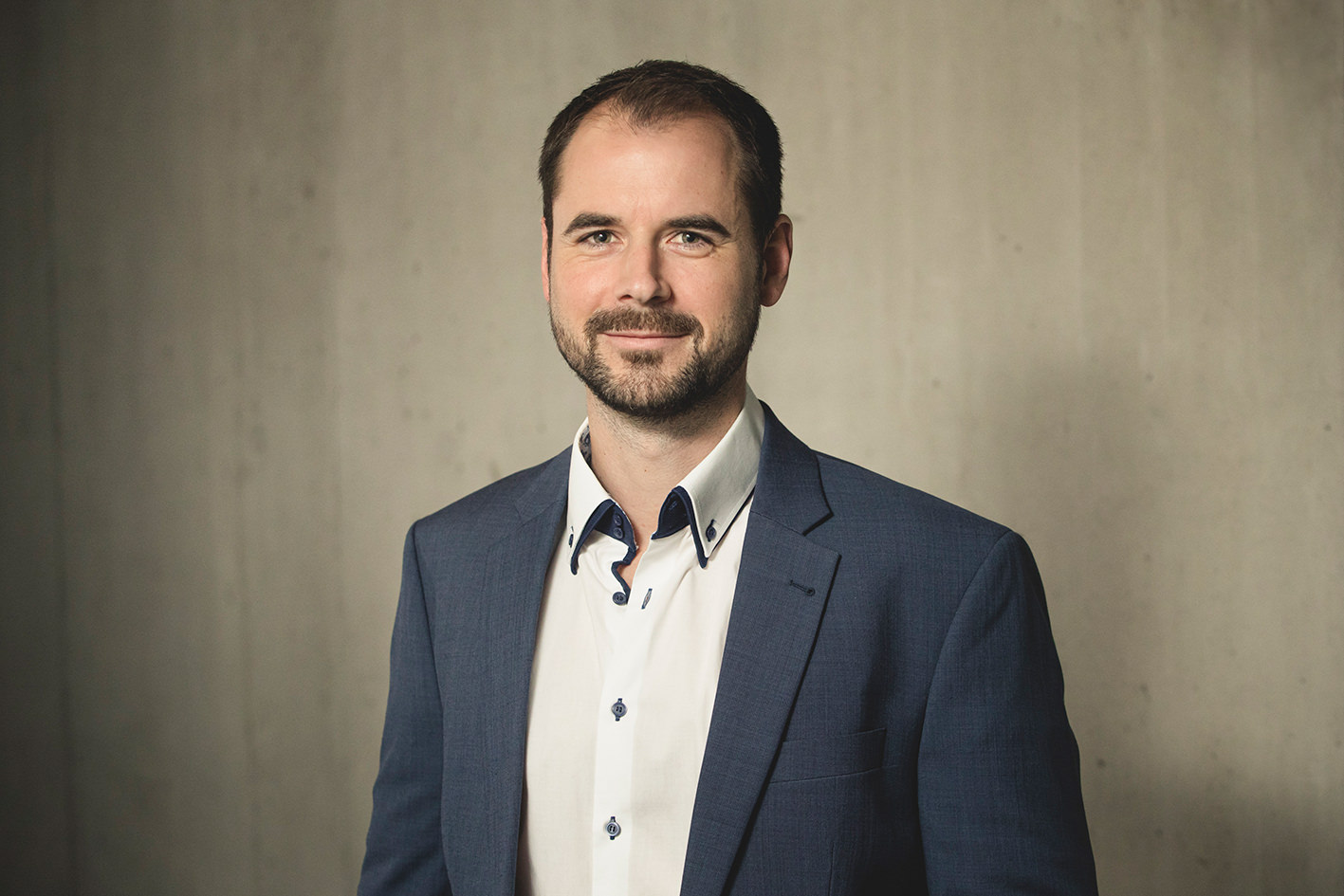}}]%
{Marco Hutter}  \: is an Associate Professor for Robotic Systems at ETH Zurich. He received his M.Sc. and PhD from ETH Zurich in 2009 and 2013 in the field of design, actuation, and control of legged robots. His research interests are in the development of novel machines and actuation concepts together with the underlying control, planning, and machine learning algorithms for locomotion and manipulation. Marco is the recipient of an ERC Starting Grant, PI of the NCCRs robotics and digital fabrication, PI in various EU projects and international challenges, a co-founder of several ETH Startups such as ANYbotics AG.
\end{IEEEbiography}

\vfill\pagebreak

\end{document}